\documentclass{article}

\usepackage[preprint,nonatbib]{neurips_2023}
\usepackage[utf8]{inputenc} %
\usepackage[T1]{fontenc}    %
\usepackage{url}            %
\usepackage{booktabs}       %
\usepackage{amsfonts}       %
\usepackage{nicefrac}       %
\usepackage{microtype}      %
\usepackage{lipsum}         %
\usepackage{graphicx}
\usepackage{doi}
\usepackage{amsmath}
\usepackage{amssymb}
\usepackage{amsthm}
\usepackage{bbding}
\usepackage{multirow}
\usepackage{subcaption}
\usepackage{algorithm}
\usepackage{algorithmic}
\usepackage{cleveref}       %
\usepackage{enumitem}
\usepackage{makecell}   %
\usepackage{color}
\usepackage{hyperref}       %
\usepackage{minitoc}
\usepackage{amssymb}
\usepackage[numbers]{natbib}   %

\title{Hacking Generative Models with Differentiable Network Bending}
\author{
Giacomo Aldegheri\thanks{Correspondence to \href{mailto:giacomo.aldegheri@gmail.com}{giacomo.aldegheri@gmail.com}}
\\University of Amsterdam\\
\And
Alina Rogalska\thanks{These authors contributed equally.}
\\Independent \\Researcher\\
\And
Ahmed Youssef\footnotemark[2]\\
University of \\ Cincinnati
\And
Eugenia Iofinova\footnotemark[2] \\
IST Austria
}
\date{August 2023}

\begin{document}

\maketitle

\begin{abstract}
    In this work, we propose a method to 'hack' generative models, pushing their outputs away from the original training distribution towards a new objective. We inject a small-scale trainable module between the intermediate layers of the model and train it for a low number of iterations, keeping the rest of the network frozen. The resulting output images display an uncanny quality, given by the tension between the original and new objectives that can be exploited for artistic purposes. Project website: \url{https://galdegheri.github.io/diffbending/}.
\end{abstract}

\begin{figure}[h!] \
\begin{center}
\includegraphics[width=\textwidth]{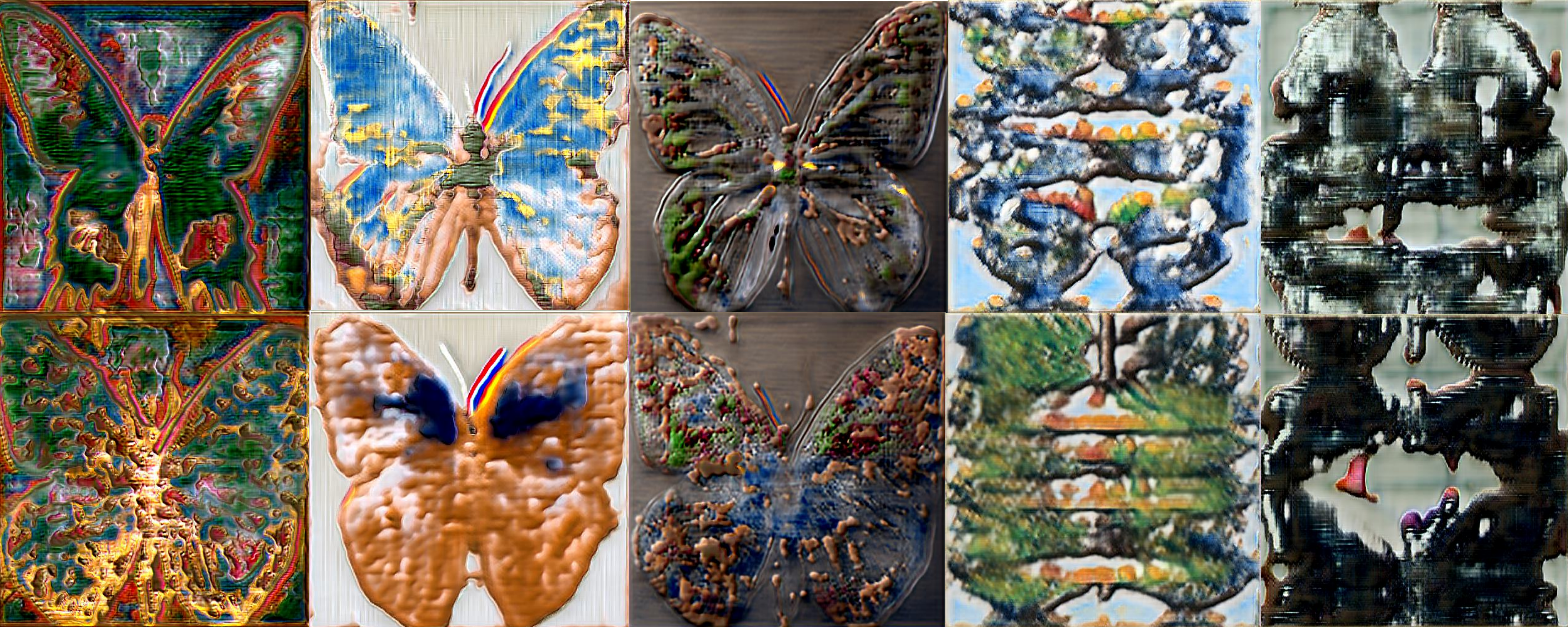}
\caption{Example outputs, using a variety of loss functions and bending modules. See Appendix~\ref{sec:full_examples} for the corresponding prompts and more examples.}
\label{fig:results}

\end{center}
\end{figure}

\section{Introduction}
Systems that fail to perform their intended task, or that are repurposed for new tasks, have long been recognized for their subversiveness and creative potential. Glitch artists~\cite{menkman2011glitch} explore the aesthetic properties of media malfunctions, while hackers~\cite{erickson2008hacking} take pleasure in the challenge of pushing existing hardware and software beyond its intended function~\cite{kotzer2016doom}. In this work, we hack existing generative models to generate images they were not originally trained for. We adapt \emph{network bending}~\cite{broad2021network}, a technique consisting of injecting a transformation between intermediate layers of a generator, by making the transformation (the \emph{bending module}, BM) differentiable. 
The images generated by these hacked models are a blend of the objects that the original model was trained to generate (butterflies) and new visual features introduced by the BMs. We find that they exhibit an uncanny quality that can be exploited for creative purposes, similar to glitch art's use of unintended media artifacts.

Our method, thanks to its low computational cost, is accessible to a wide variety of artists and practitioners. Unlike current state of the art text-to-image pipelines, which aim to generate perfect-looking images, it provides a tool to explore strange and unexpected variations of existing models.


\section{Method}

\begin{figure}[h] \
\begin{center}
\includegraphics[width=1\textwidth]{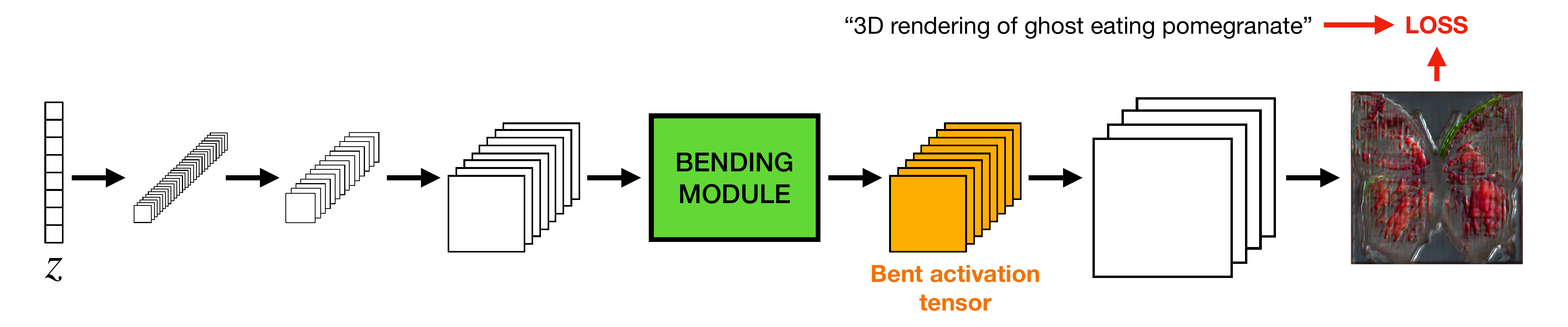}
\caption{Overview of our proposed method. The BM takes the activation map of any chosen layer as input, and outputs a 'bent' activation map (shown in orange) which is fed as an input to the subsequent layers.}

\end{center}
\end{figure}

\paragraph{Architecture.} The differentiable BM can be injected after any layer of a generator network. Here we use a lightweight GAN based on the architecture in~\cite{liu2020towards} and trained on a dataset of butterfly images\footnote{This model is available at \url{https://huggingface.co/ceyda/butterfly_cropped_uniq1K_512}.}. We chose a model trained on a narrow domain of images in order to be able to clearly distinguish the effect of the BM from the model's original structure. Specifically, in our outputs the outline of the butterfly remains visible to varying degrees depending on the specific BM used. The BM takes as input the activation map of the chosen layer, and outputs a tensor of the same dimensionality. We use a variety of small-scale network architectures for our BMs (see \textbf{Appendix} for architectural details): \textbf{(1) Convolutional:} a plain convolutional neural network (CNN), using either $ReLU(x)$ or $\sin(x)$ as an activation function. \textbf{(2) Convolutional + Coordinates:} a CNN with spatial coordinates ($x$, $y$ and optionally $r$, the distance from the center) concatenated with the input features. This allows the BM to generate spatially-varying structures. \textbf{(3) Convolutional + Sorting:} a differentiable sorting network~\cite{petersen2022monotonic}, operating across the width and/or height of the input feature map, followed by a CNN. This allows the BM to rearrange the spatial structure of the input activation map.



\paragraph{Loss function.} While the differentiable BM can be trained with any objective, here we experiment with two loss functions. The first one is the squared great circle distance between the CLIP~\cite{radford2021learning} embeddings of the output images and of a user-provided prompt, to generate semantically evocative images. The second one is a loss function that minimizes the distance between images and prompt in CLIP space, while maximizing the distance among different images in the batch, to increase output diversity in a semantically meaningful space. Following the approach of~\cite{yu2022towards}, we use the InfoNCE~\cite{oord2018representation} contrastive objective to maximize the mutual information between image and caption embeddings, while minimizing that among images in the batch, as expressed by the following equation:
\[  \mathcal{L}_{NCE} = -\log\frac{ e^{(Q \cdot K^{+}/\tau)}}{e^{(Q \cdot K^{+}/\tau)} + \sum_{K^{-}} e^{(Q \cdot K^{-}/\tau)} }  \]

Where $Q$ is the image embedding, $K^{+}$ the prompt embedding, $K^{-}$ are the embeddings of the other images in the batch, and $\tau$ is a temperature hyperparameter.

\section{Ethical Implications}
The work was done entirely on open-source or publicly available models and data. In particular, we avoided, as much as possible, the use of tools and data that use the uncompensated work of (traditional) artists. The only possible exception is the use of the pretrained CLIP model, for which we were not aware of alternatives.

Further, since our method provides only a very coarse control of the output, we do not believe that it aids the creation of works that break copyright or produce obscene or incendiary images, over what is already available.

\bibliography{bibliography.bib}
\clearpage

\appendix

\section*{Acknowledgments}
We would like to thank Viorica Patraucean, Razvan Pascanu, and the entire staff of the Eastern European Machine Learning Summer School '22 for bringing us together, and Viorica in particular for her support in getting this project off the ground. We would also like to thank Piotr Mirowski for his advice and feedback on our early ideas, and the former members of the project for their enthusiasm and ideas in the brainstorming sessions. In particular, Ivan Vrkic, for setting up the initial Google Colab environment.

\section{Network architecture and training}
\label{sec:net_details}
The BMs can be applied at any layer of a generator network. In all experiments reported here, we keep this fixed to the 6th layer (for the convolutional BMs) and to the 4th layer (for the Convolutional + Sorting BM) of the ButterflyGAN generator. These layers have $64$ and $256$ channels, respectively. However, we encourage users to experiment with applying the BMs to different layers. 

All the BMs used here share a basic CNN architecture, comprising two convolutional layers with 3x3 kernels, 'same' padding (to keep the size of the output the same as the input) and a number of input, hidden and output channels equal to those of the chosen layer's activation map. The activation function ($ReLU(x)$ or $\sin{x}$ is applied after the first convolutional layer. In the coordinate-aware BM, the $x$ and $y$ coordinates indicating locations on the activation map (scaled between $0-1$) are concatenated with the input features along the channel dimension. The Convolutional + Sorting BM comprises a CNN that assigns scores to each row of the activation map, followed by a Bitonic differentiable sorting layer~\cite{petersen2022monotonic}, with a steepness hyperparameter fixed at $50$, that sorts these scores in increasing order. The permutation matrix generated by the sorting layer is then used to reorder the input activation map along the chosen dimension (width or height). The sorting network is followed by a standard convolutional BM.

All networks are trained for a fixed number of $1000$ iterations, with batch size $16$, using the Adam~\cite{kingma2014adam} optimizer with a learning rate of $1e-3$. Training takes around 3 minutes on a single NVIDIA GeForce RTX 3080 Ti.

\section{Sample outputs for a range of queries}
\label{sec:full_examples}

In the next pages, we present the first 16 example outputs (not filtered, sorted, or cherry-picked in any way) for a range of prompts and training configurations.

\begin{figure}[h] \
\begin{center}
\setlength{\tabcolsep}{2pt}
    \begin{tabular}{cccc}
      \includegraphics[width=0.24\linewidth]{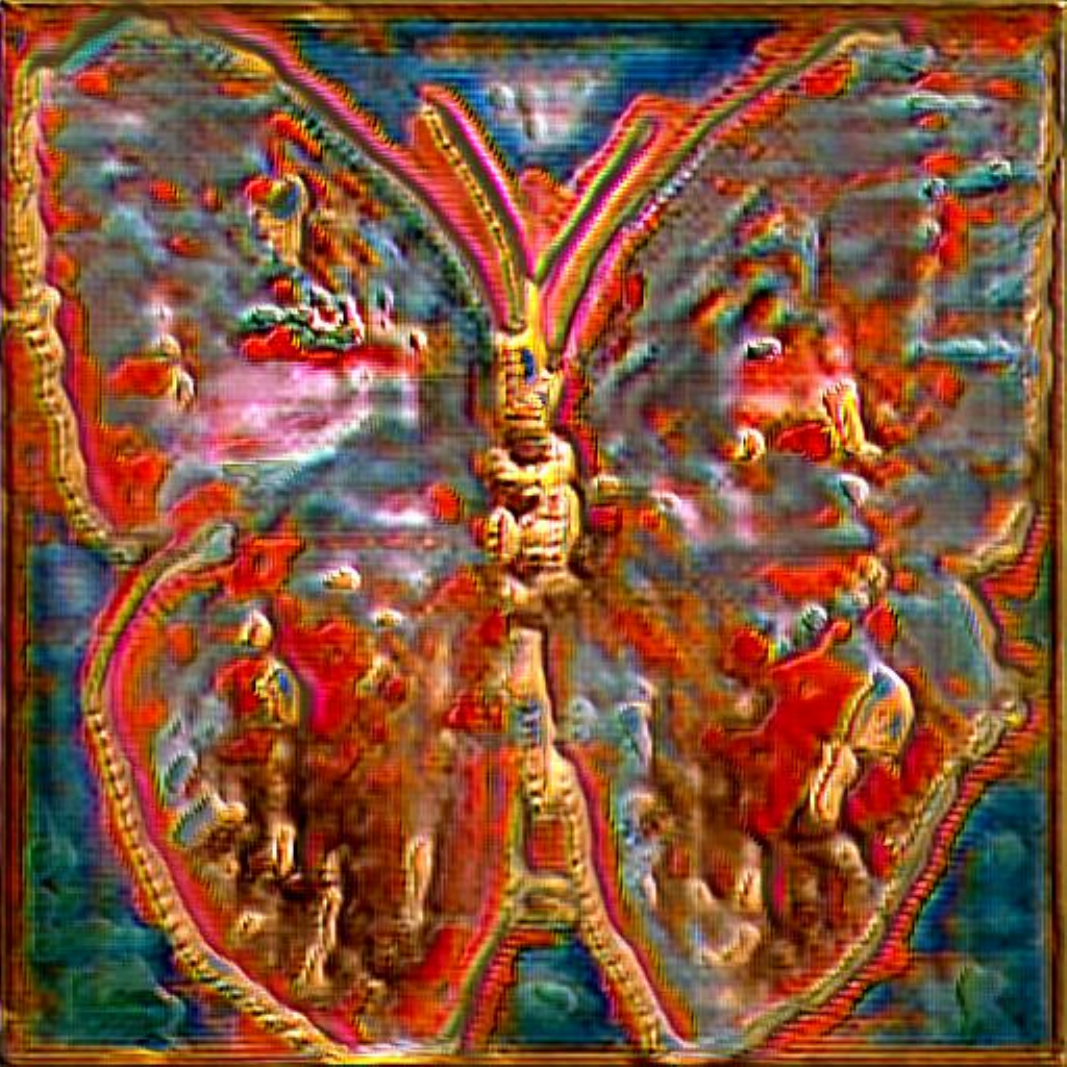}   &
      \includegraphics[width=0.24\linewidth]{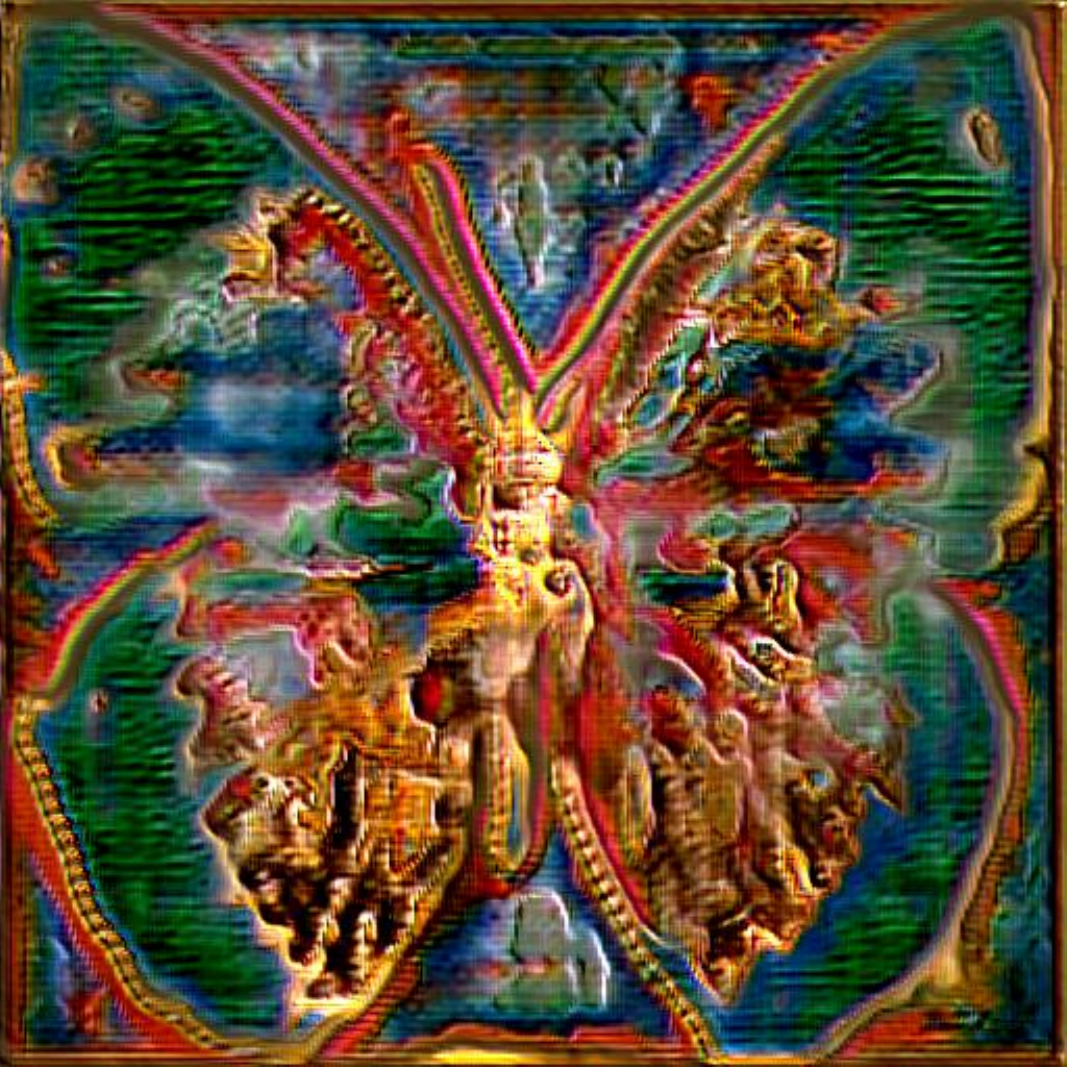}   &
      \includegraphics[width=0.24\linewidth]{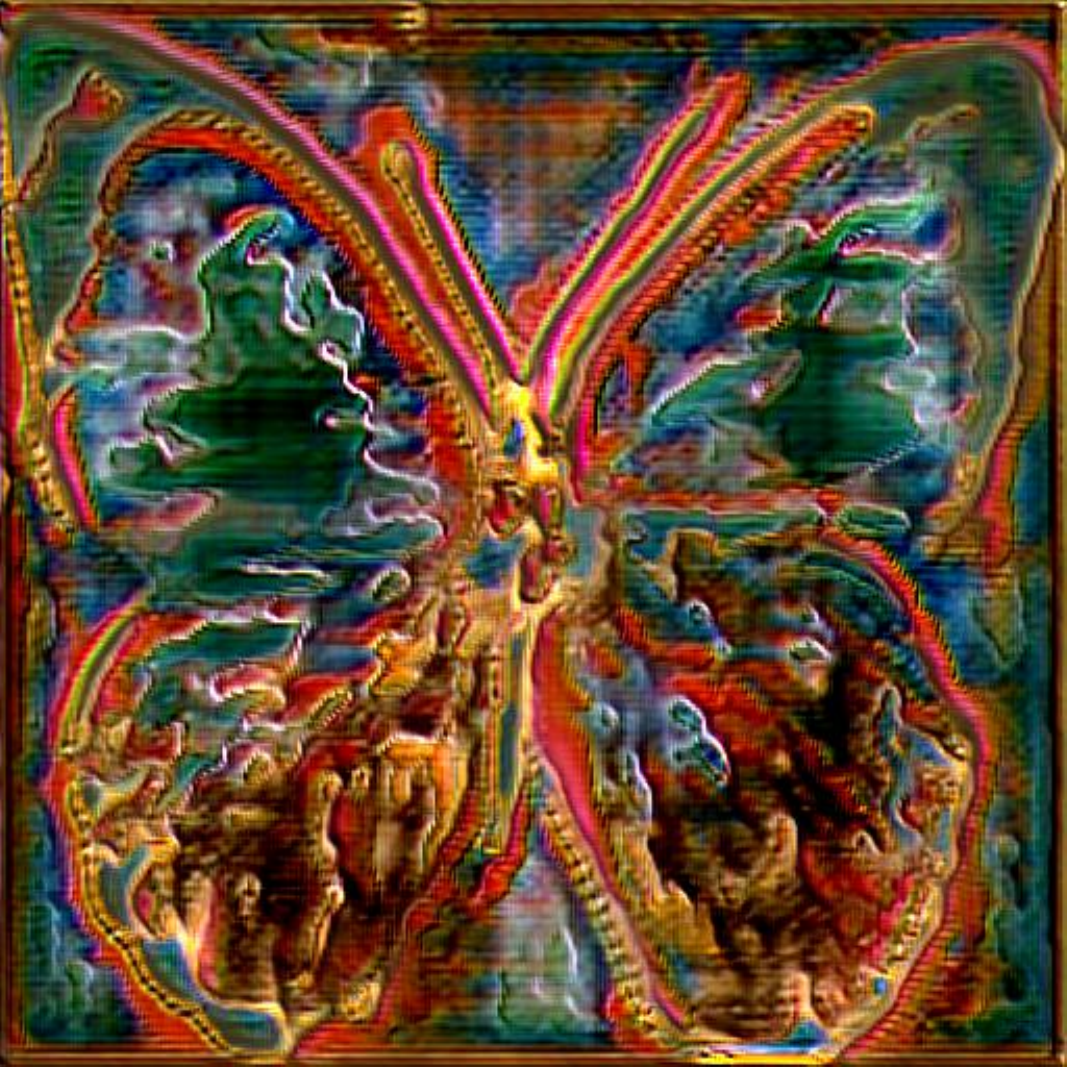}   &
      \includegraphics[width=0.24\linewidth]{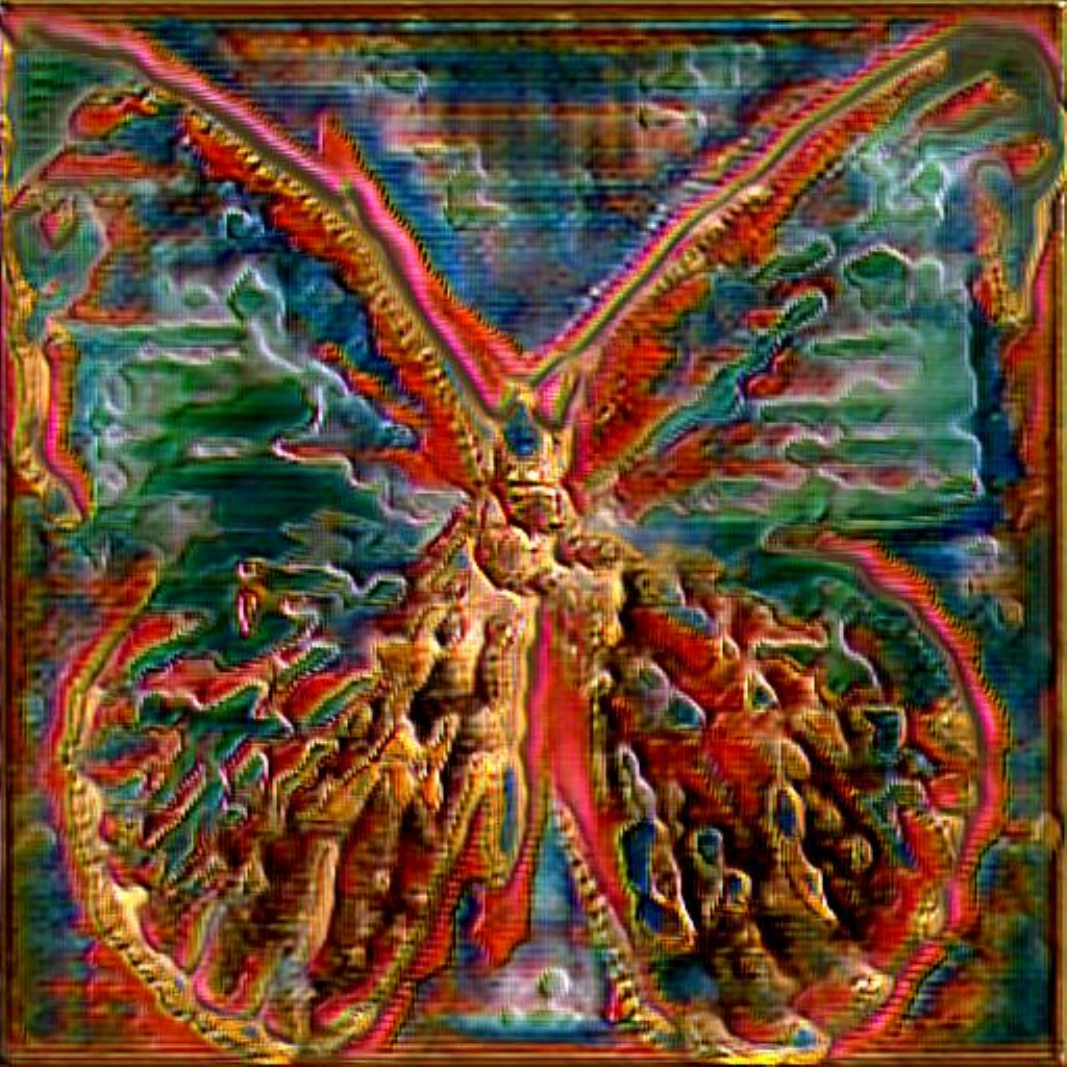}   \\
      \includegraphics[width=0.24\linewidth]{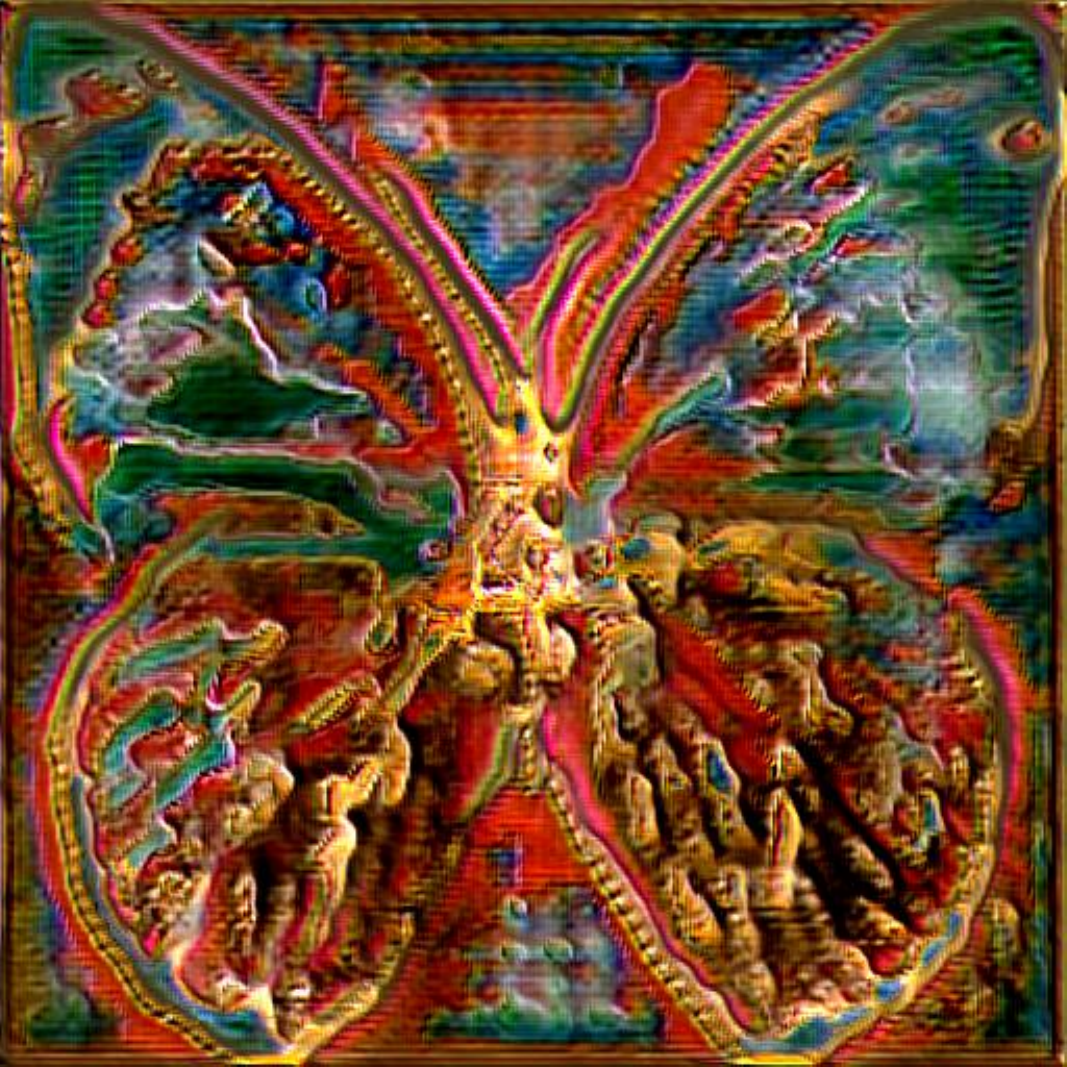}   &
      \includegraphics[width=0.24\linewidth]{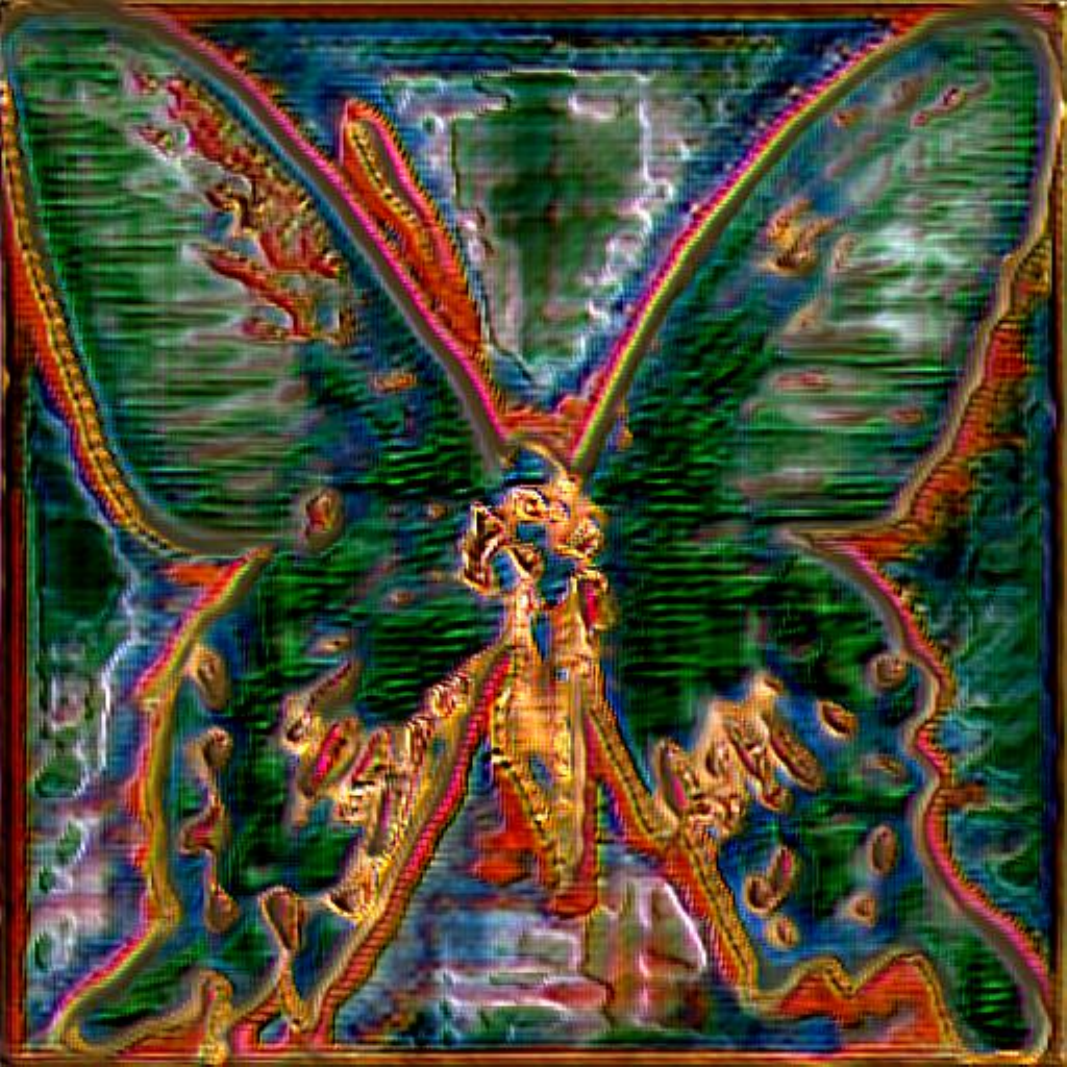}   &
      \includegraphics[width=0.24\linewidth]{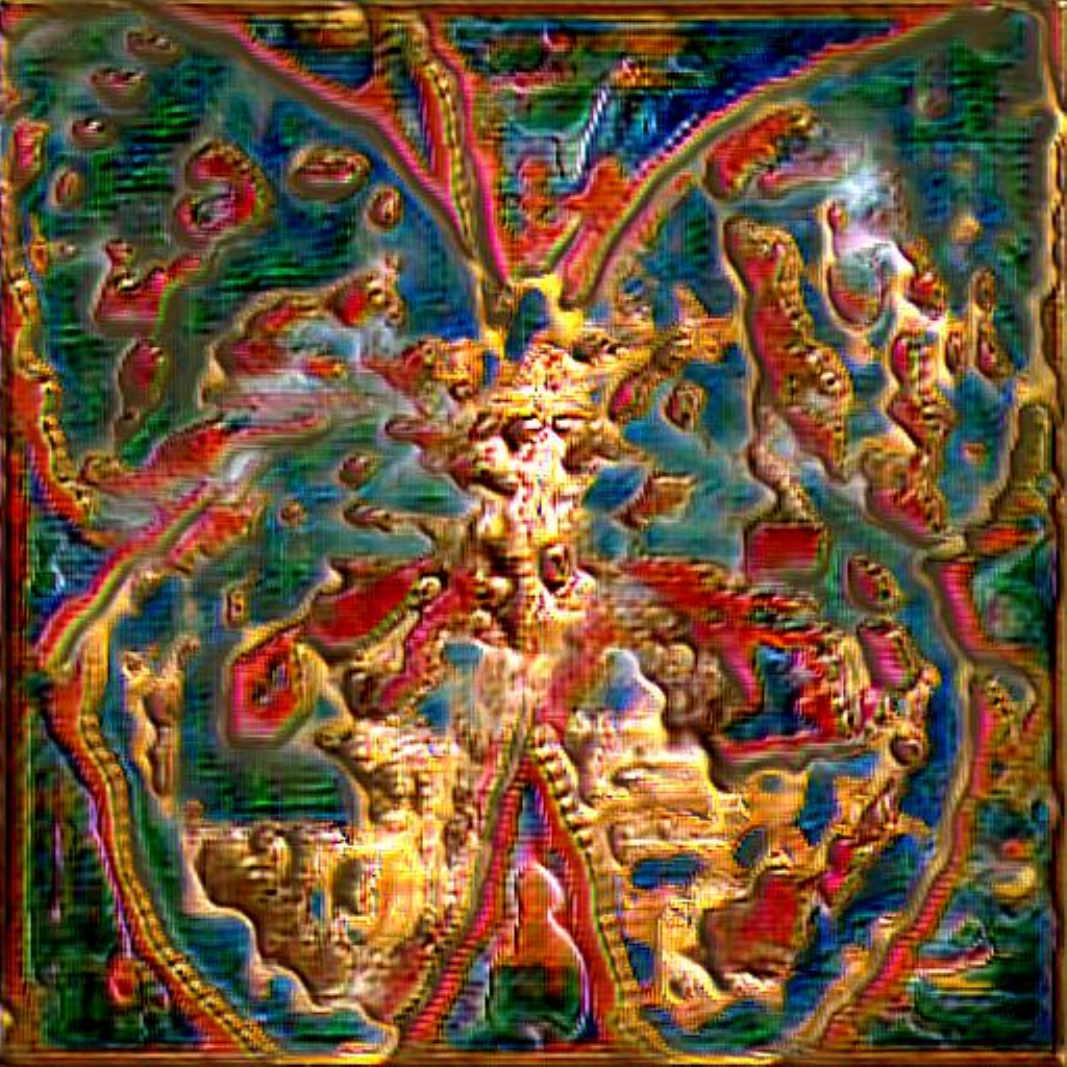}   &
      \includegraphics[width=0.24\linewidth]{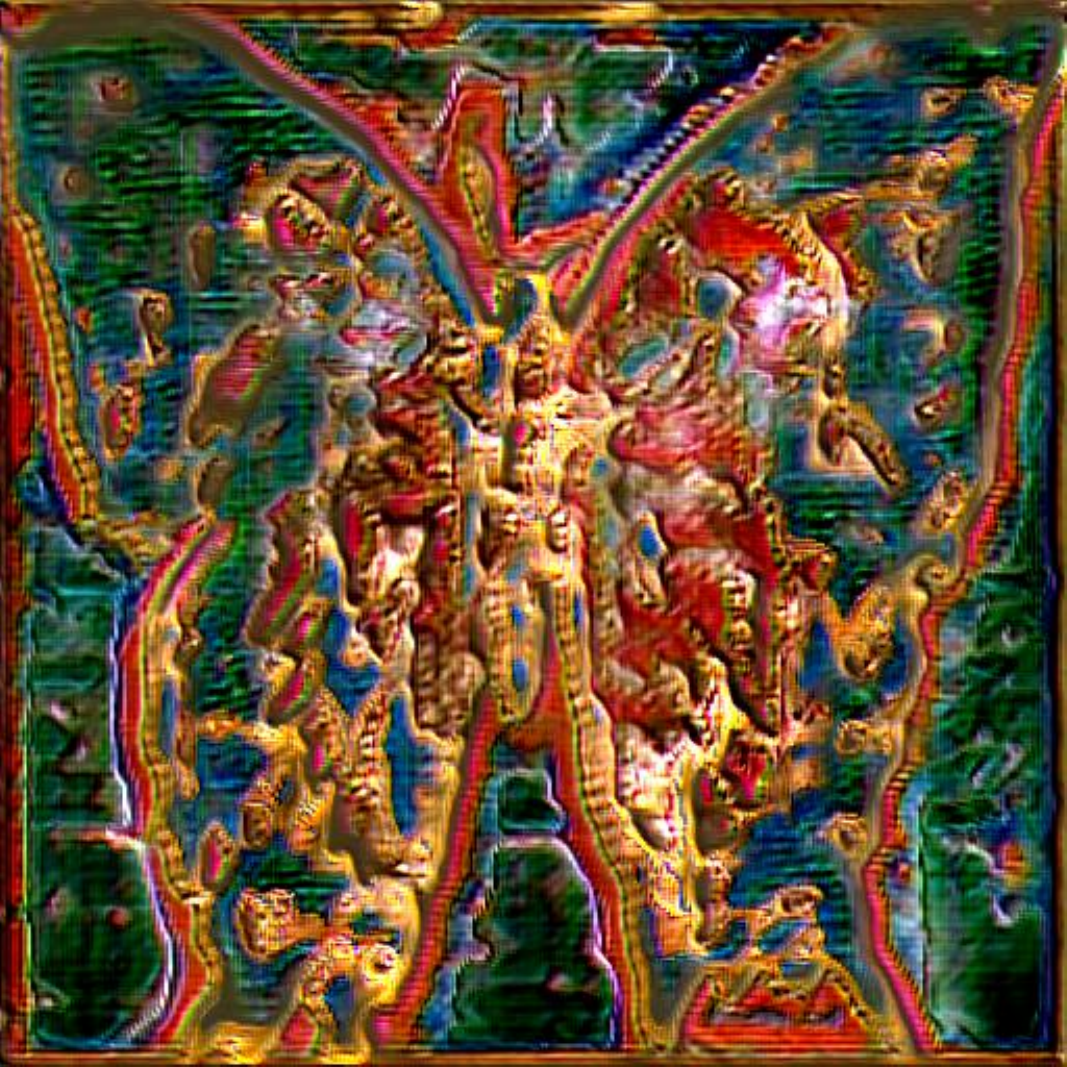}   \\
      \includegraphics[width=0.24\linewidth]{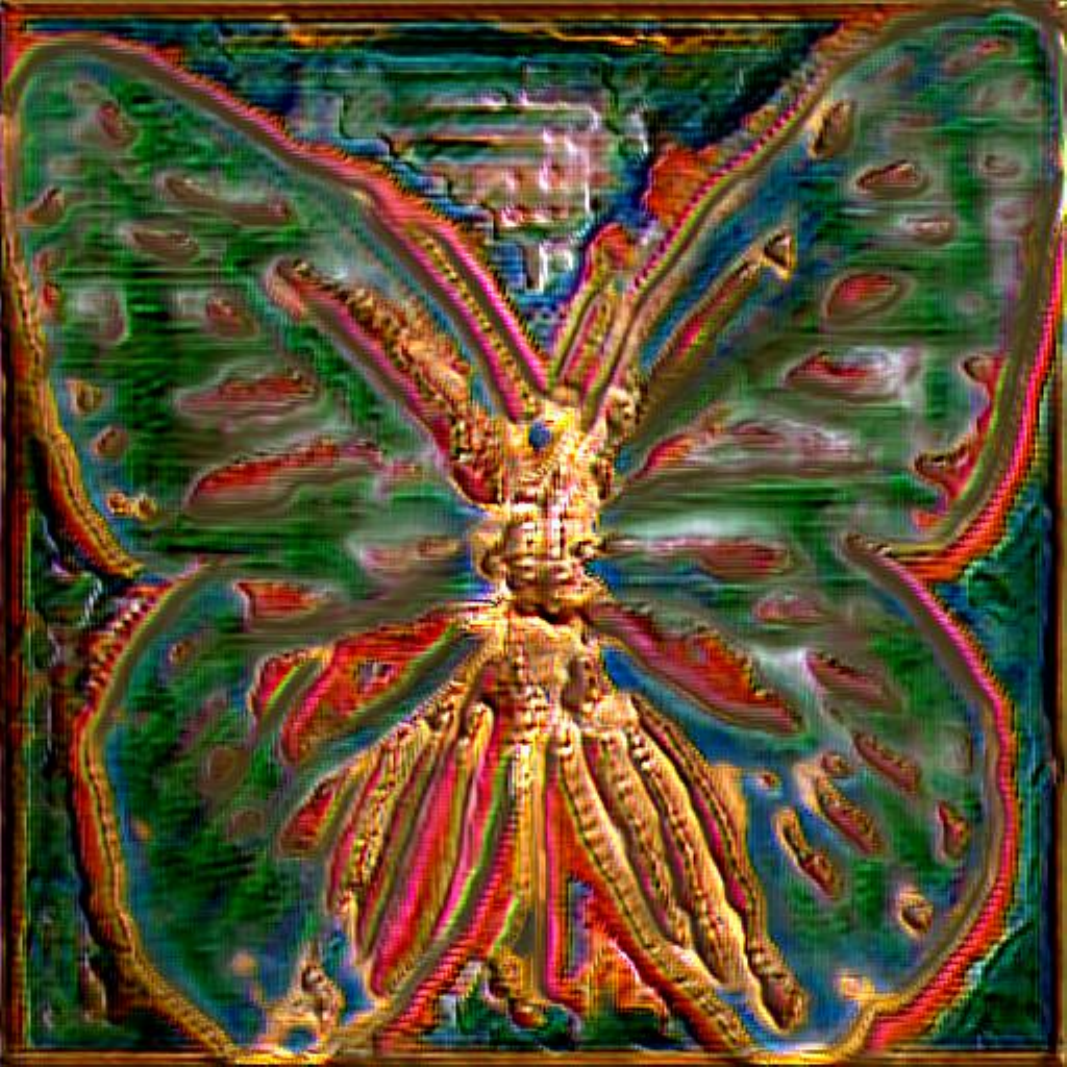}   &
      \includegraphics[width=0.24\linewidth]{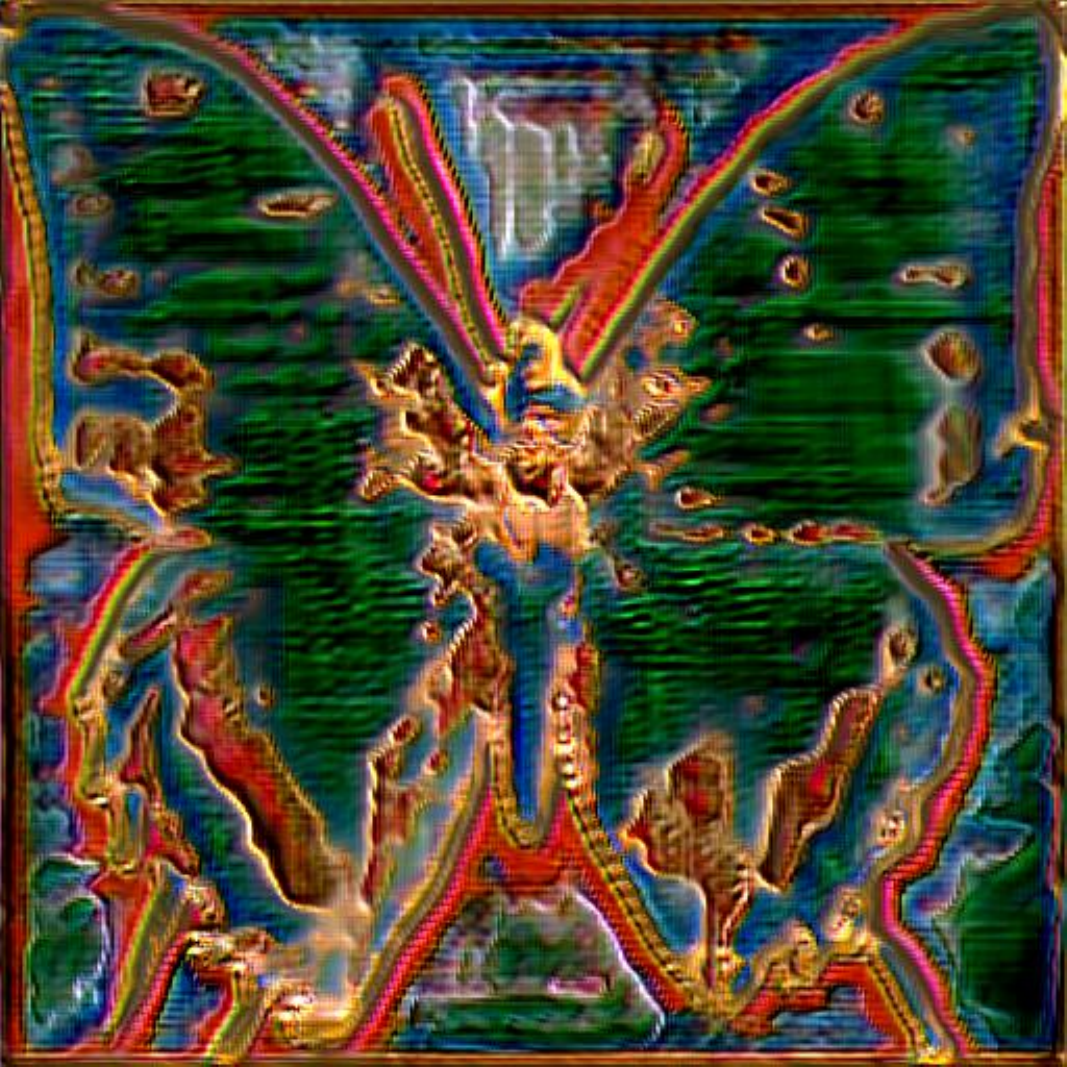}   &
      \includegraphics[width=0.24\linewidth]{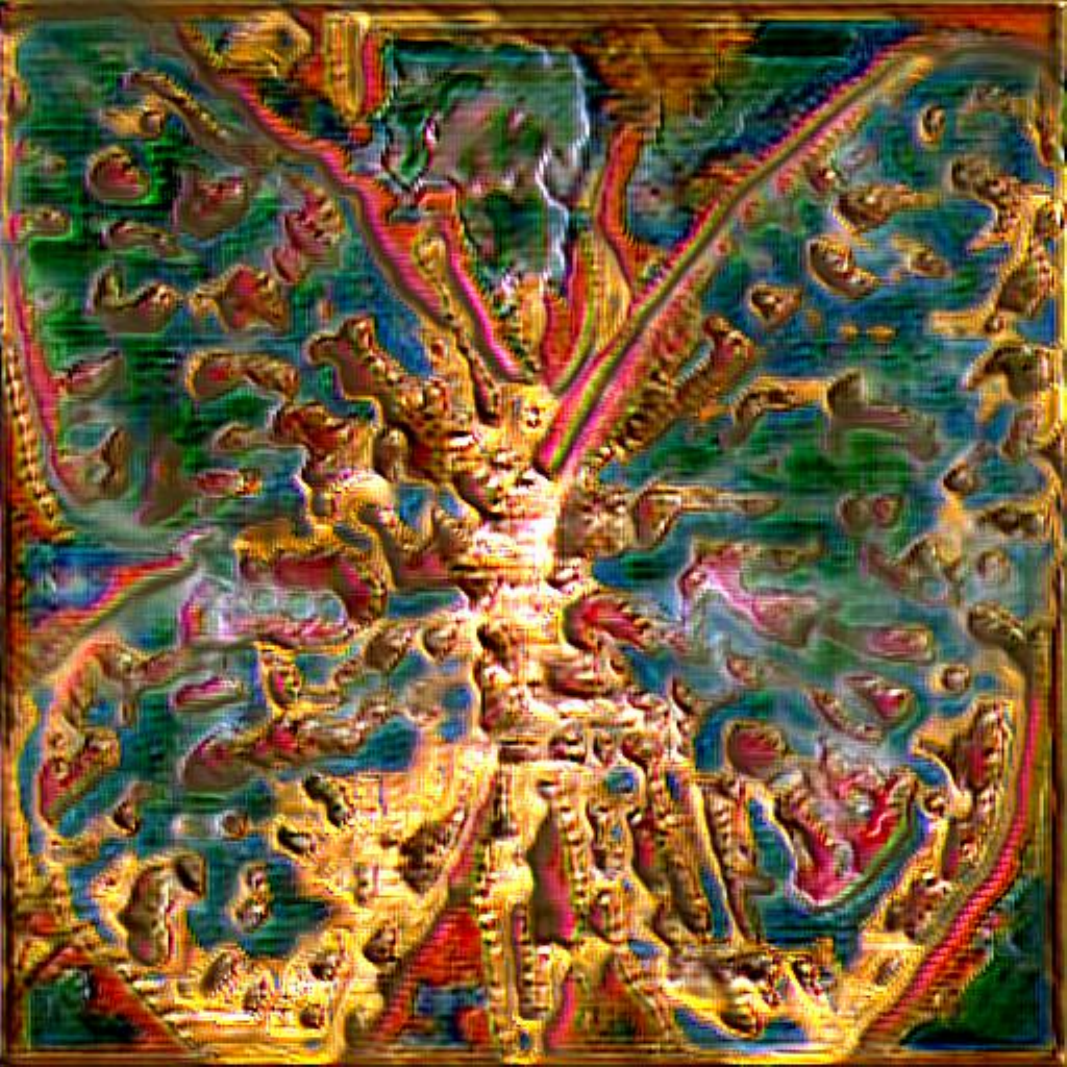}   &
      \includegraphics[width=0.24\linewidth]{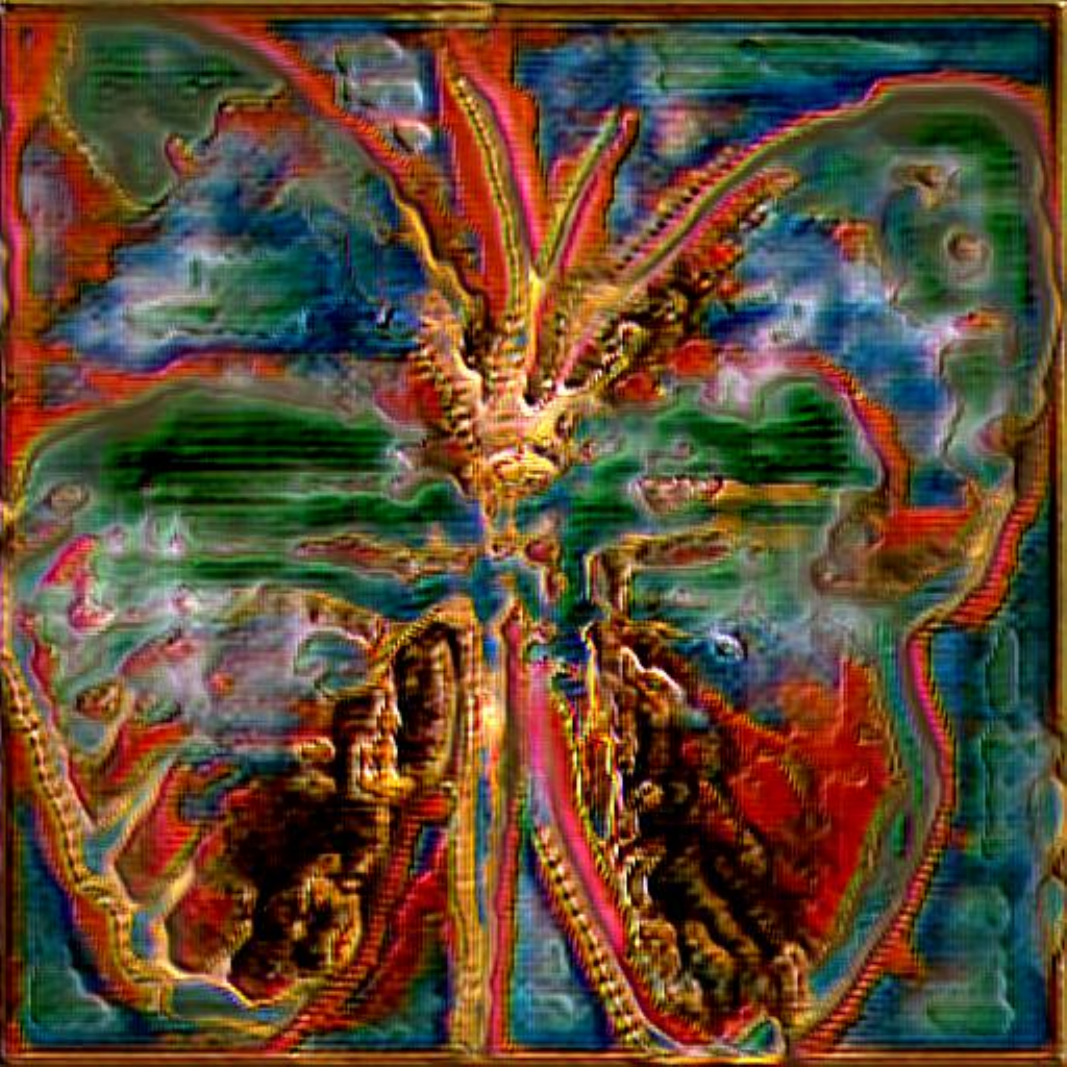}   \\
      \includegraphics[width=0.24\linewidth]{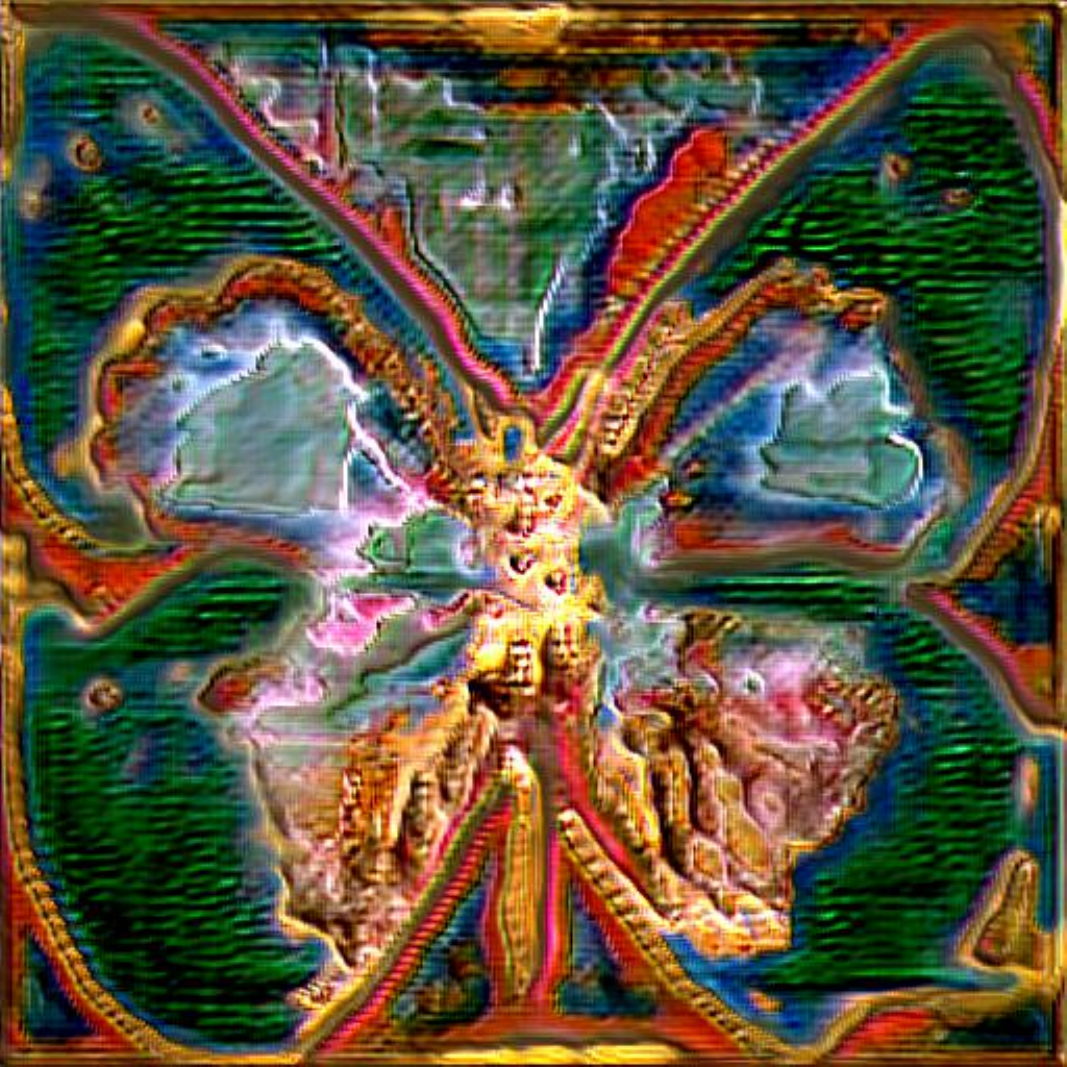}   &
      \includegraphics[width=0.24\linewidth]{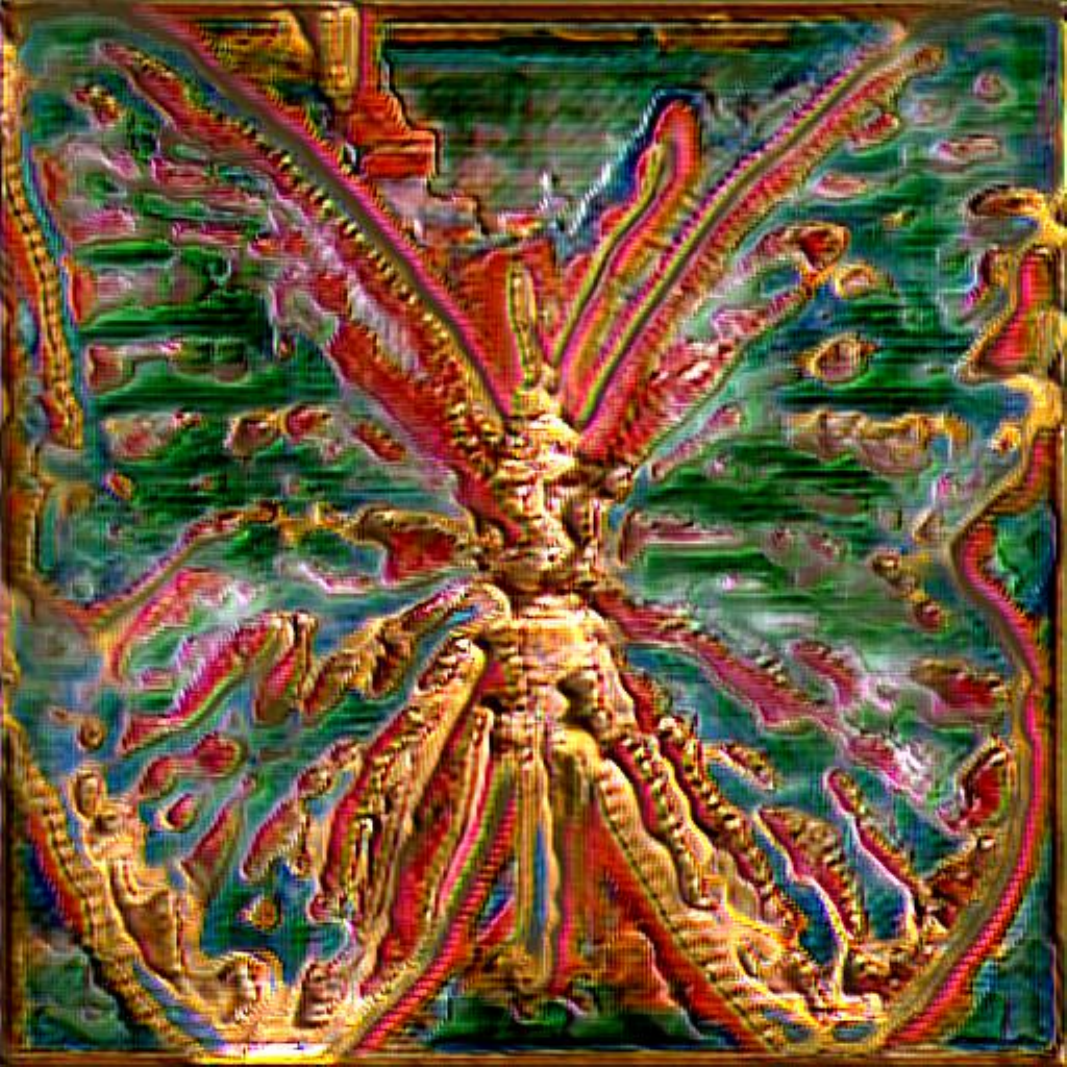}   &
      \includegraphics[width=0.24\linewidth]{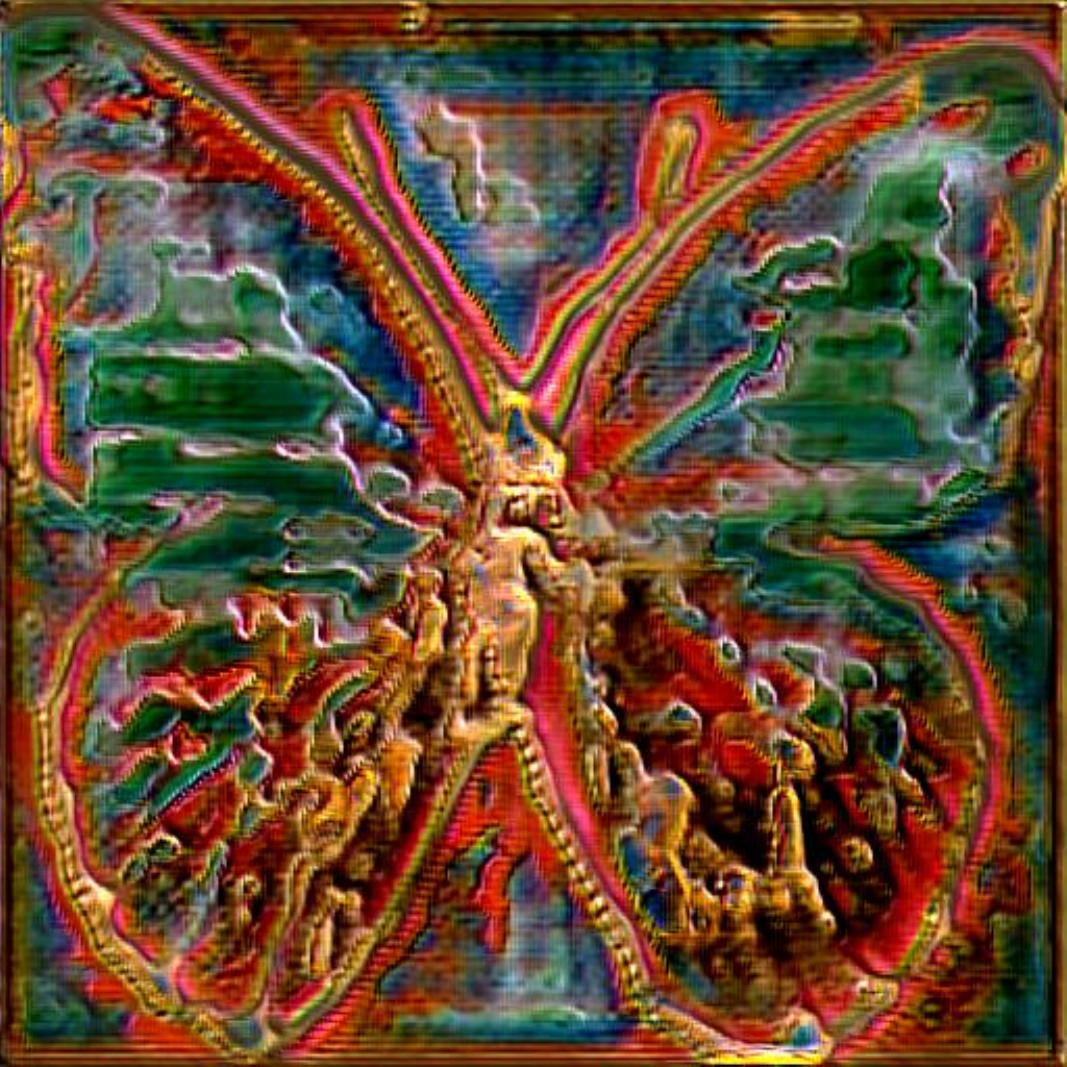}   &
      \includegraphics[width=0.24\linewidth]{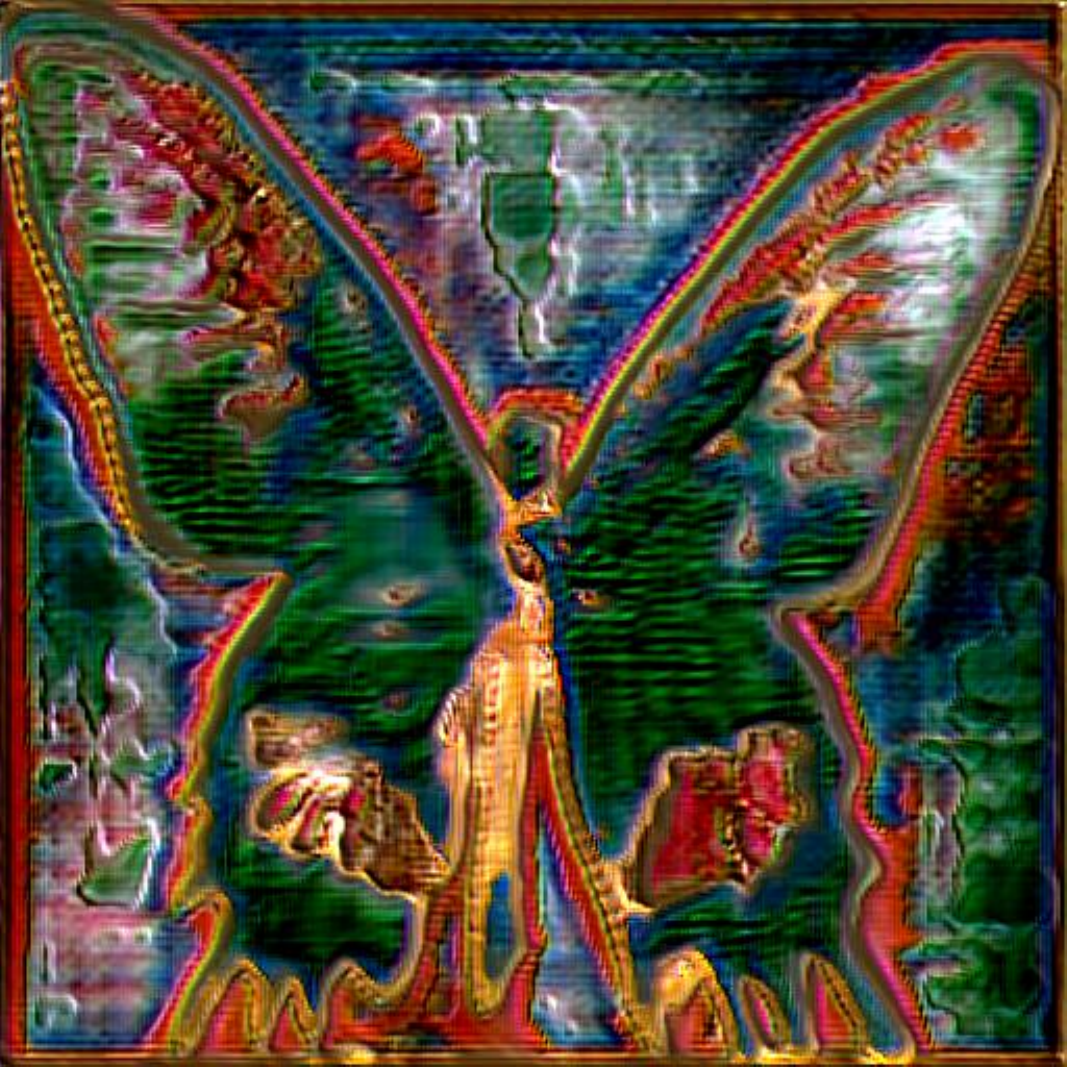}   
    \end{tabular}
\caption{First 16 outputs for the prompt "Psychedelic painting of a multi-armed deity" using the vanilla convolutional BM and standard CLIP loss.}
\label{fig:clip_tree}
\end{center}
\end{figure}

\begin{figure}[h] \
\begin{center}
\setlength{\tabcolsep}{2pt}
    \begin{tabular}{cccc}
      \includegraphics[width=0.24\linewidth]{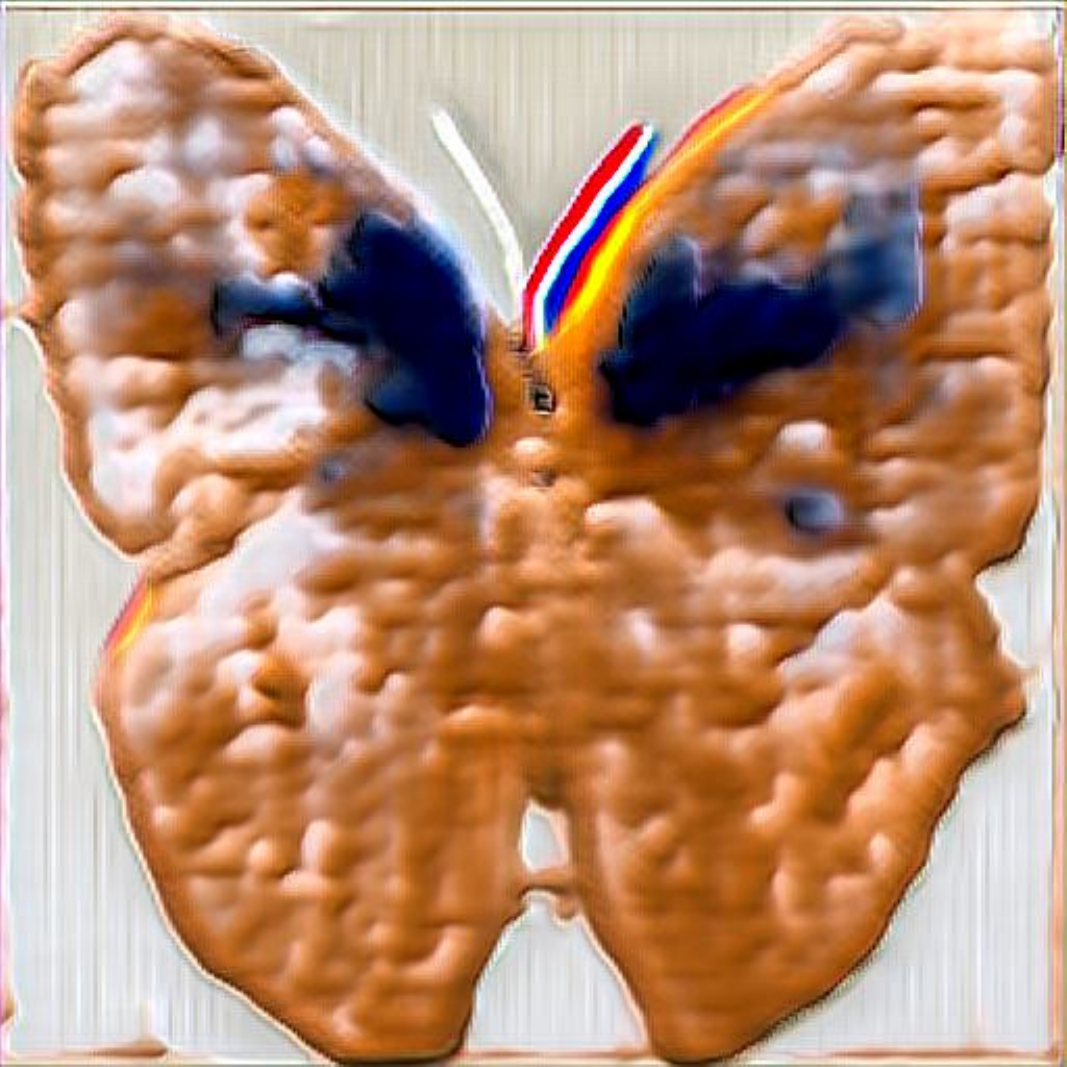}   &
      \includegraphics[width=0.24\linewidth]{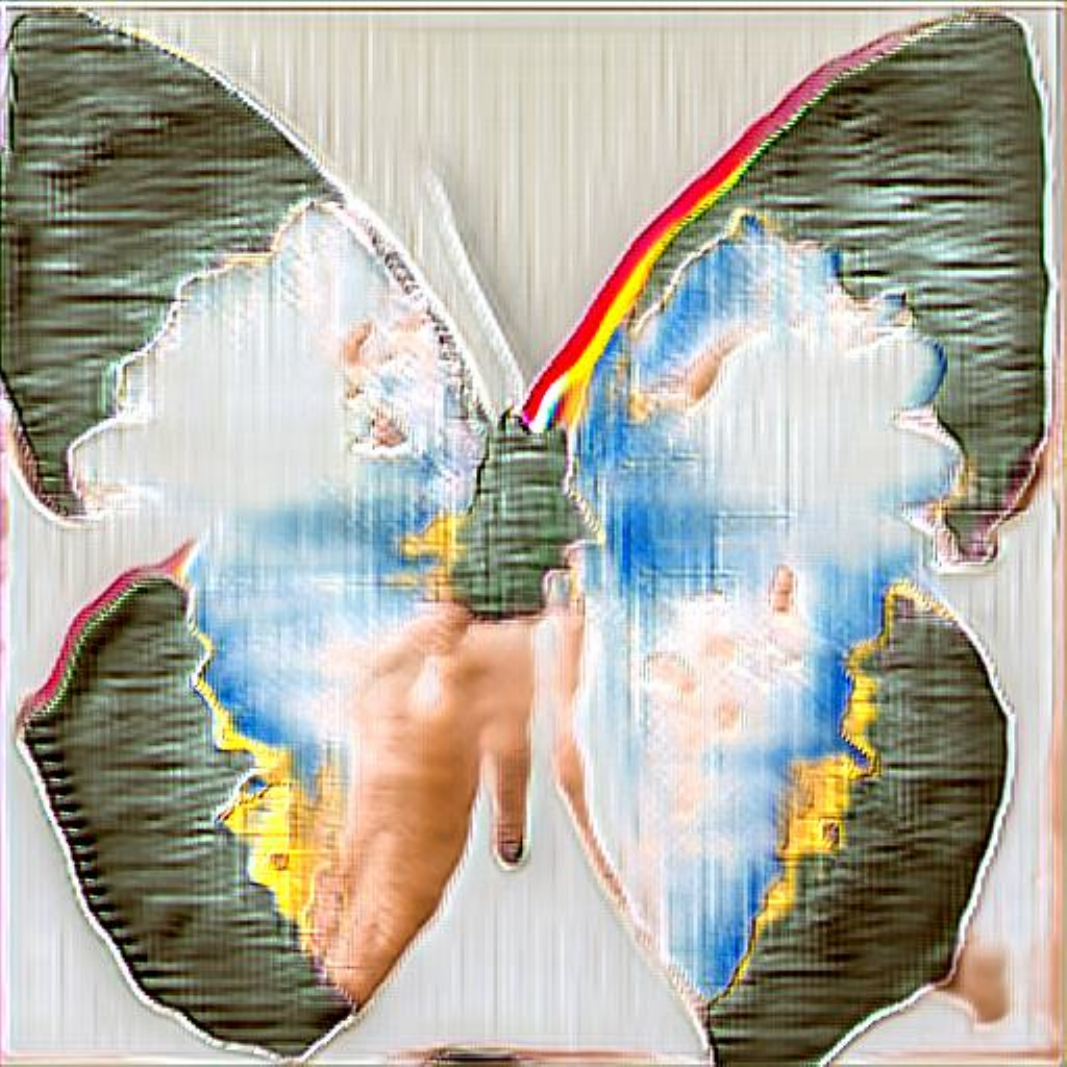}   &
      \includegraphics[width=0.24\linewidth]{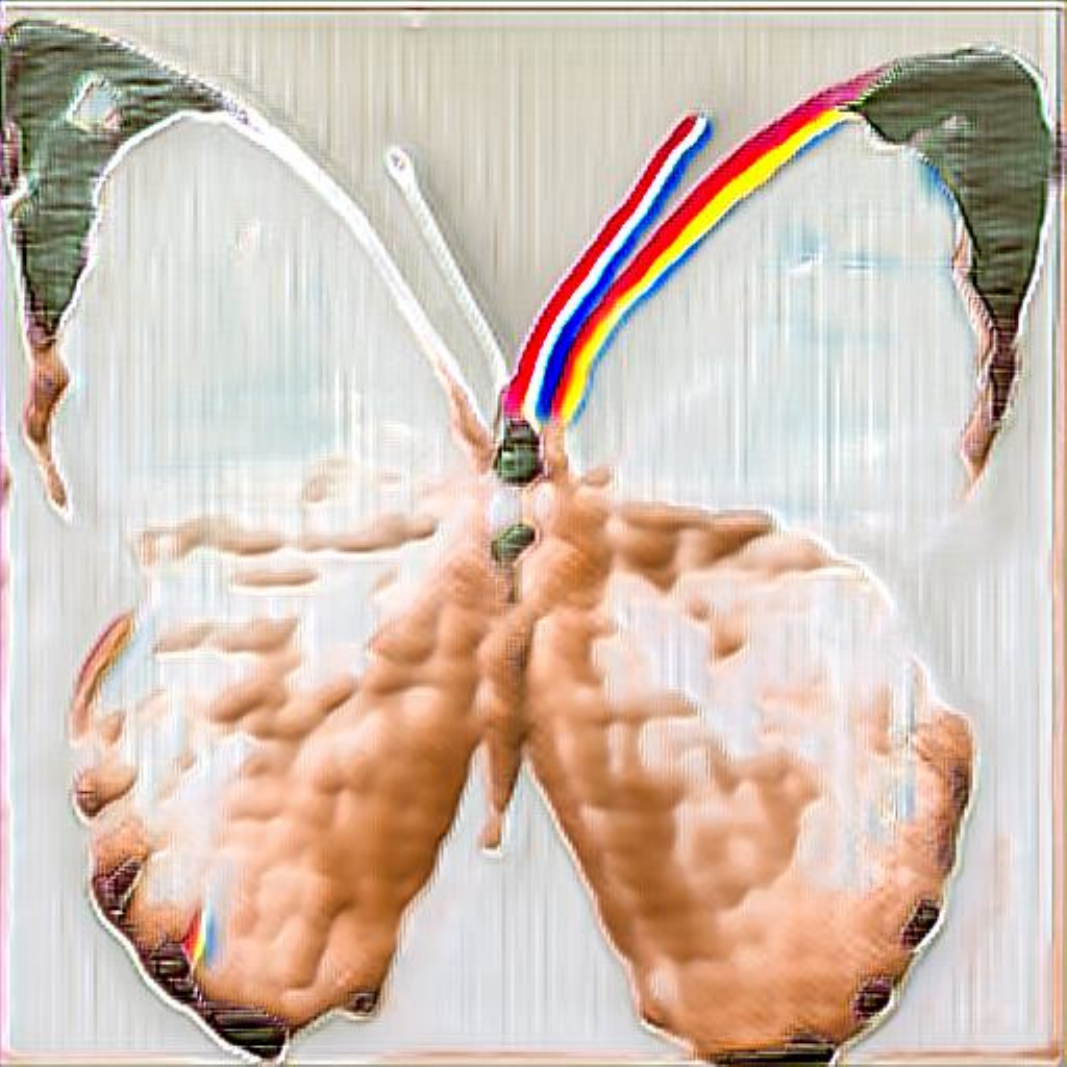}   &
      \includegraphics[width=0.24\linewidth]{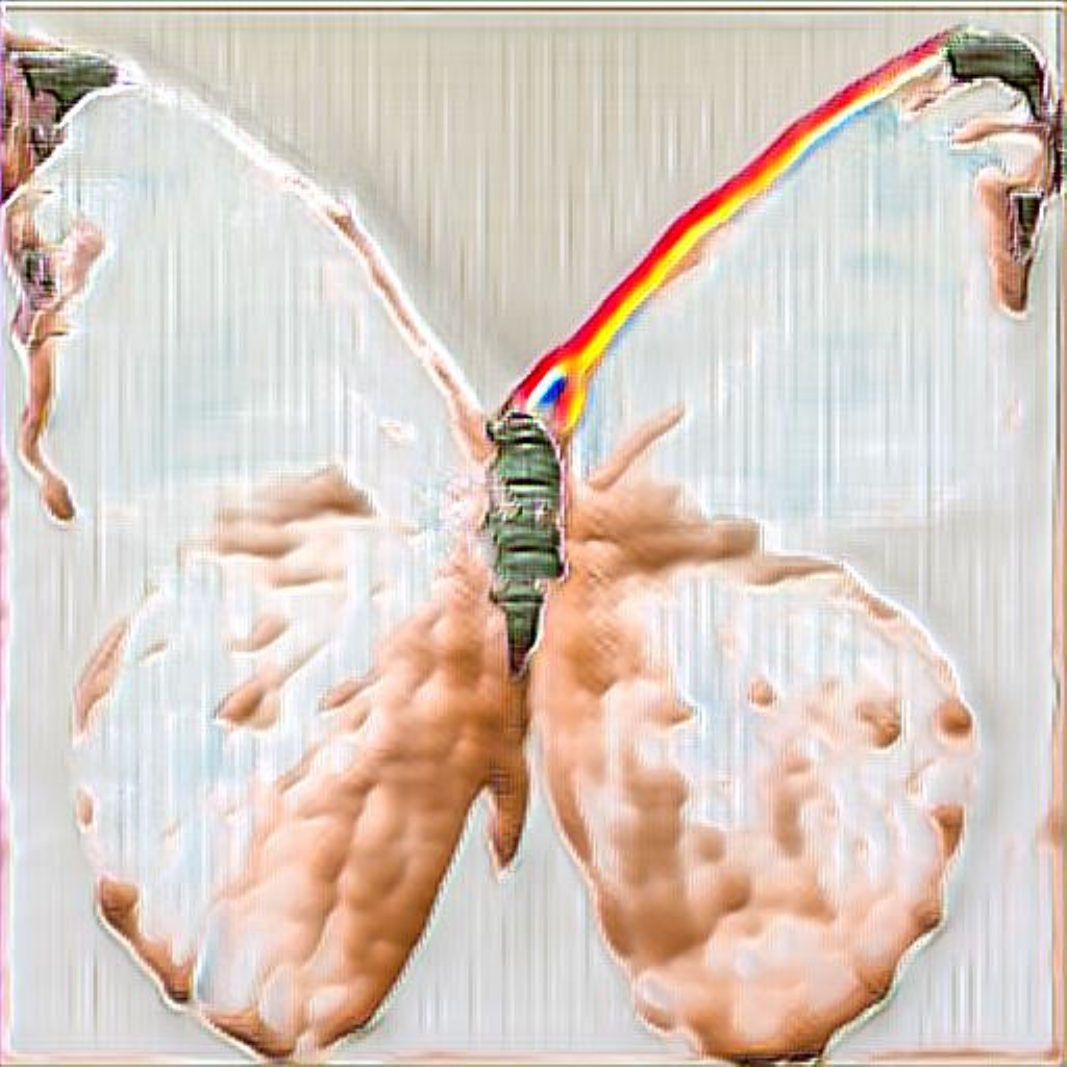}   \\
      \includegraphics[width=0.24\linewidth]{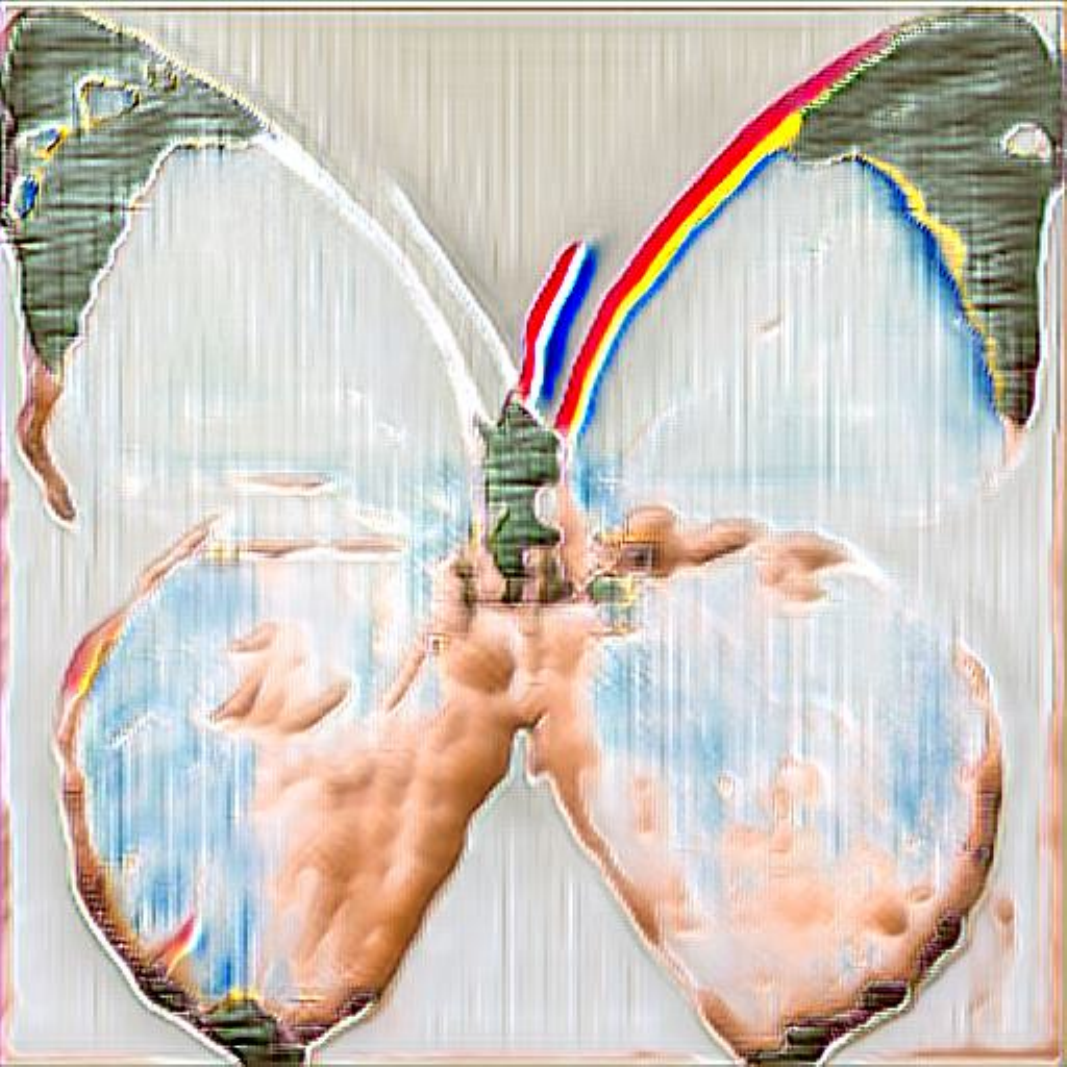}   &
      \includegraphics[width=0.24\linewidth]{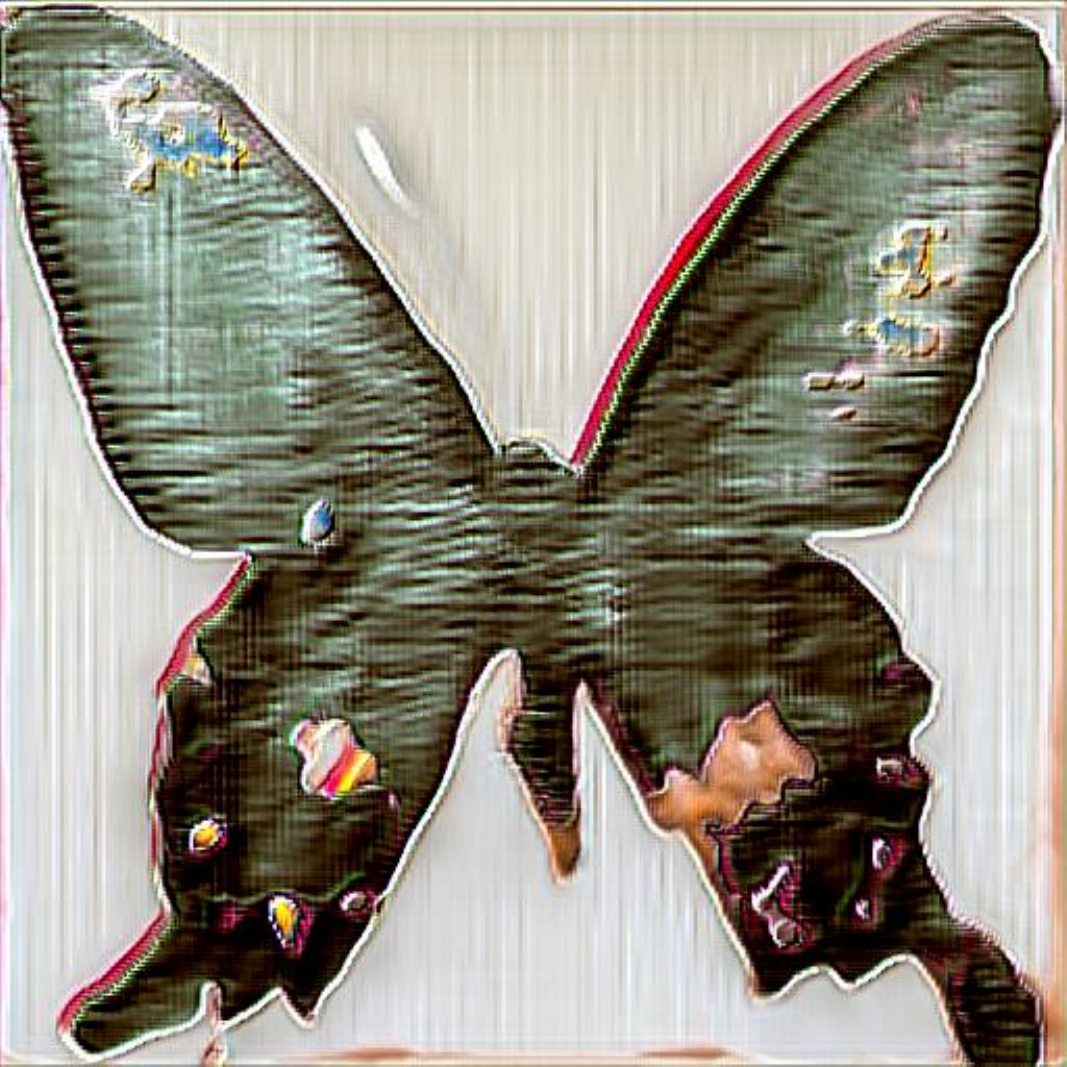}   &
      \includegraphics[width=0.24\linewidth]{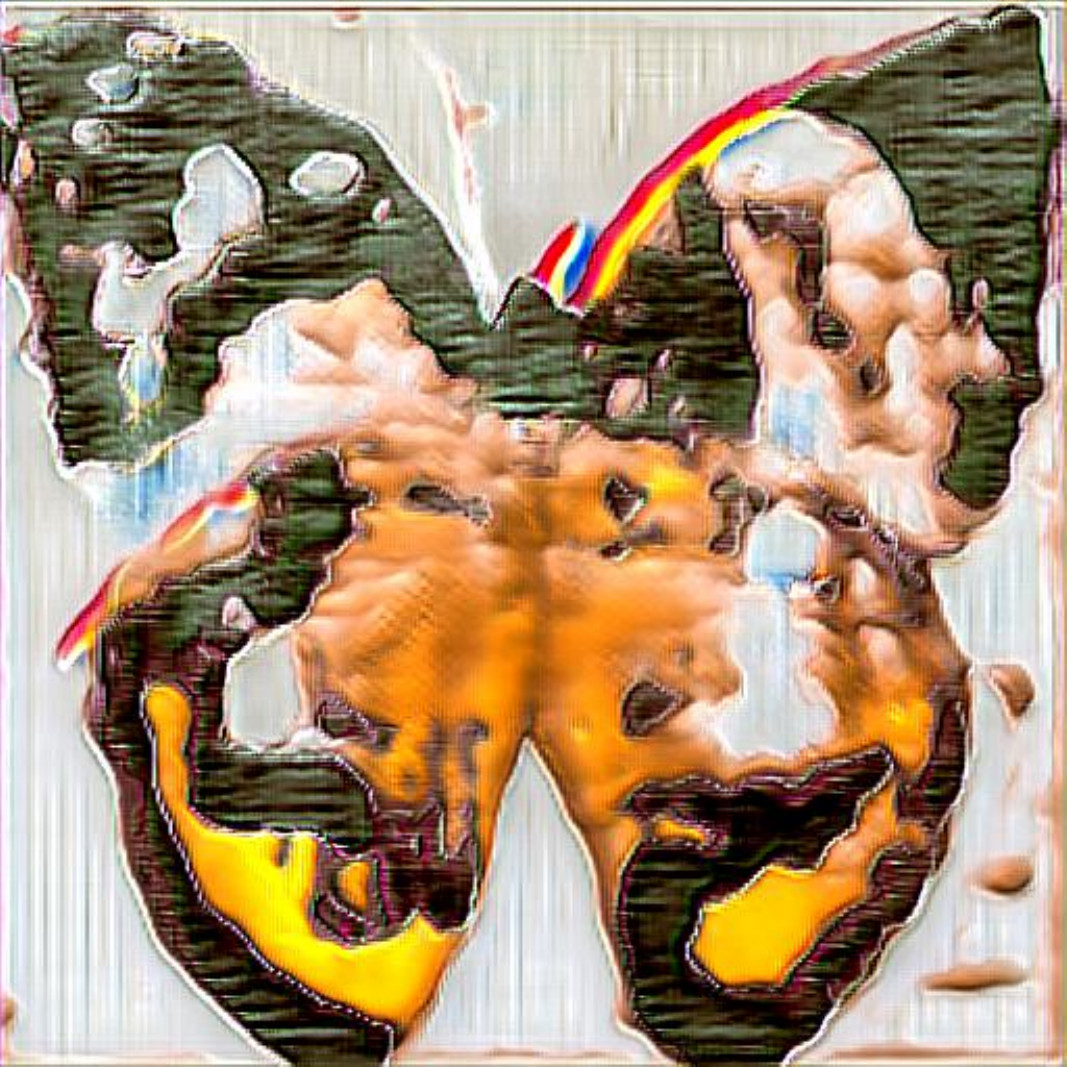}   &
      \includegraphics[width=0.24\linewidth]{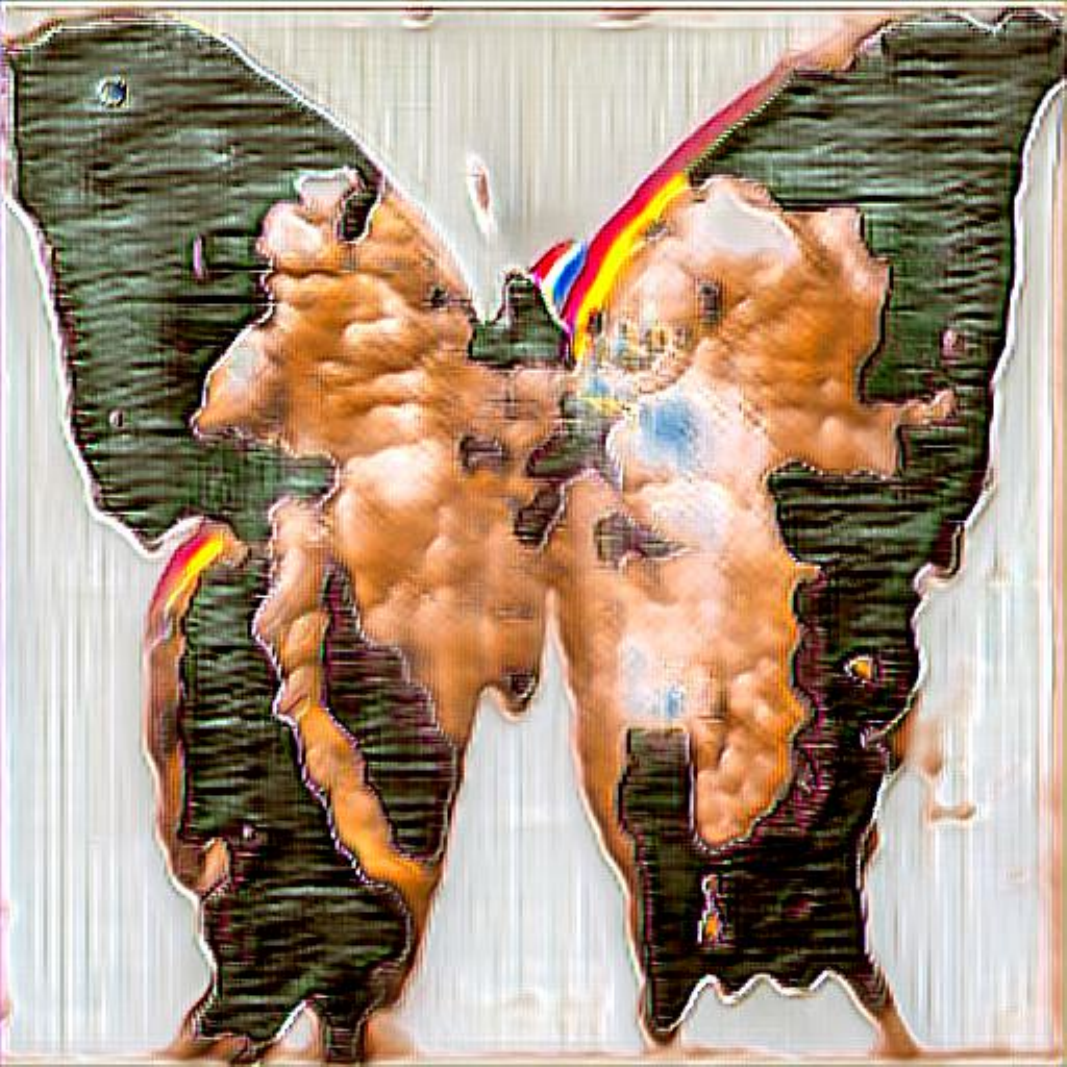}   \\
      \includegraphics[width=0.24\linewidth]{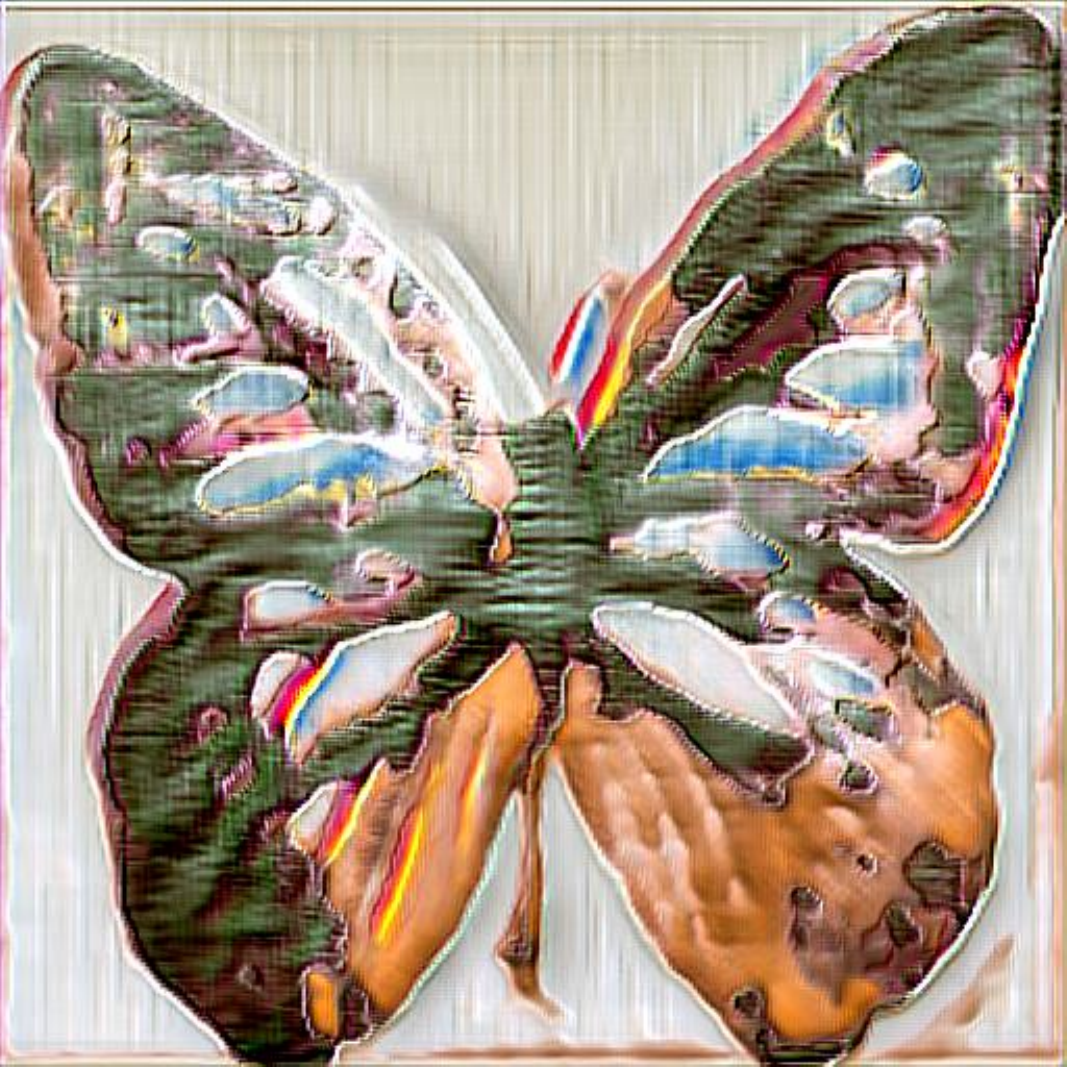}   &
      \includegraphics[width=0.24\linewidth]{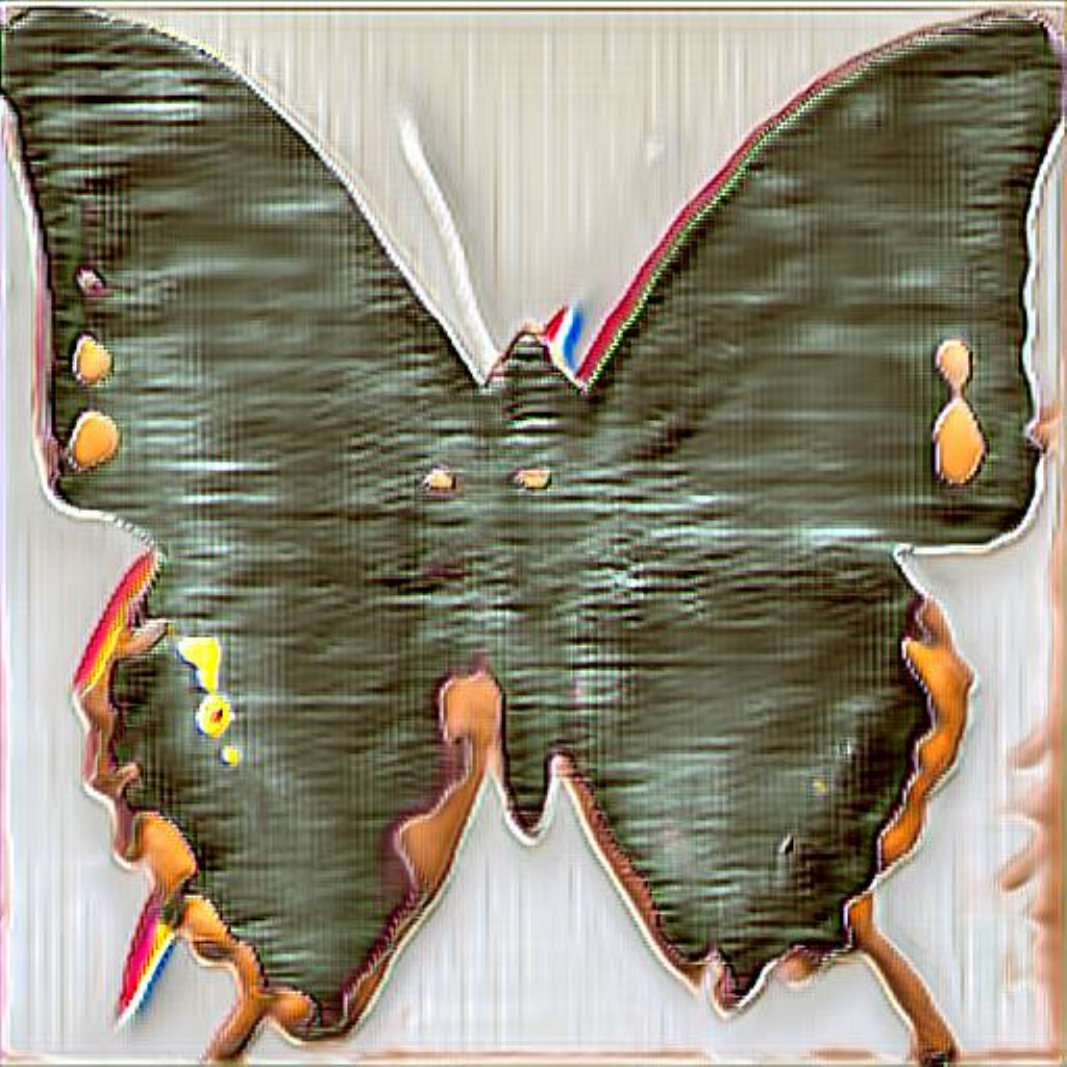}   &
      \includegraphics[width=0.24\linewidth]{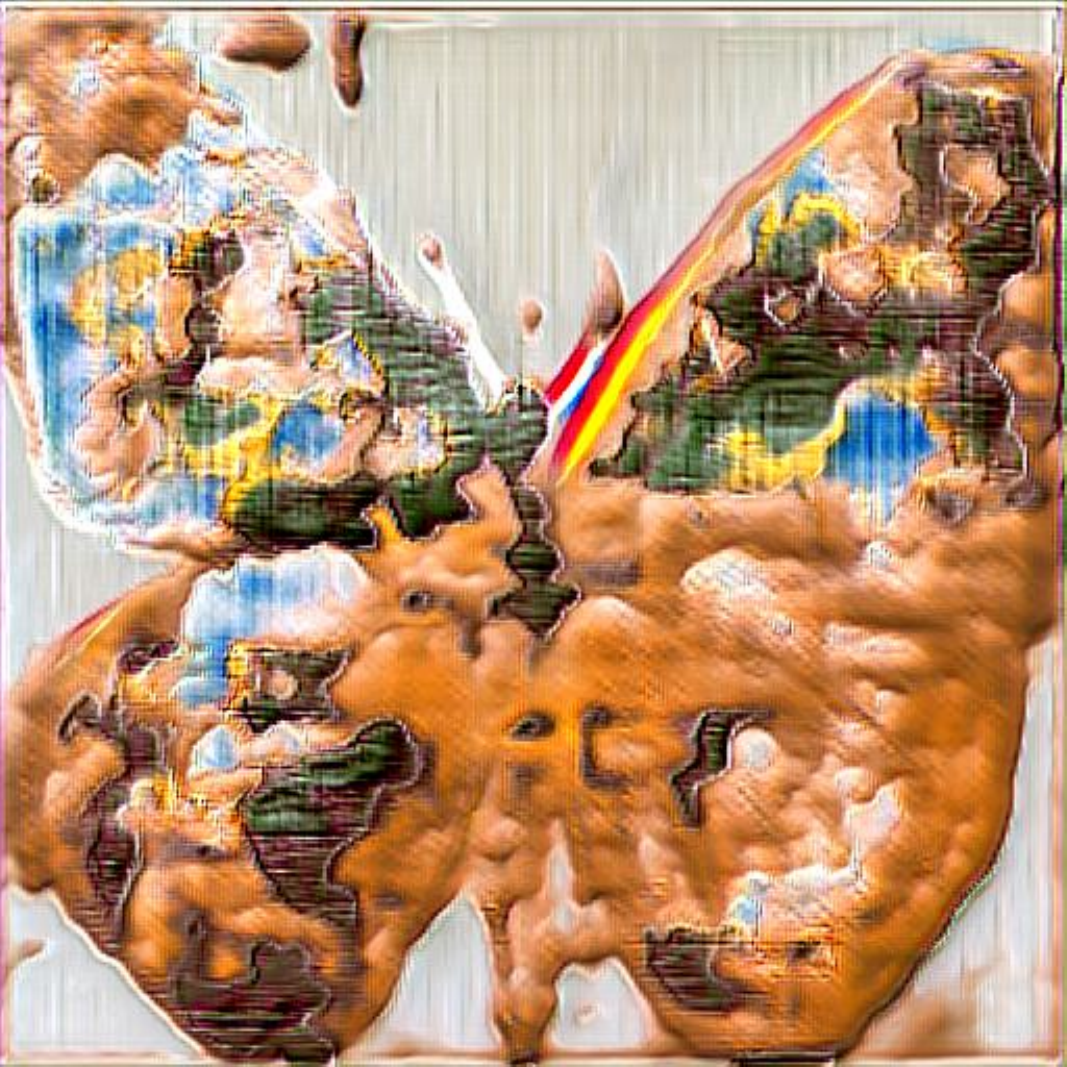}   &
      \includegraphics[width=0.24\linewidth]{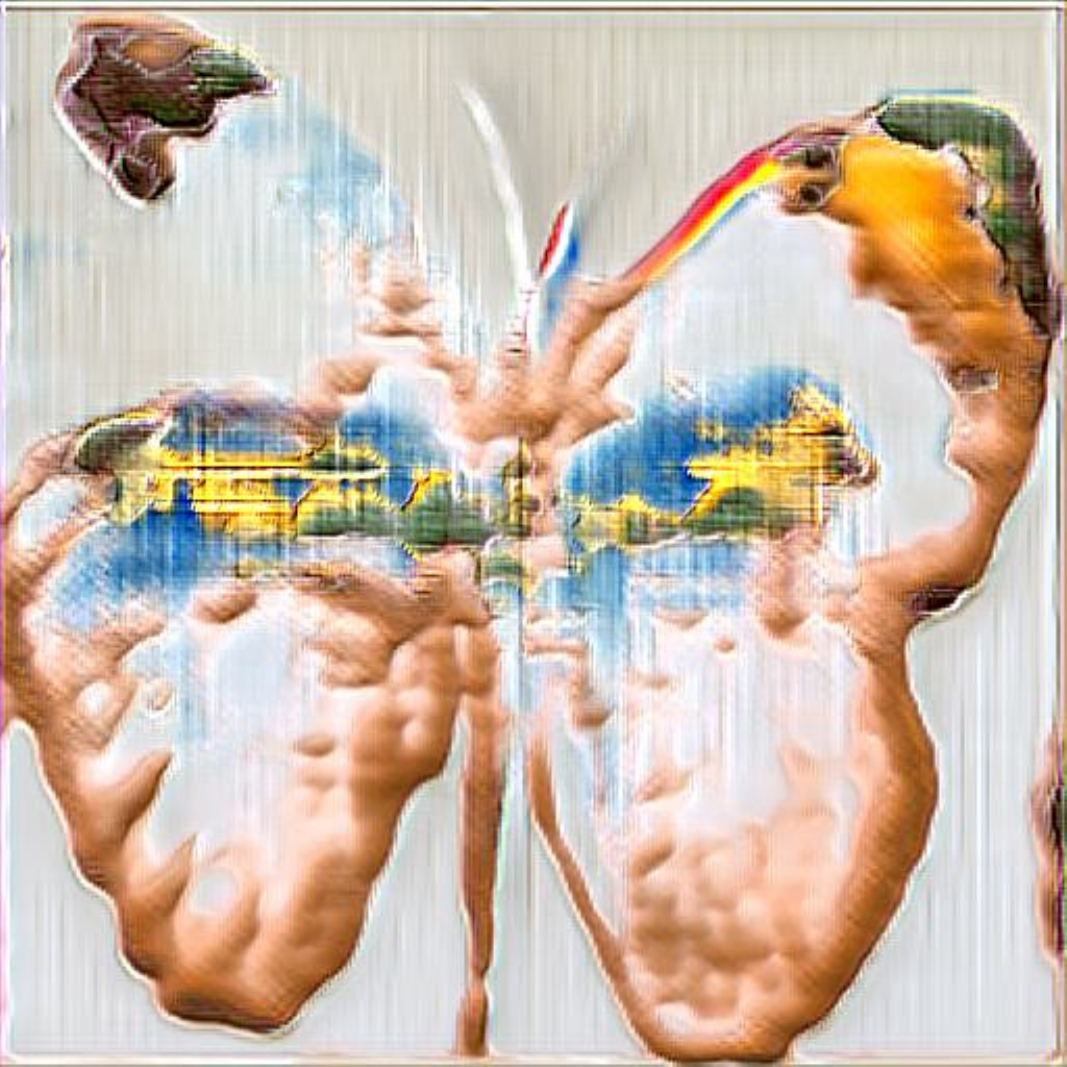}   \\
      \includegraphics[width=0.24\linewidth]{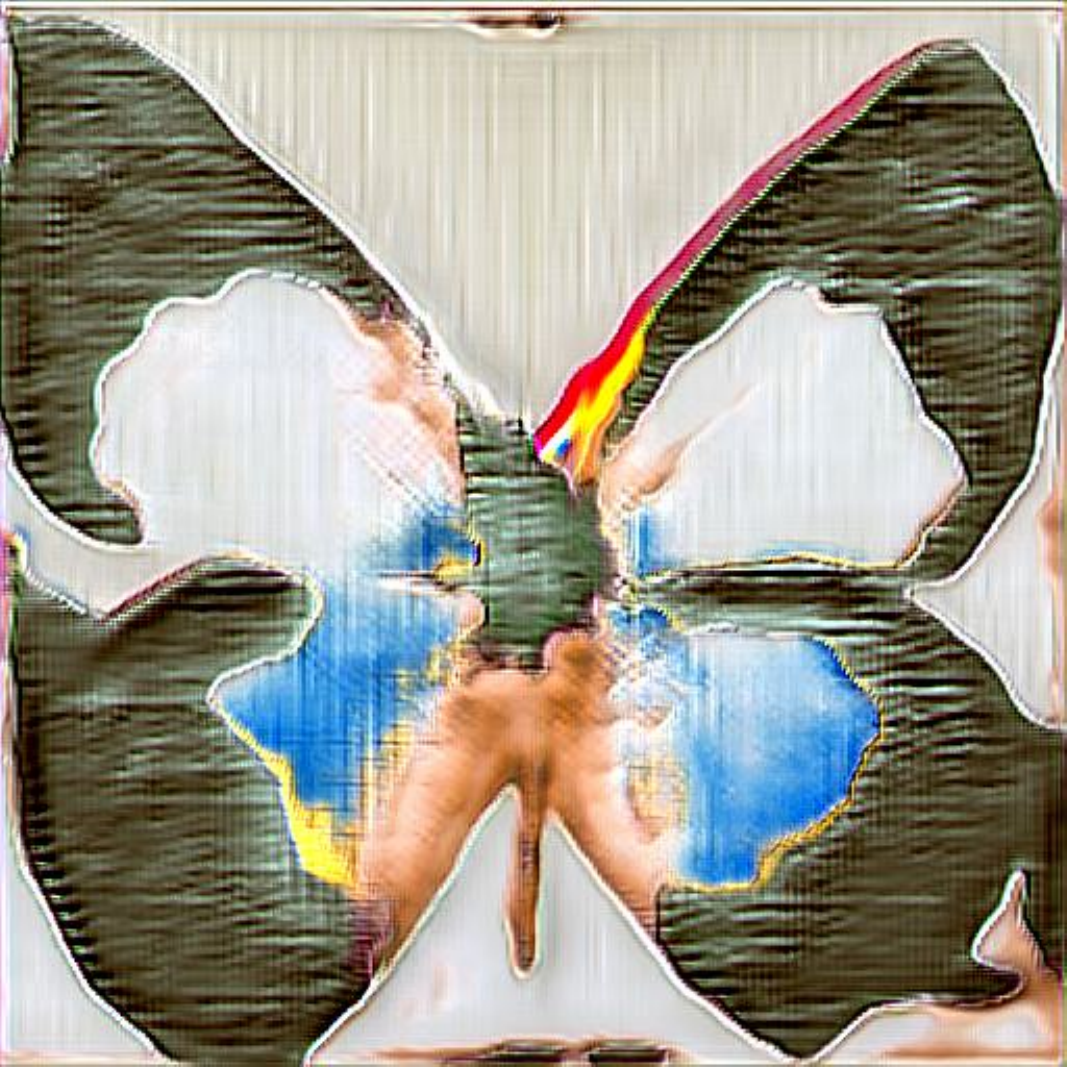}   &
      \includegraphics[width=0.24\linewidth]{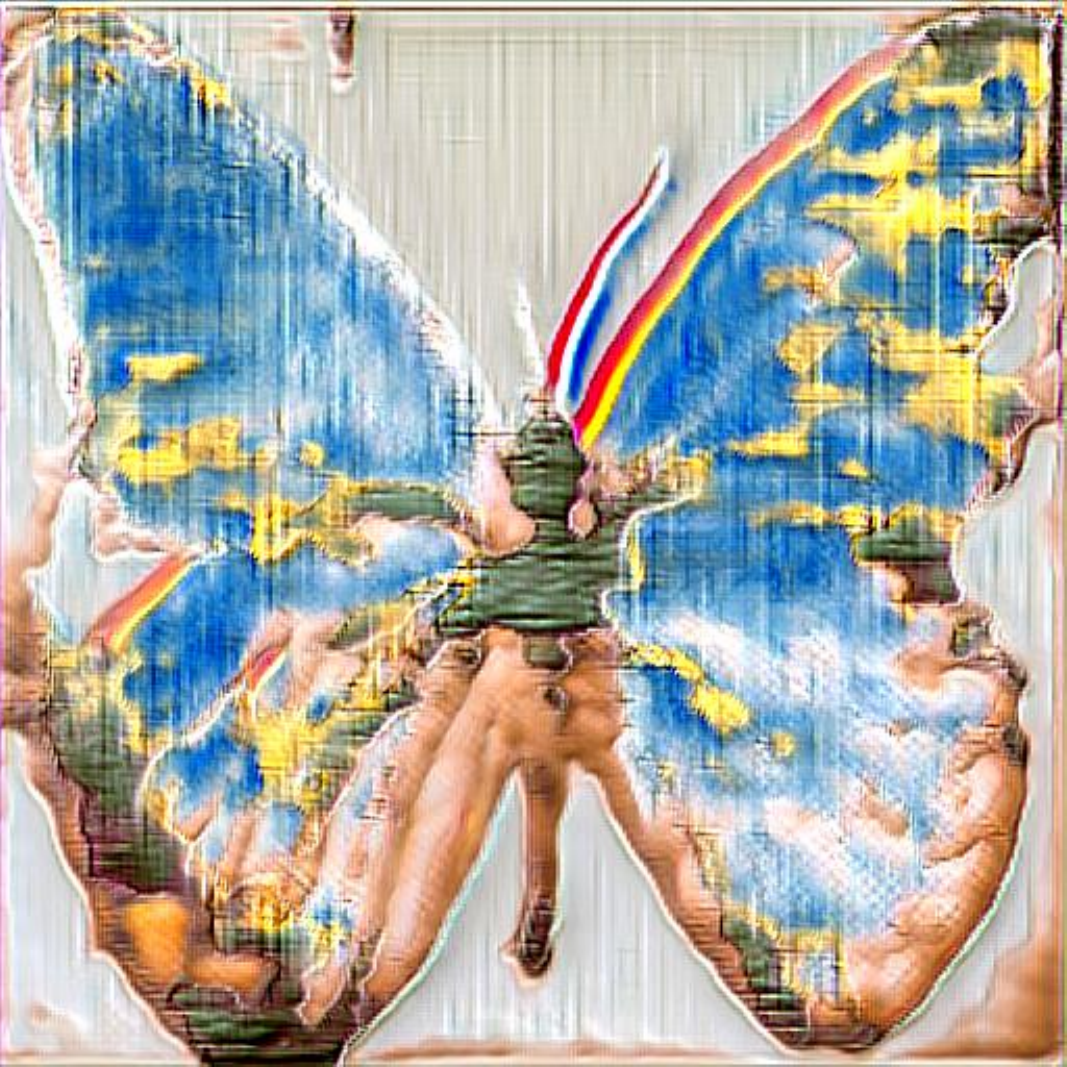}   &
      \includegraphics[width=0.24\linewidth]{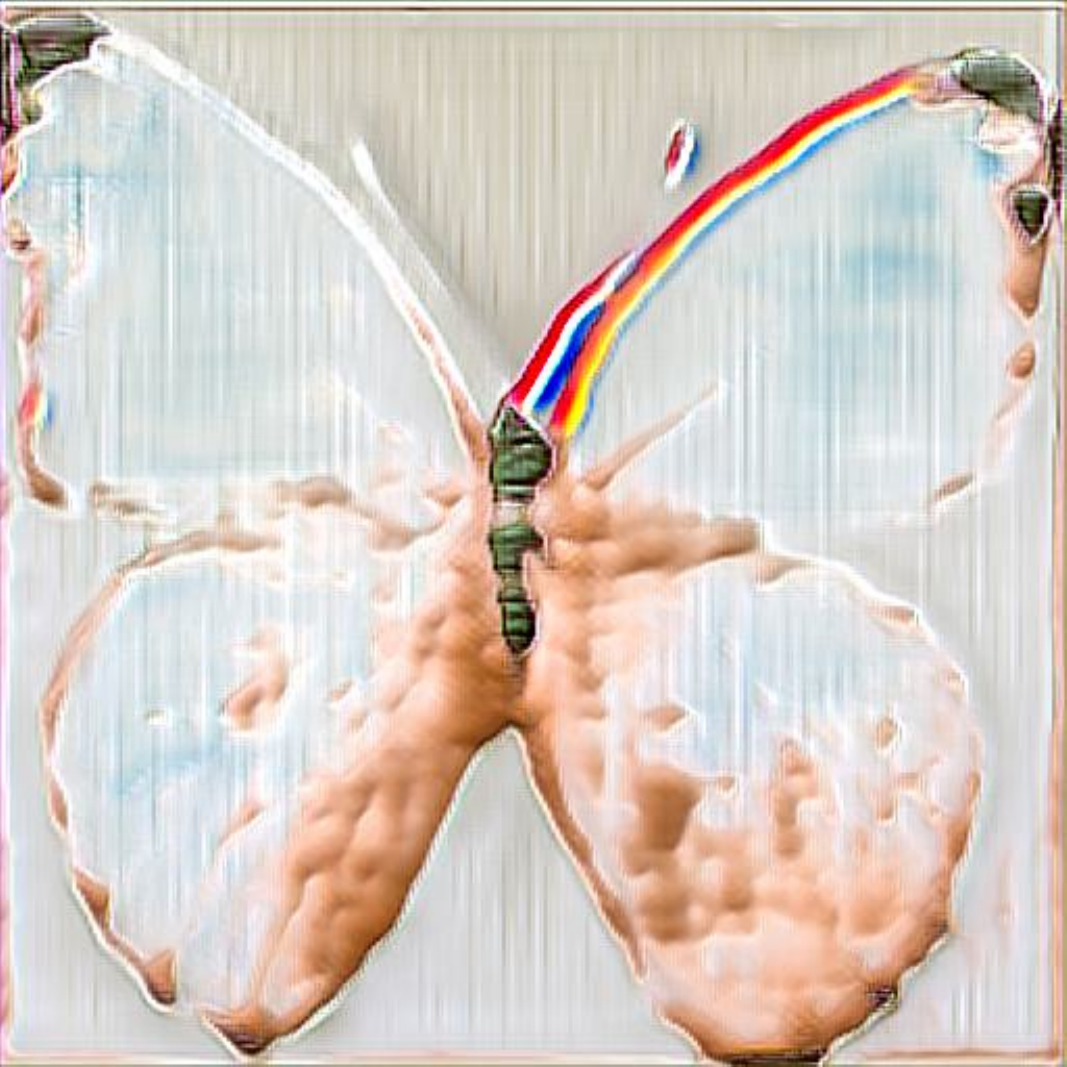}   &
      \includegraphics[width=0.24\linewidth]{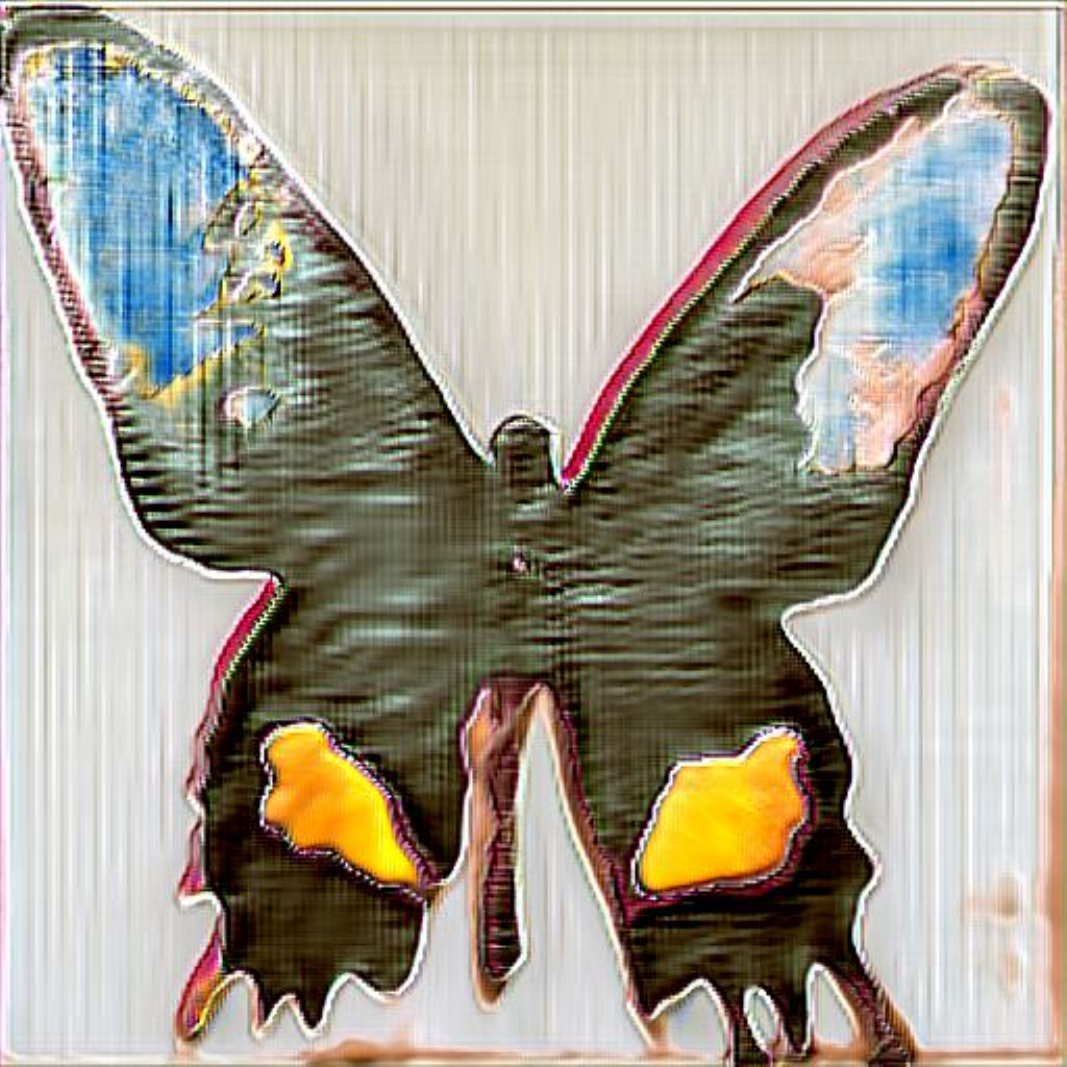}   
    \end{tabular}
\caption{First 16 outputs for the prompt "Inflatable plastic bodybuilder in a colorful album cover painted by Magritte" using the coordinates-aware BM and InfoNCE loss.}
\label{fig:clip_tree}
\end{center}
\end{figure}

\begin{figure}[h!] \
\begin{center}
\setlength{\tabcolsep}{2pt}
    \begin{tabular}{cccc}
      \includegraphics[width=0.24\linewidth]{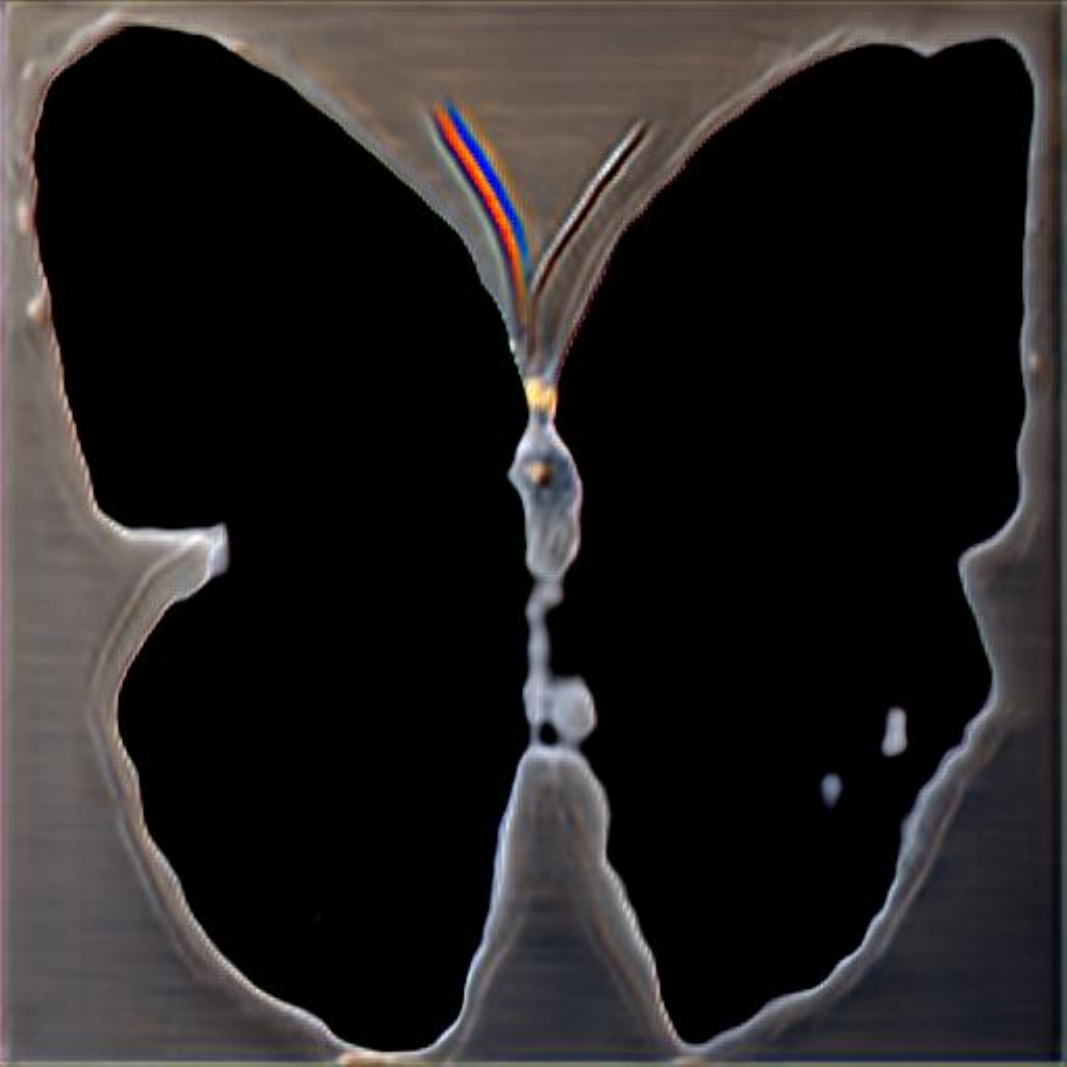}   &
      \includegraphics[width=0.24\linewidth]{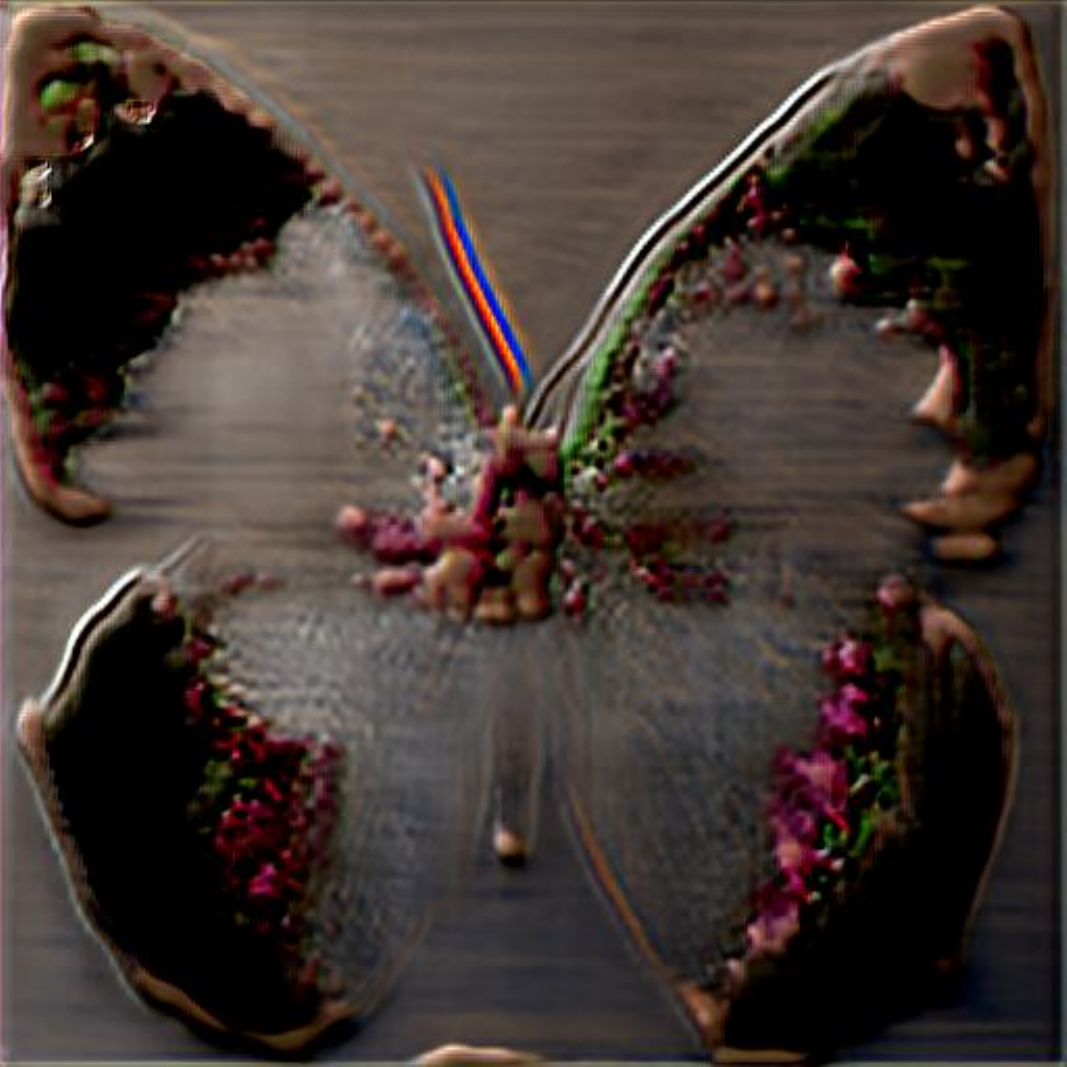}   &
      \includegraphics[width=0.24\linewidth]{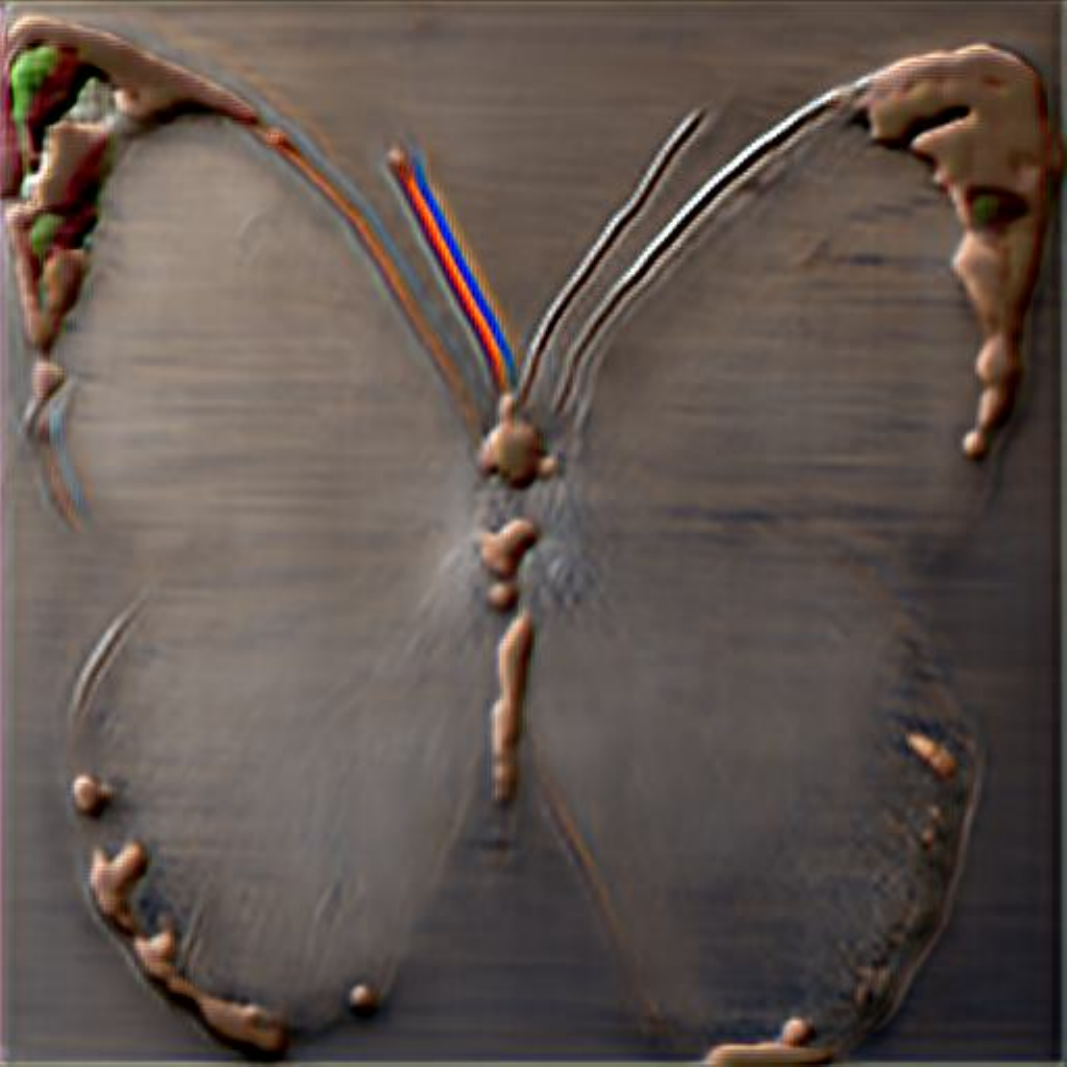}   &
      \includegraphics[width=0.24\linewidth]{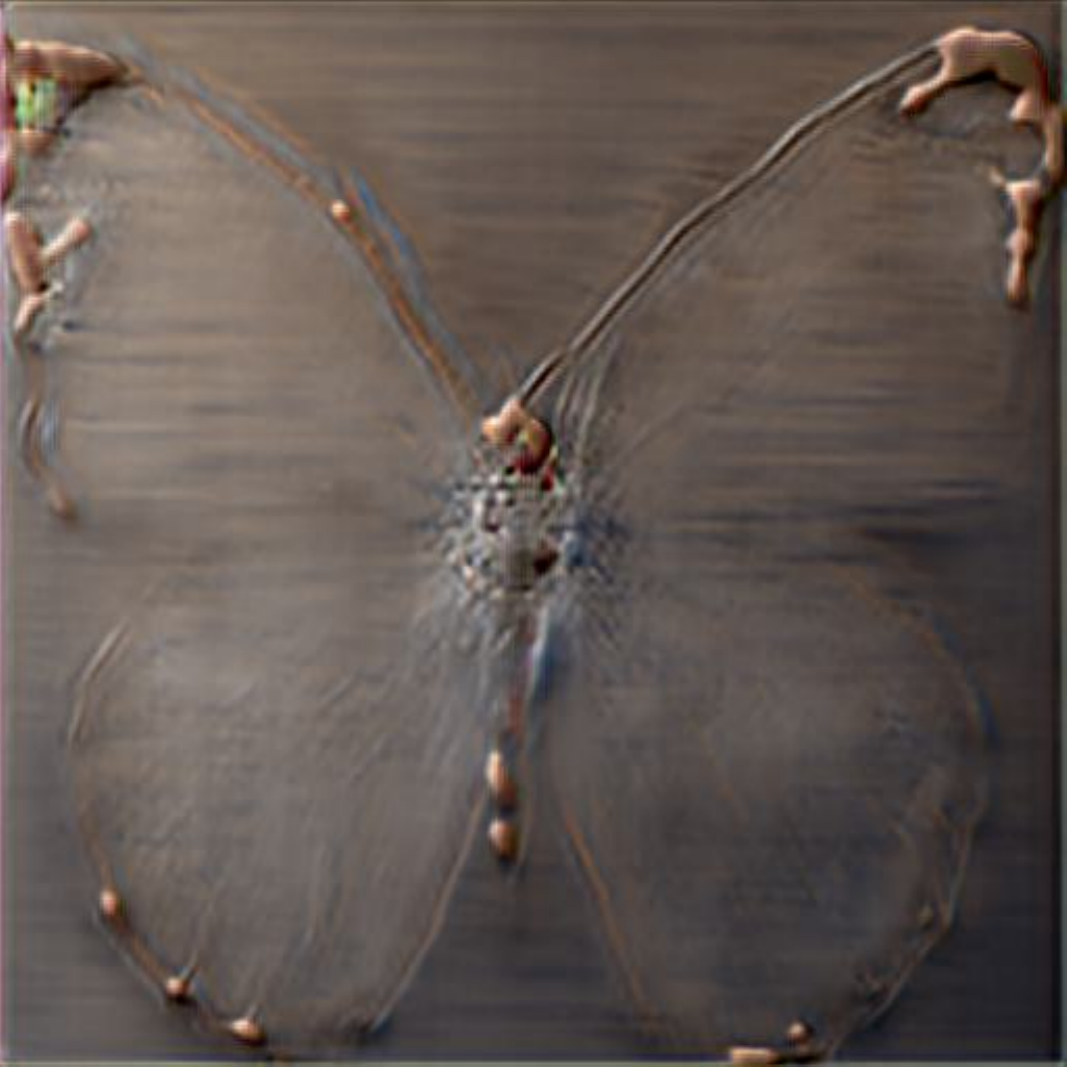}   \\
      \includegraphics[width=0.24\linewidth]{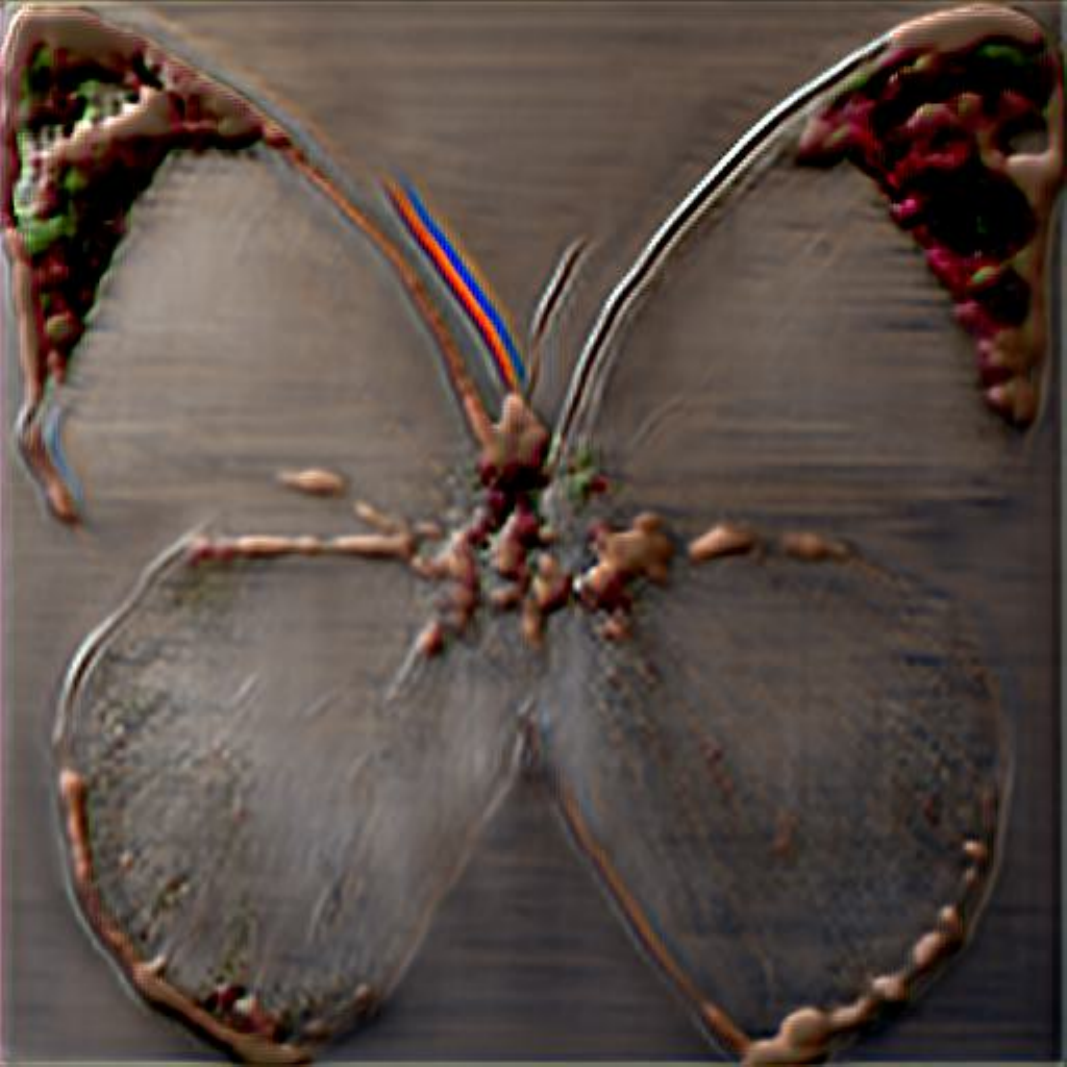}   &
      \includegraphics[width=0.24\linewidth]{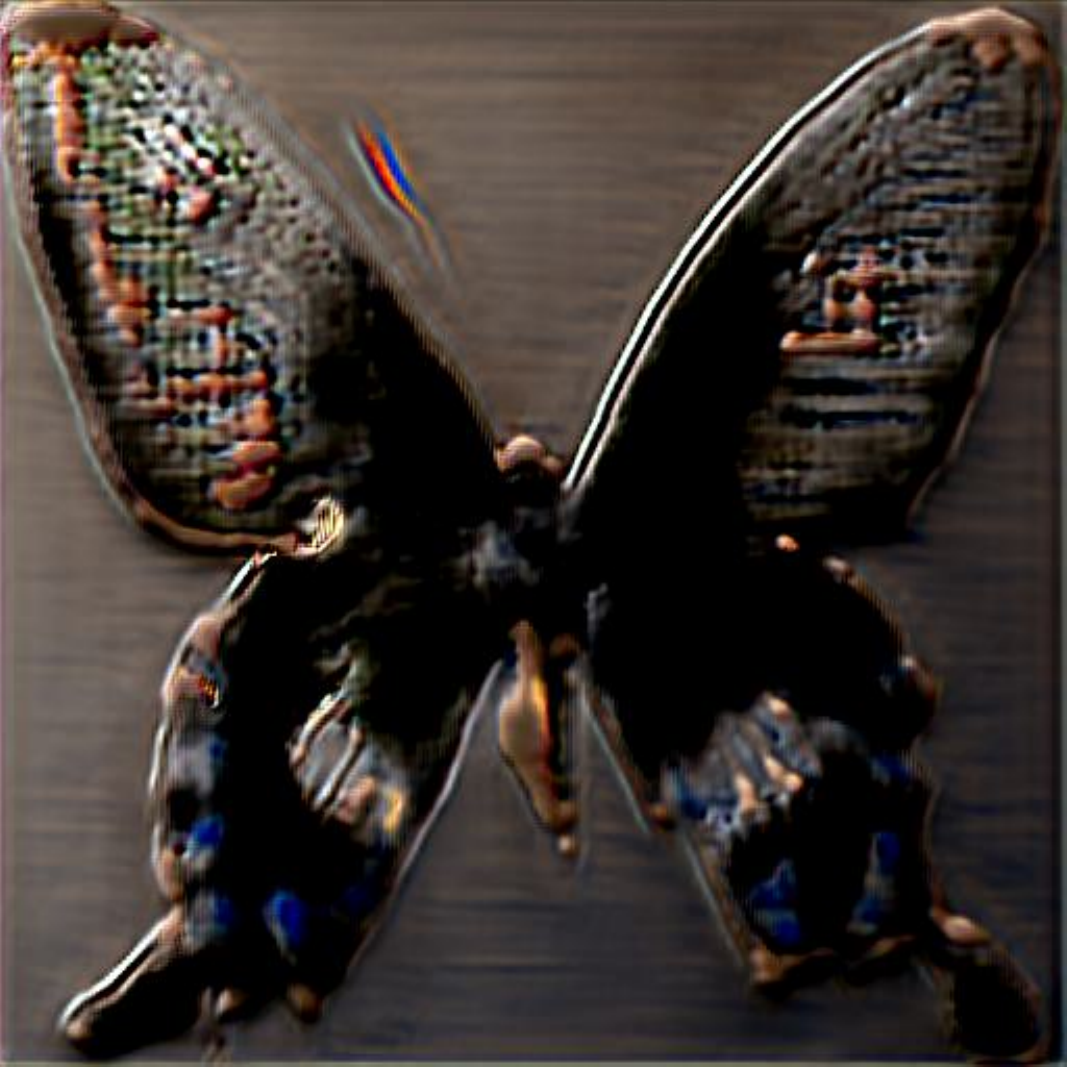}   &
      \includegraphics[width=0.24\linewidth]{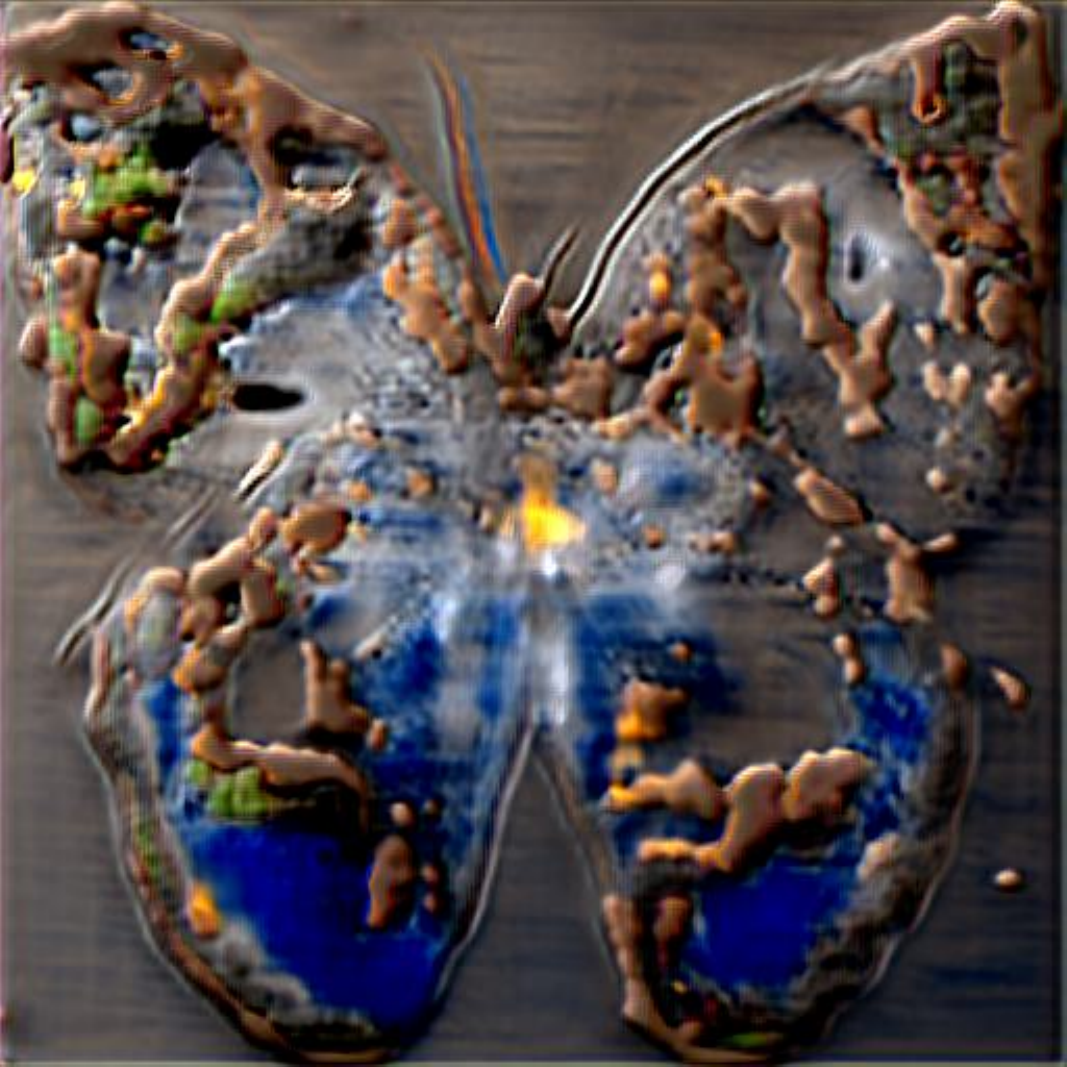}   &
      \includegraphics[width=0.24\linewidth]{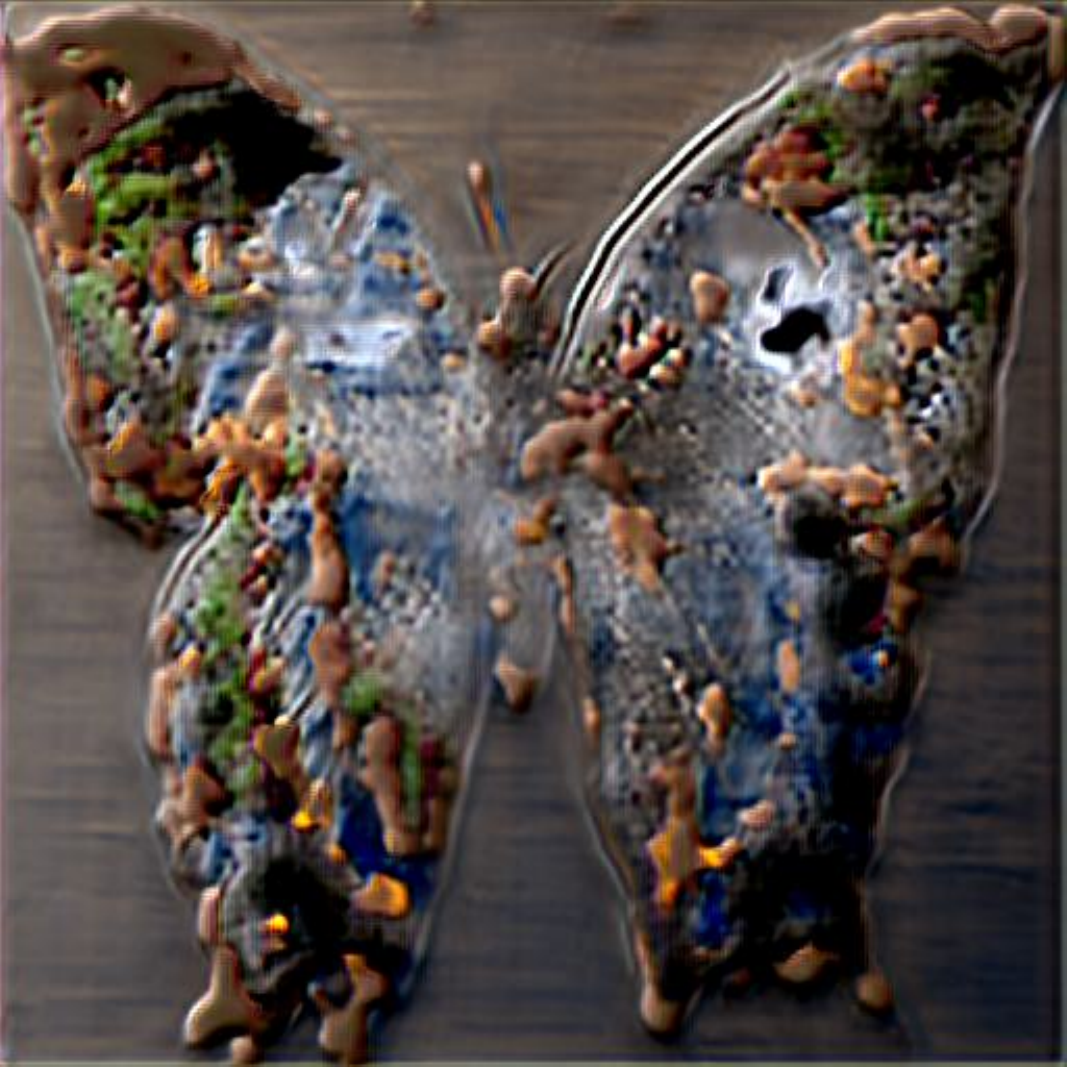}   \\
      \includegraphics[width=0.24\linewidth]{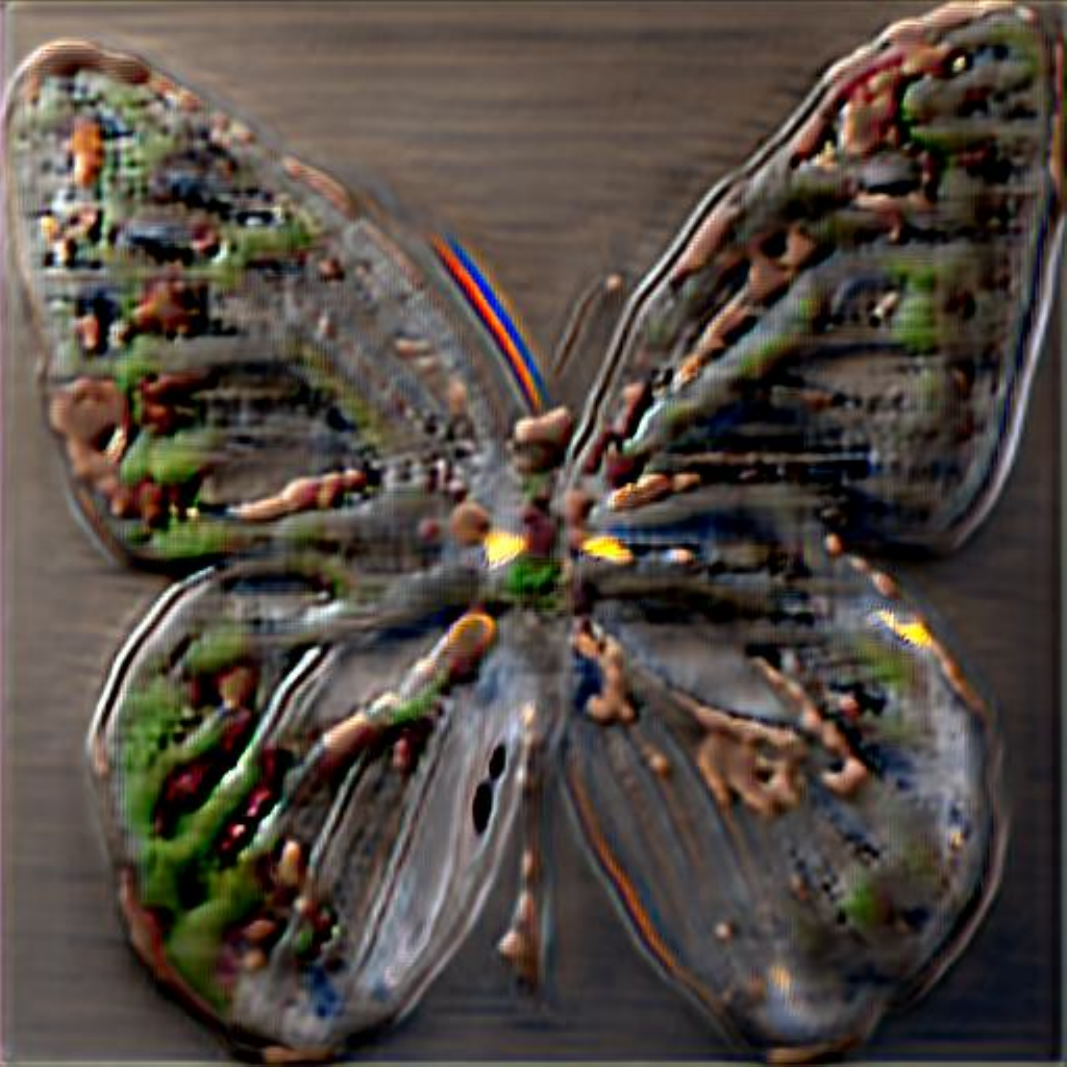}   &
      \includegraphics[width=0.24\linewidth]{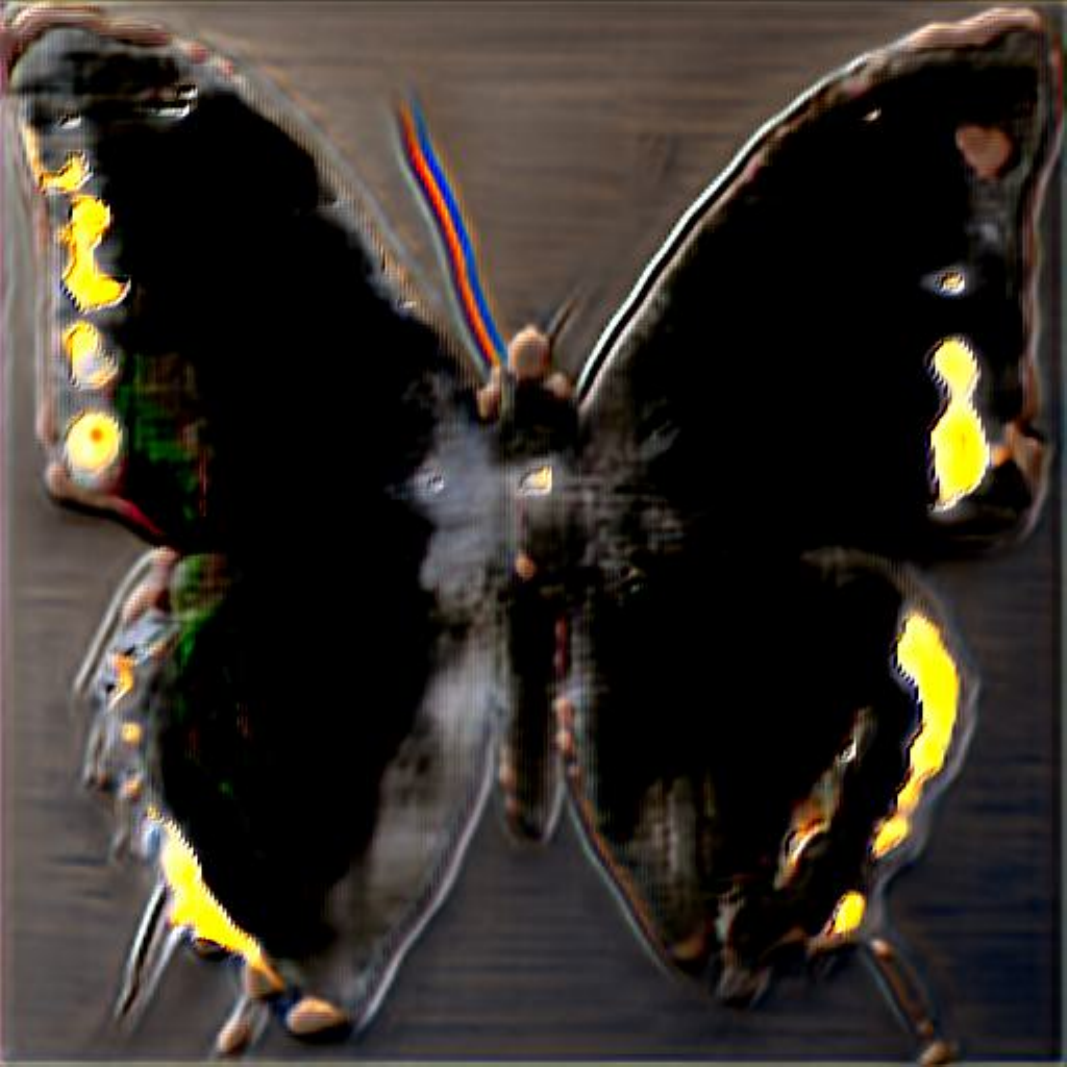}   &
      \includegraphics[width=0.24\linewidth]{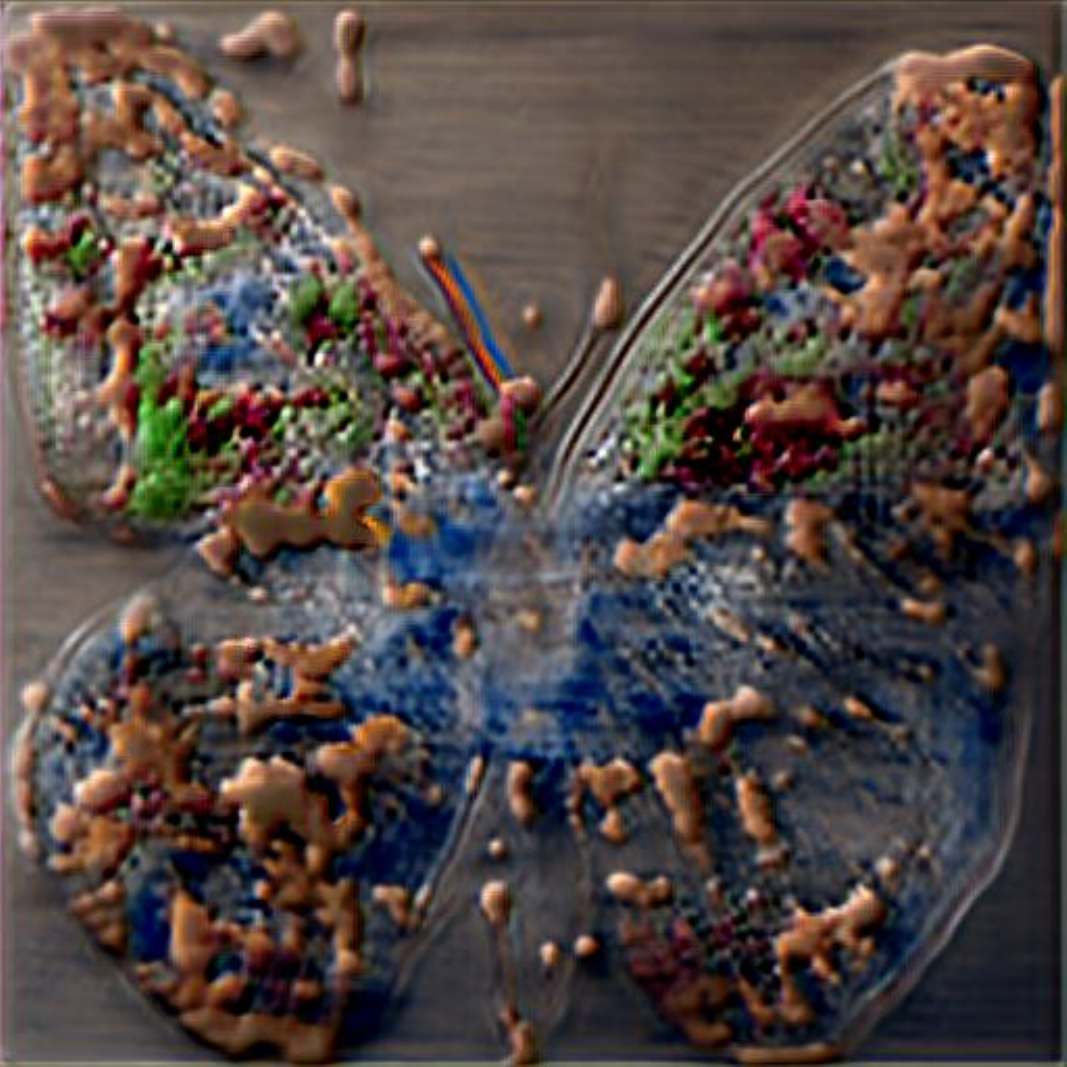}   &
      \includegraphics[width=0.24\linewidth]{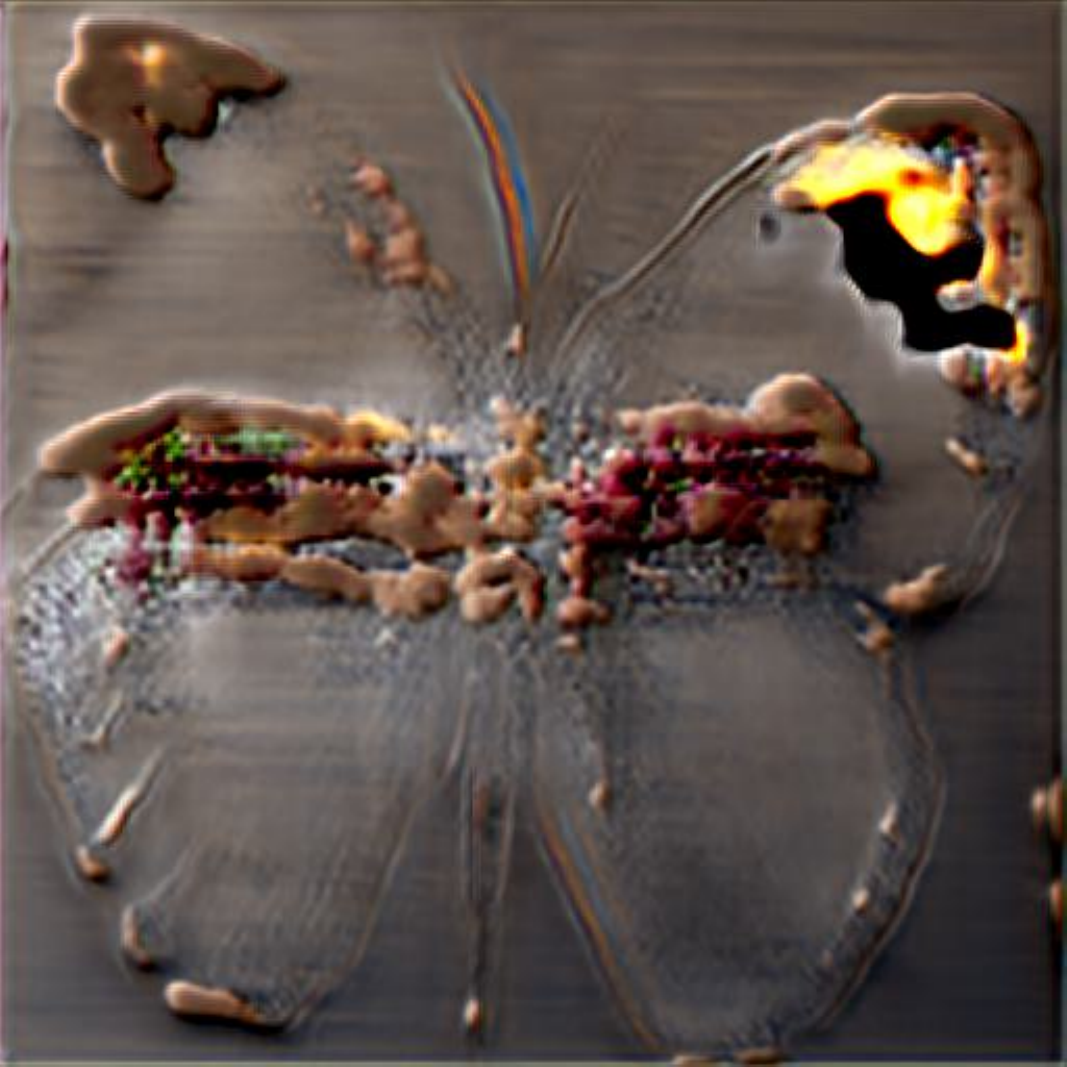}   \\
      \includegraphics[width=0.24\linewidth]{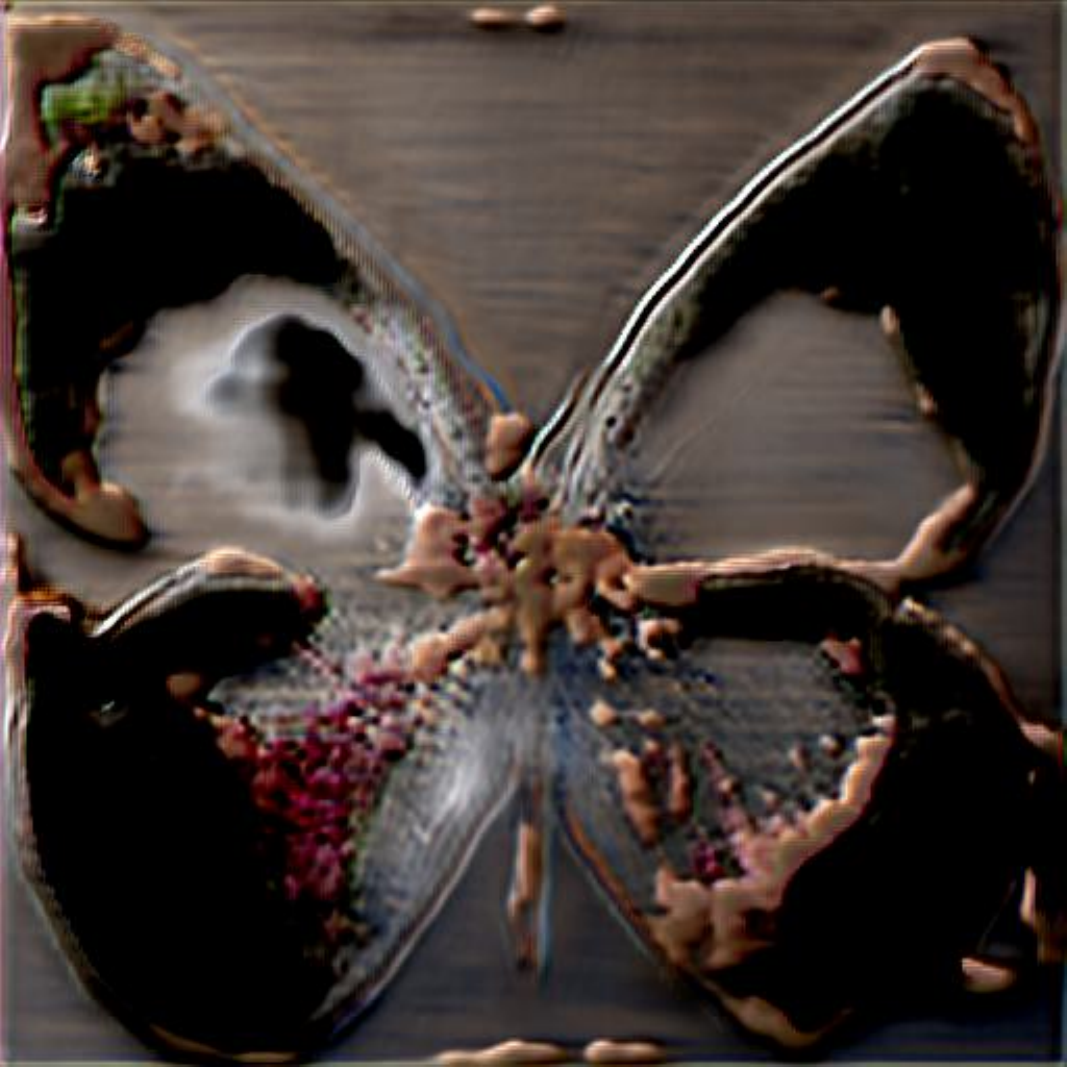}   &
      \includegraphics[width=0.24\linewidth]{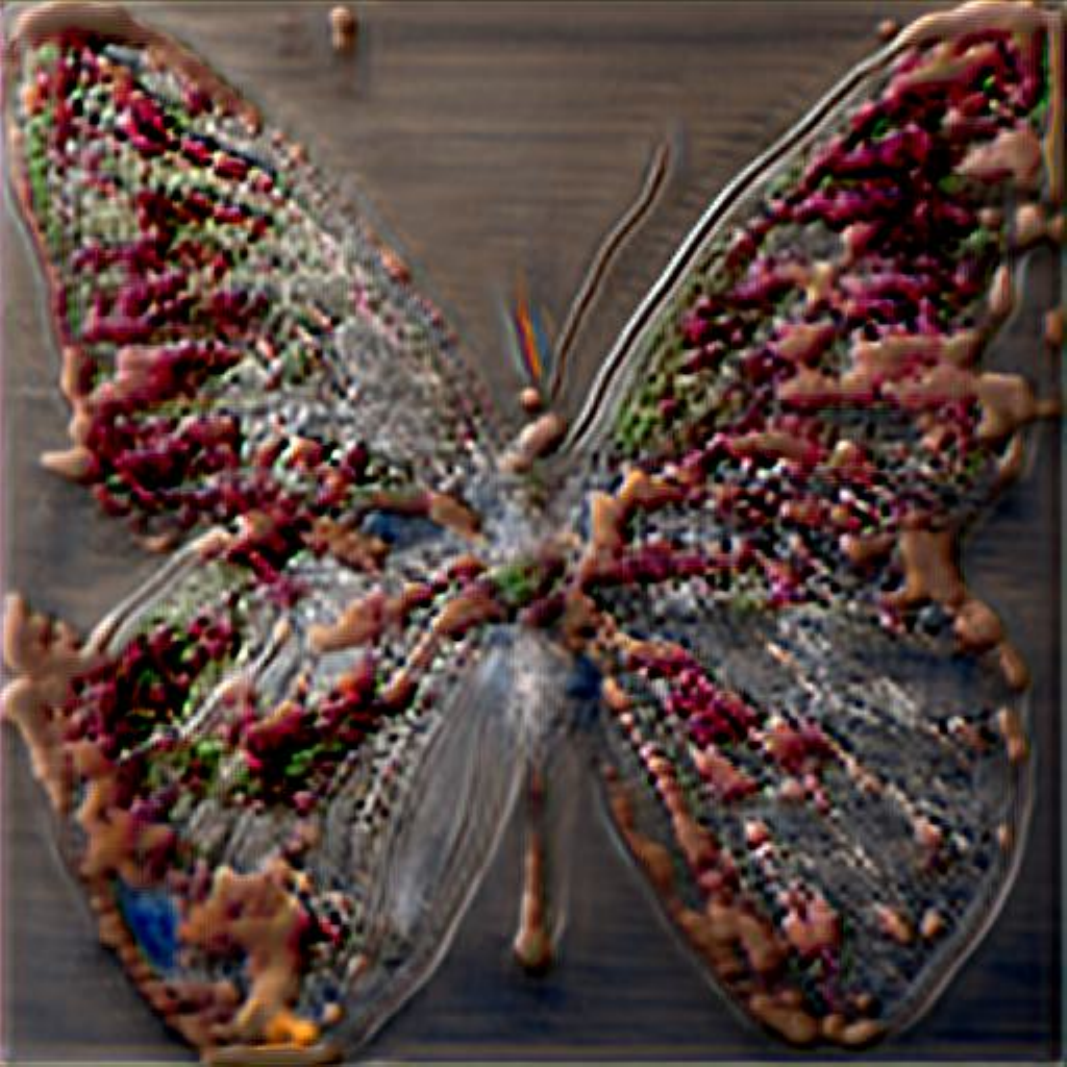}   &
      \includegraphics[width=0.24\linewidth]{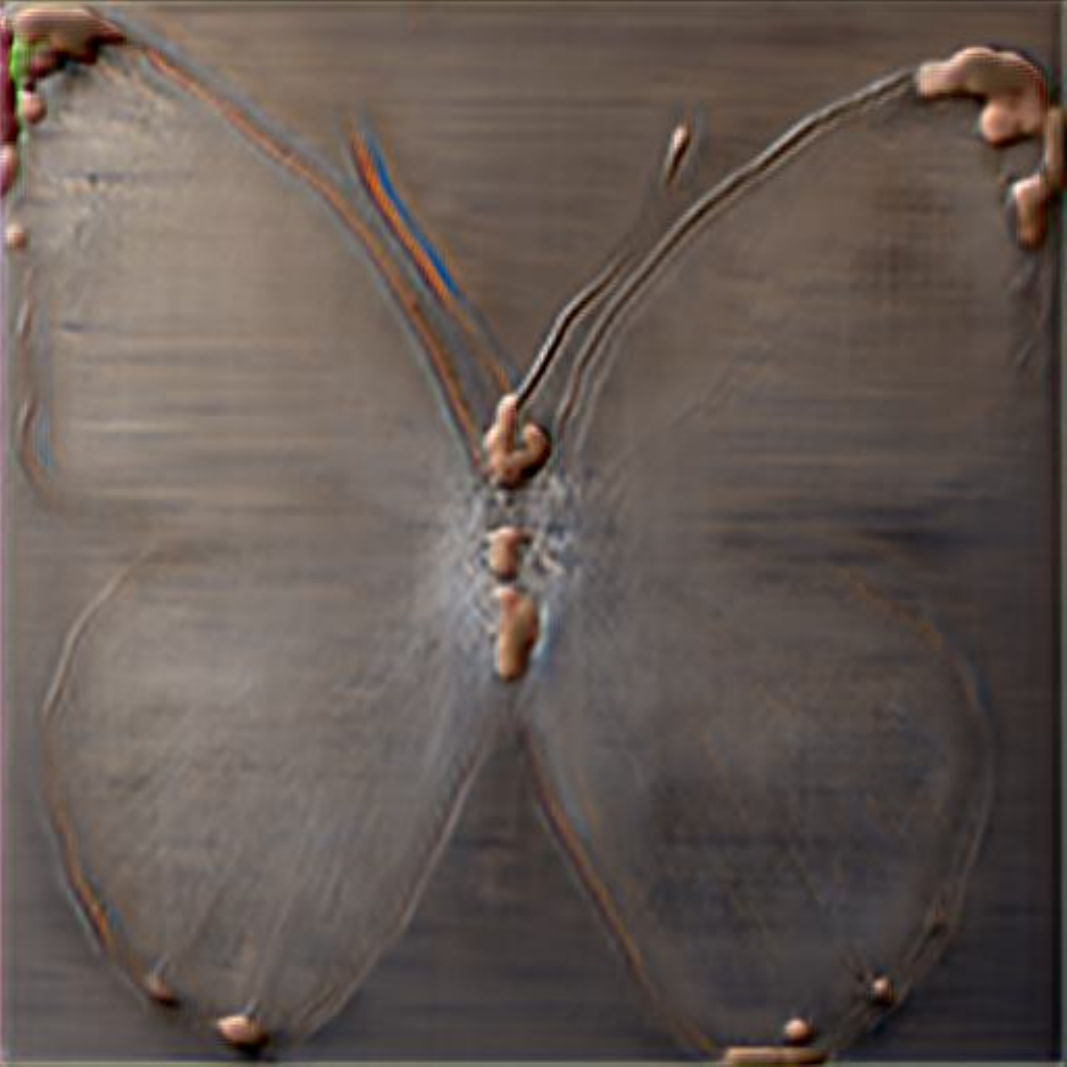}   &
      \includegraphics[width=0.24\linewidth]{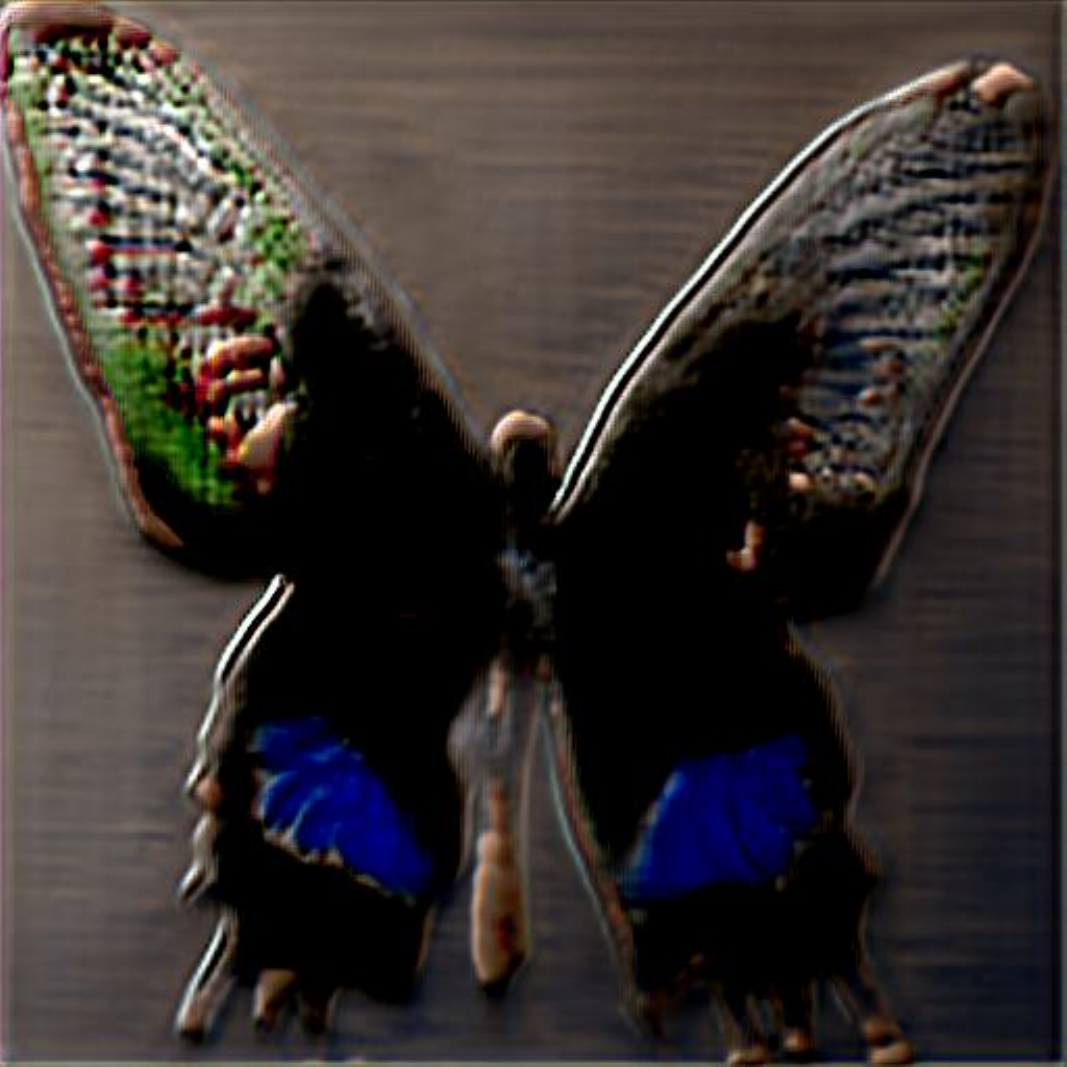}   
    \end{tabular}
\caption{First 16 outputs for the prompt "Hieronymus Bosch particle system rendered in Blender" using the coordinates-aware BM and InfoNCE loss.}
\label{fig:clip_tree}
\end{center}
\end{figure}

\begin{figure}[h!] \
\begin{center}
\setlength{\tabcolsep}{2pt}
    \begin{tabular}{cccc}
      \includegraphics[width=0.24\linewidth]{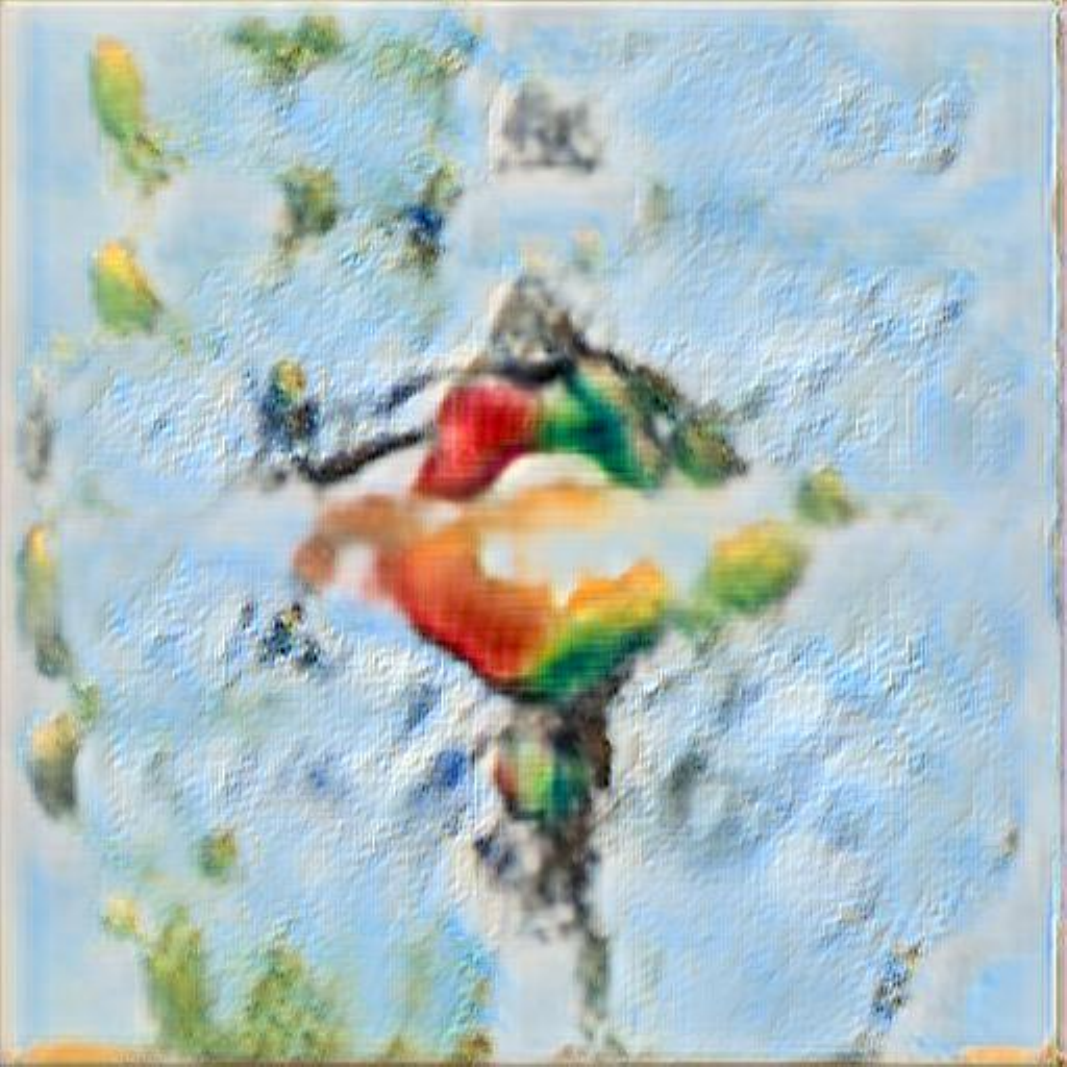}   &
      \includegraphics[width=0.24\linewidth]{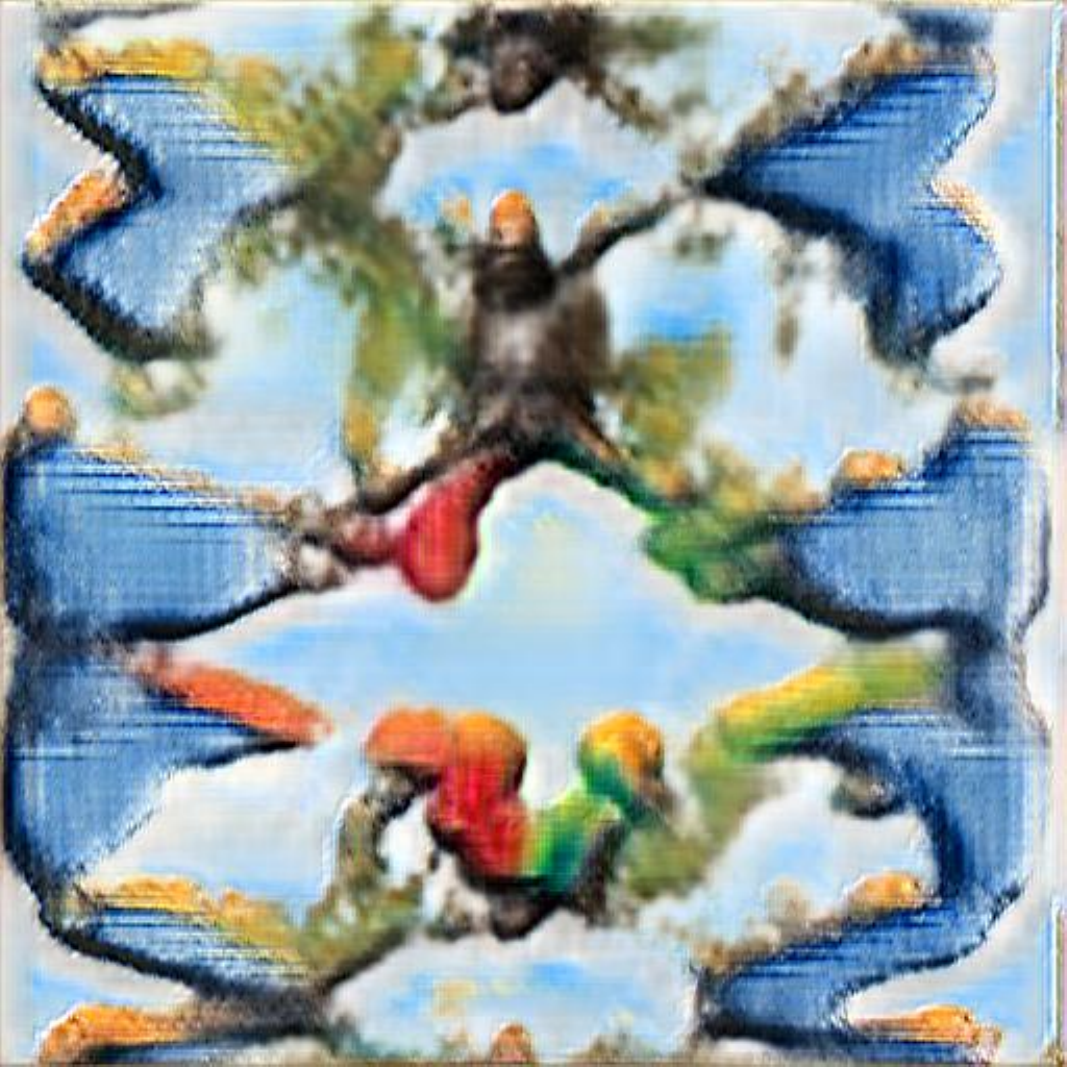}   &
      \includegraphics[width=0.24\linewidth]{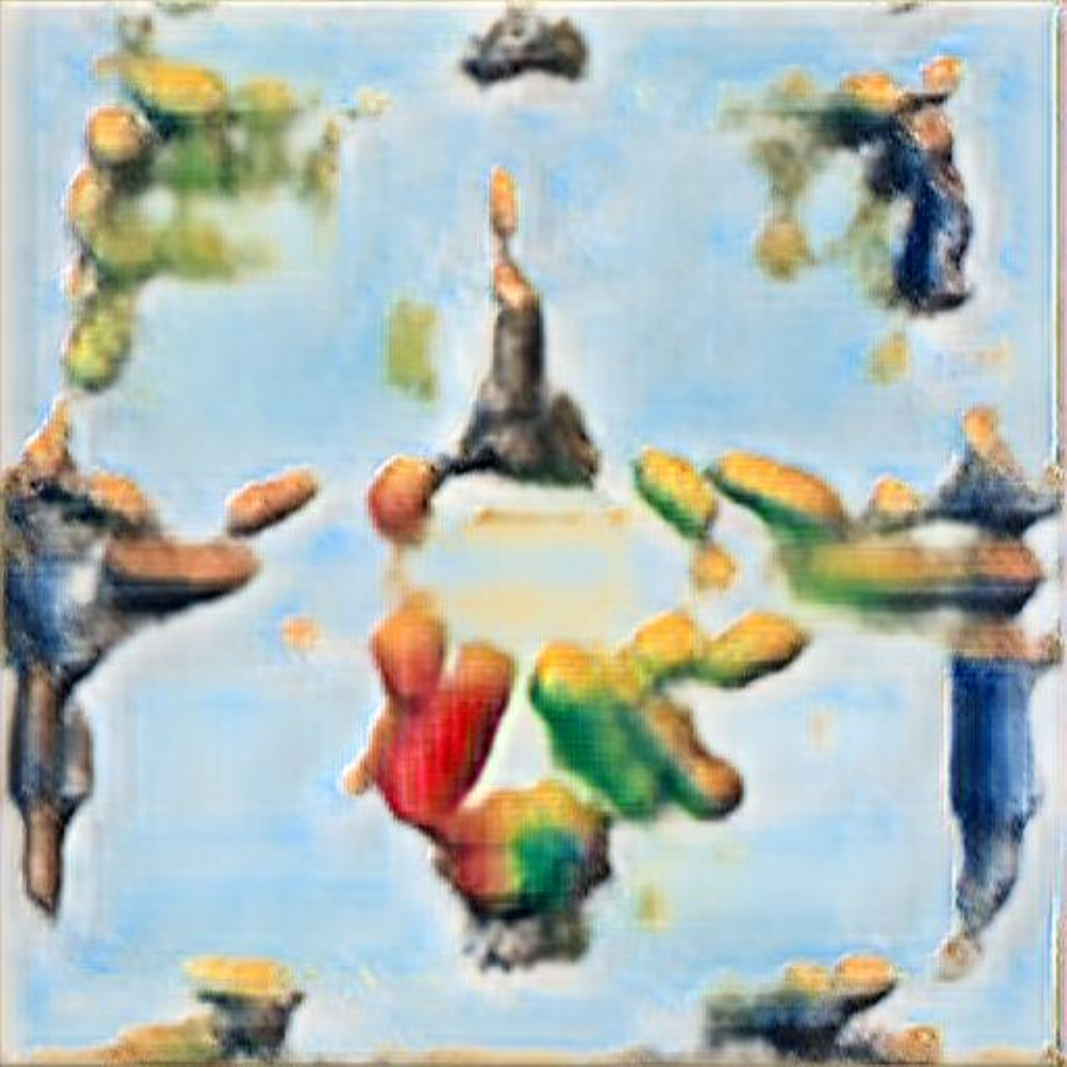}   &
      \includegraphics[width=0.24\linewidth]{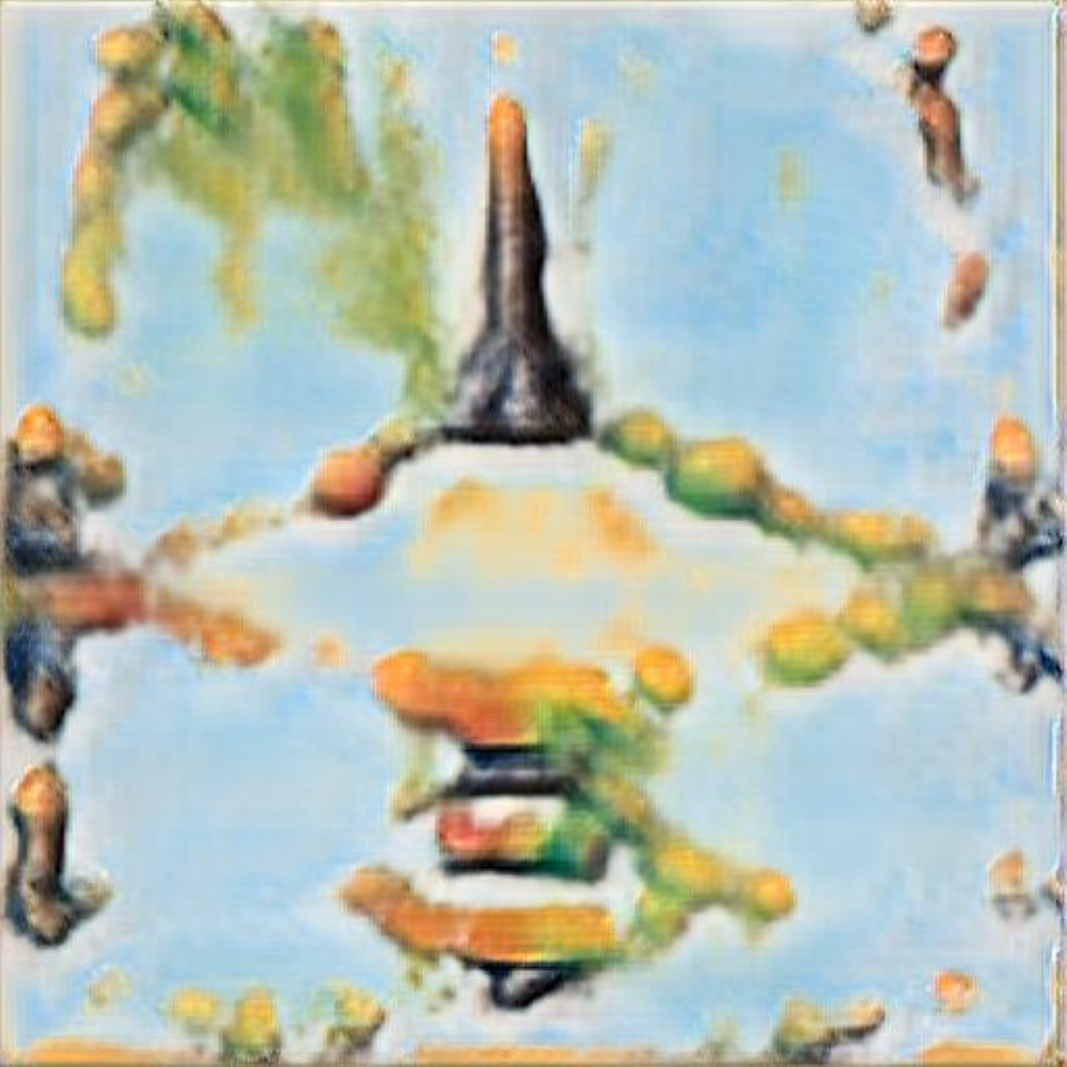}   \\
      \includegraphics[width=0.24\linewidth]{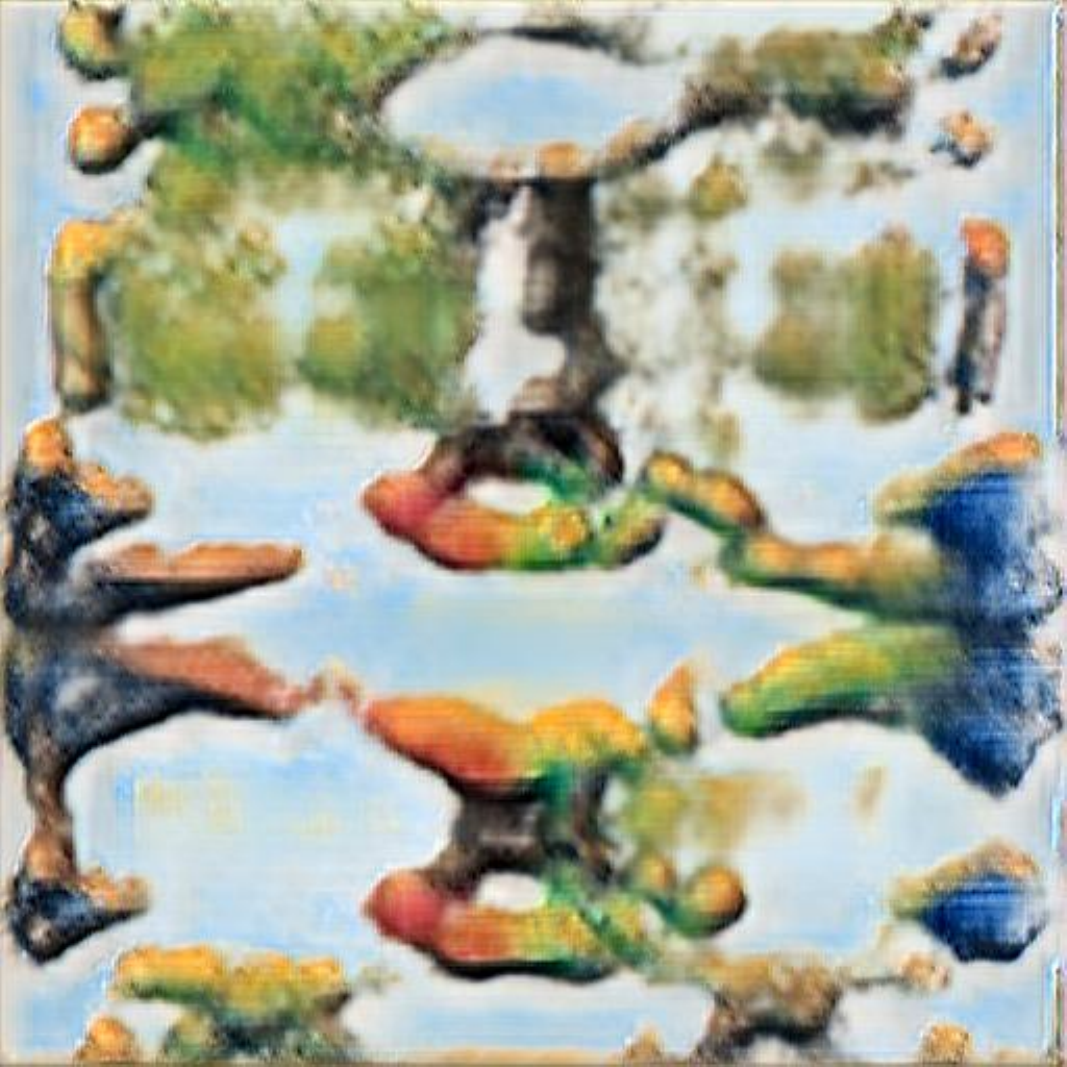}   &
      \includegraphics[width=0.24\linewidth]{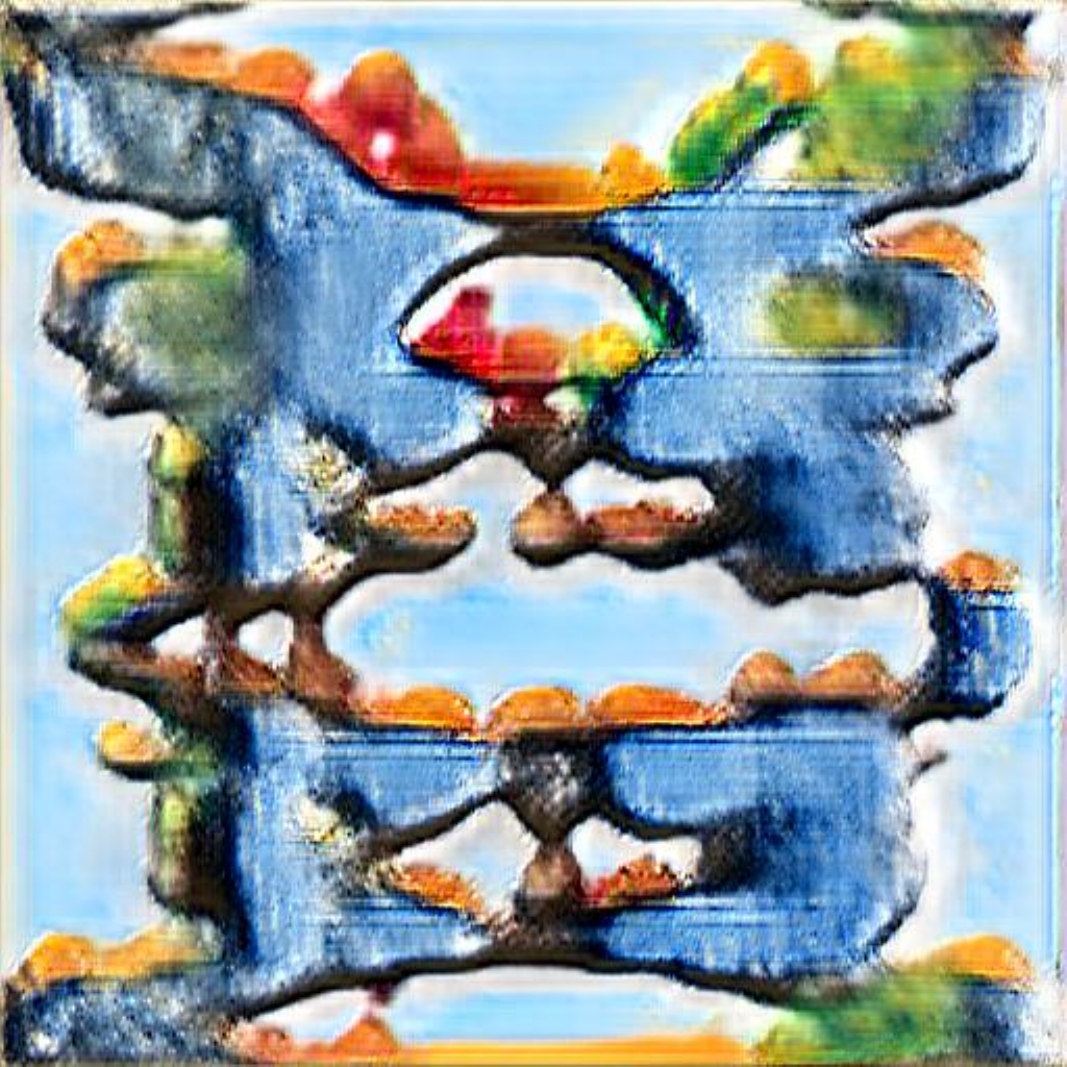}   &
      \includegraphics[width=0.24\linewidth]{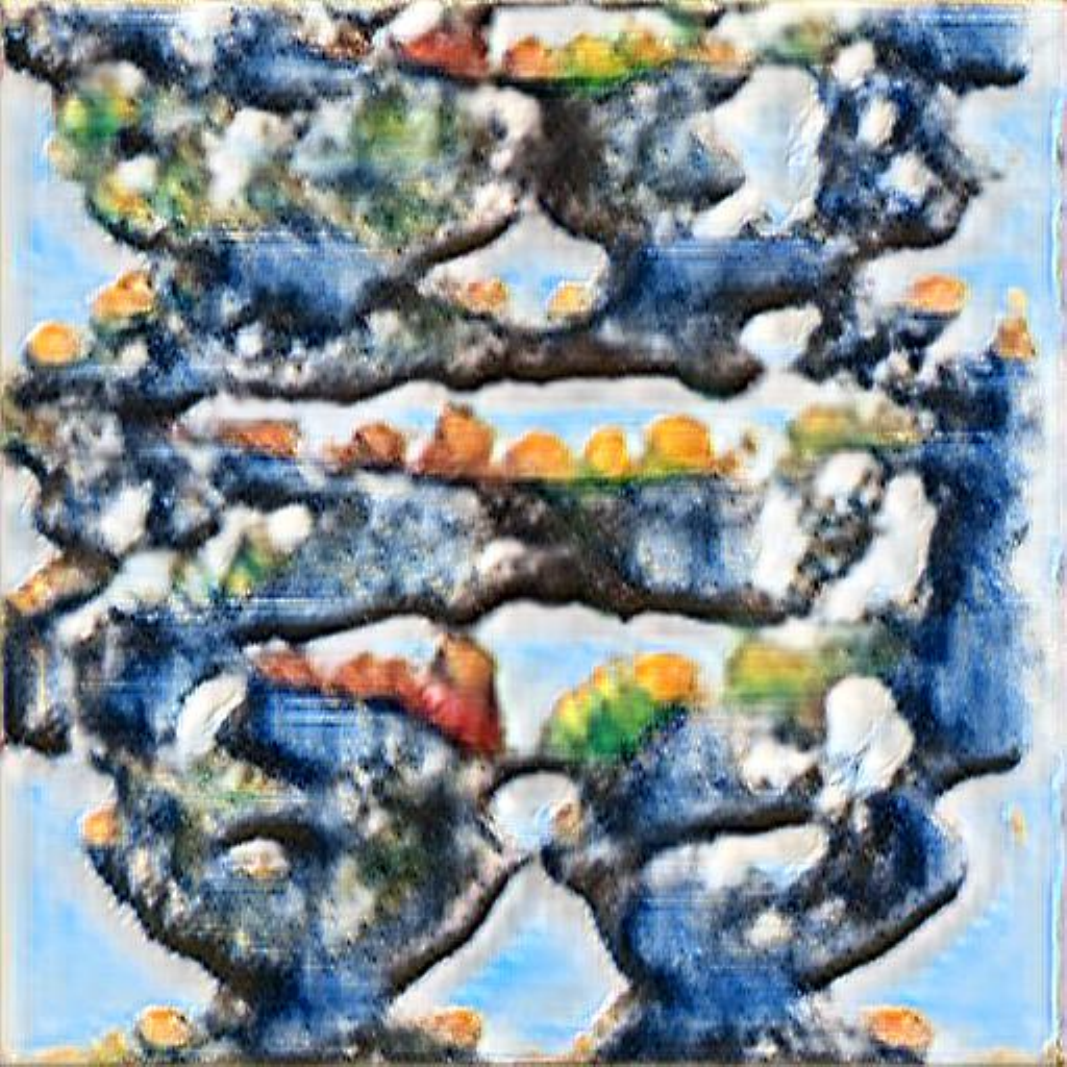}   &
      \includegraphics[width=0.24\linewidth]{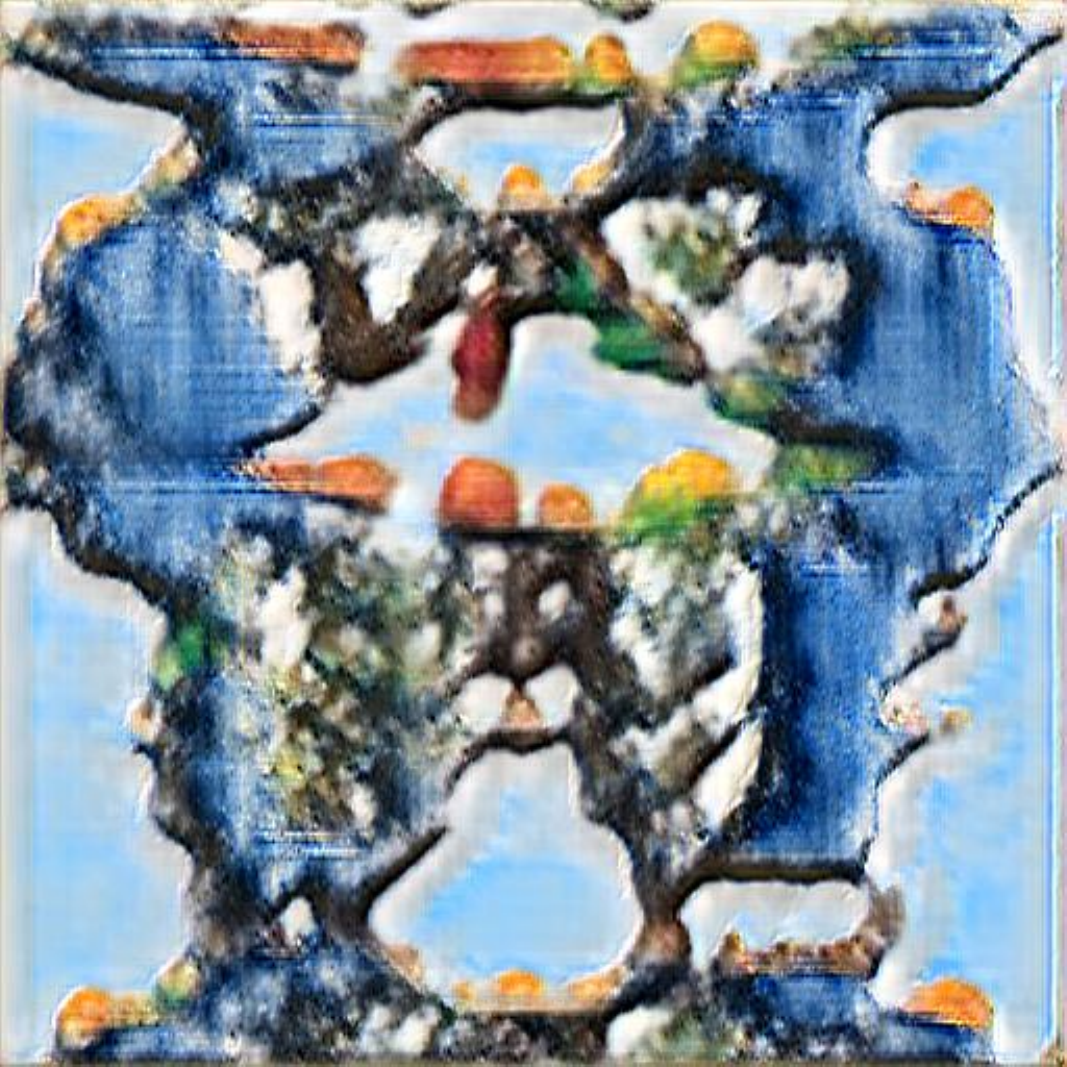}   \\
      \includegraphics[width=0.24\linewidth]{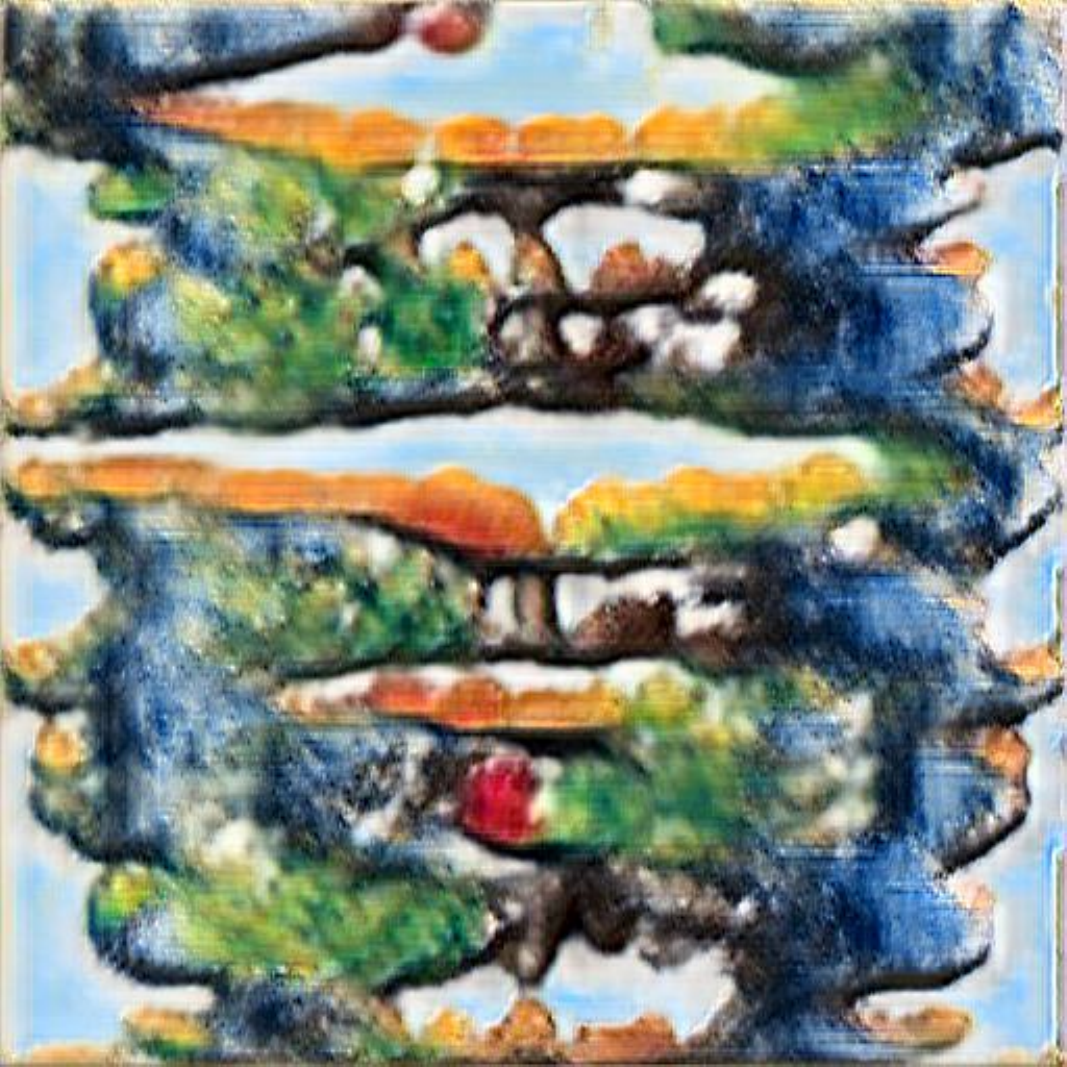}   &
      \includegraphics[width=0.24\linewidth]{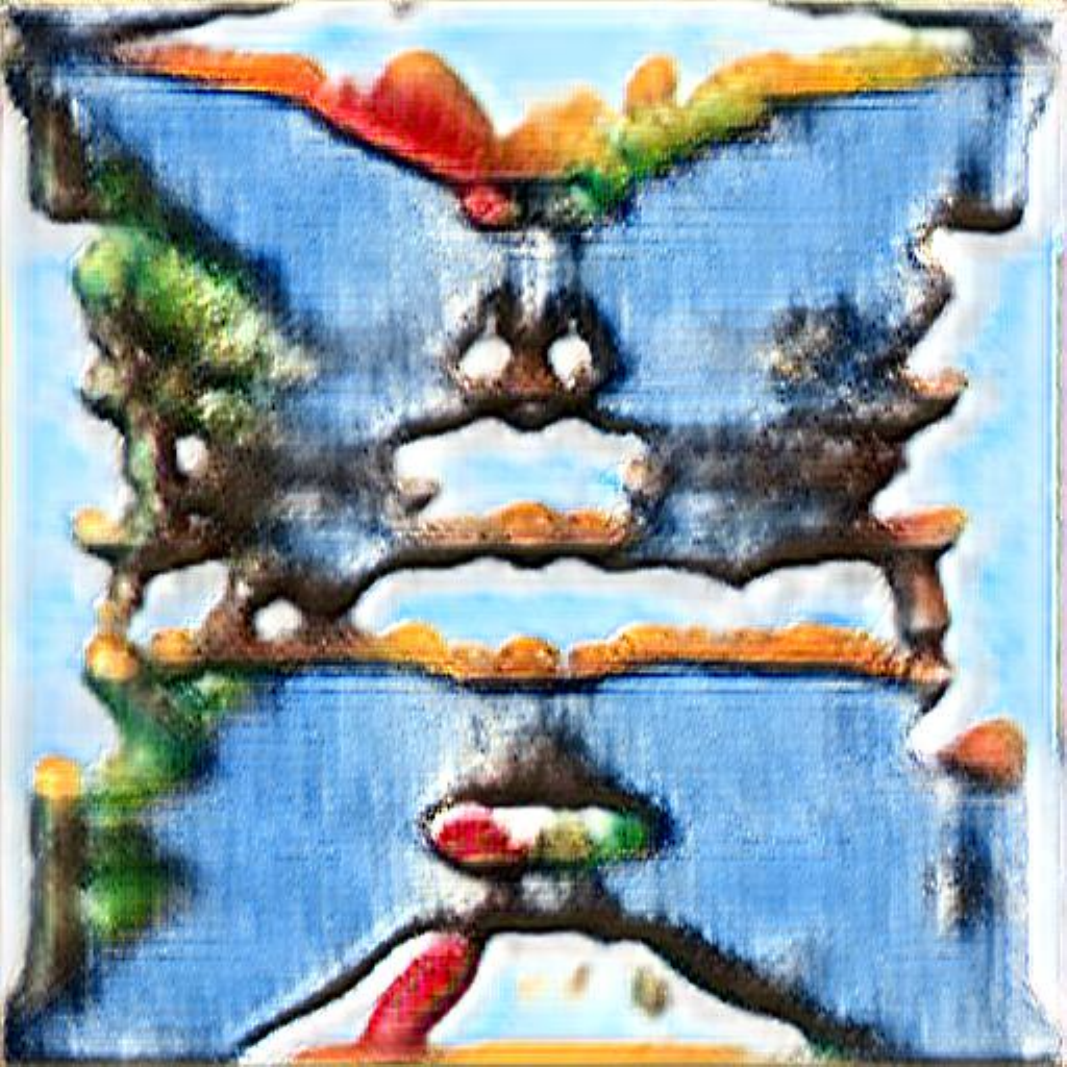}   &
      \includegraphics[width=0.24\linewidth]{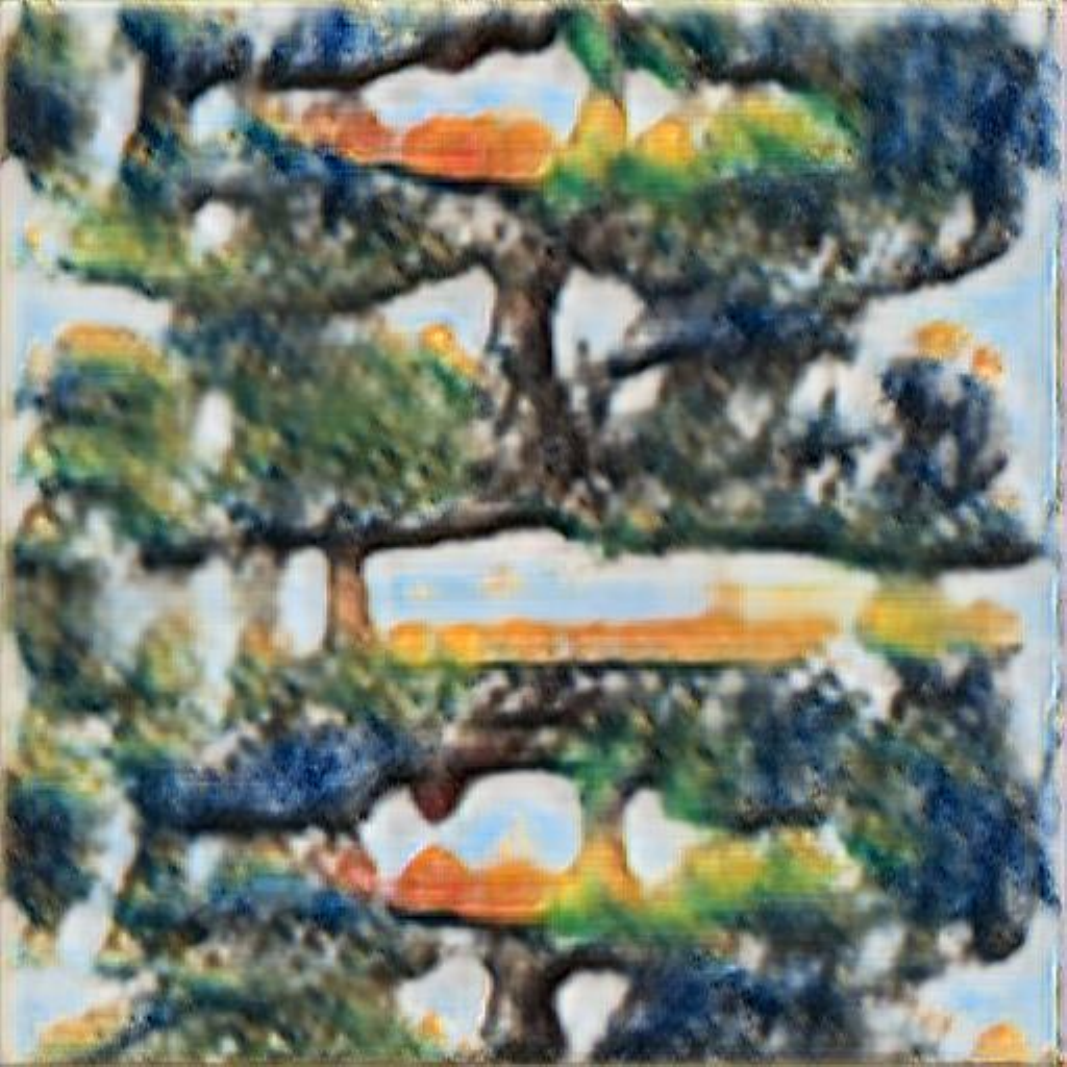}   &
      \includegraphics[width=0.24\linewidth]{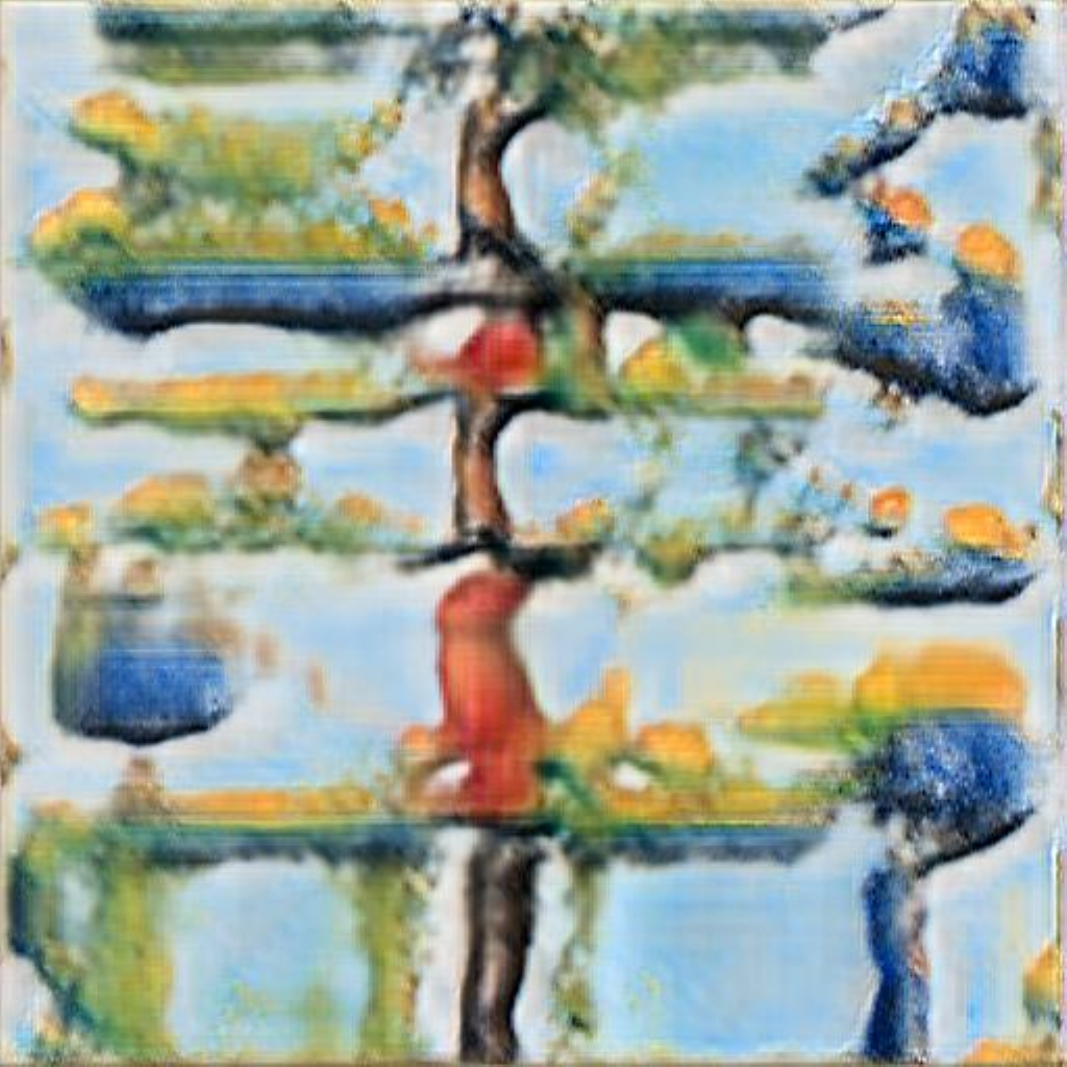}   \\
      \includegraphics[width=0.24\linewidth]{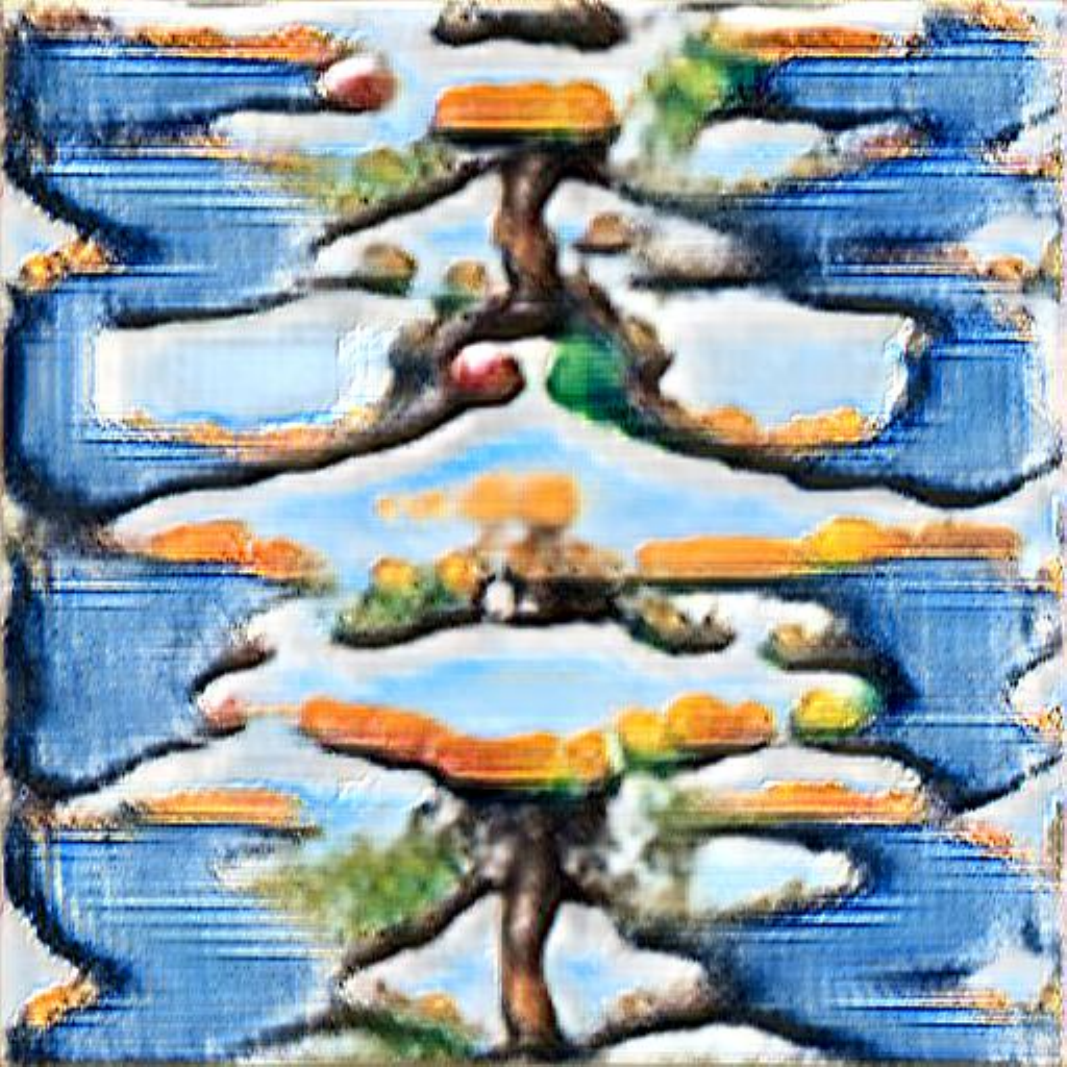}   &
      \includegraphics[width=0.24\linewidth]{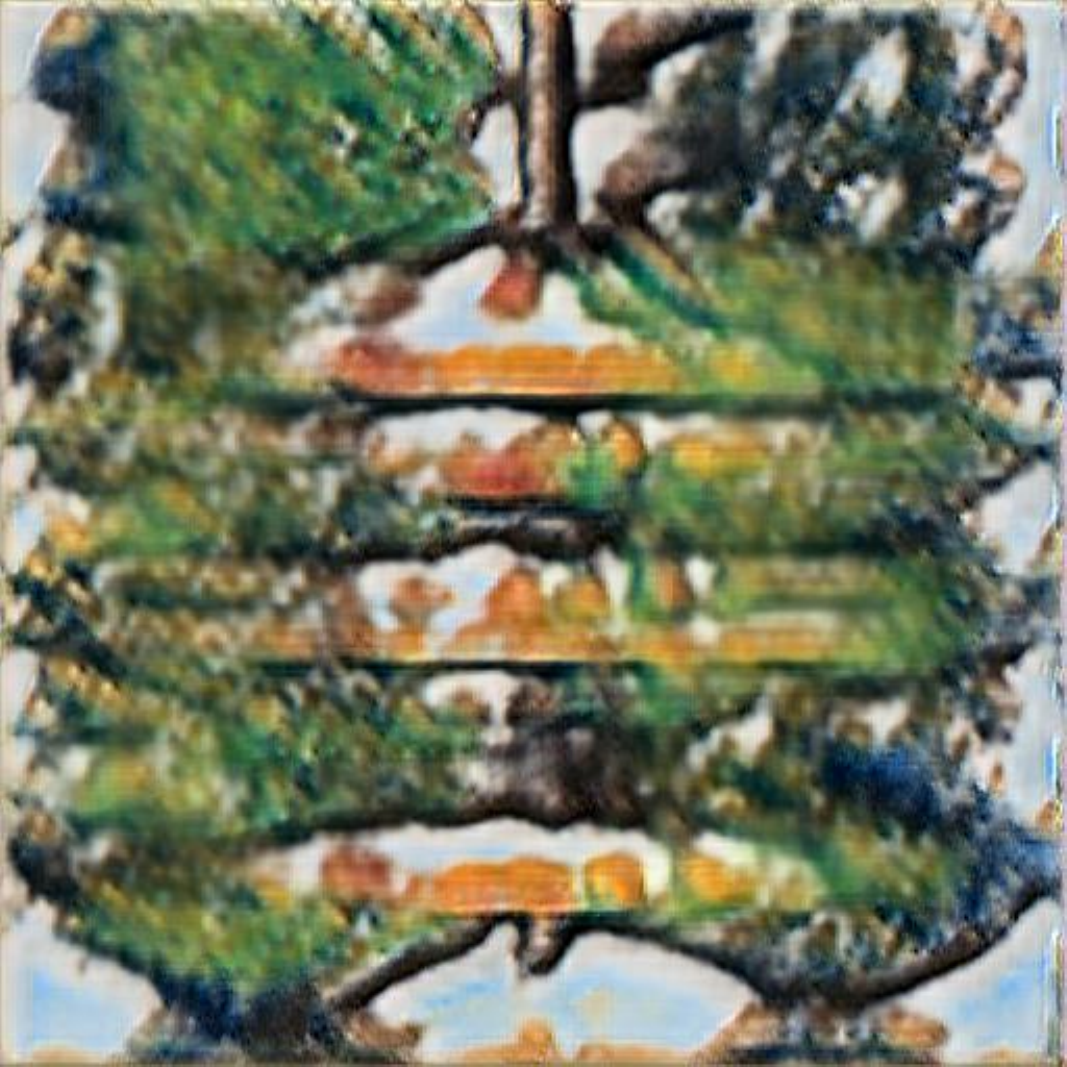}   &
      \includegraphics[width=0.24\linewidth]{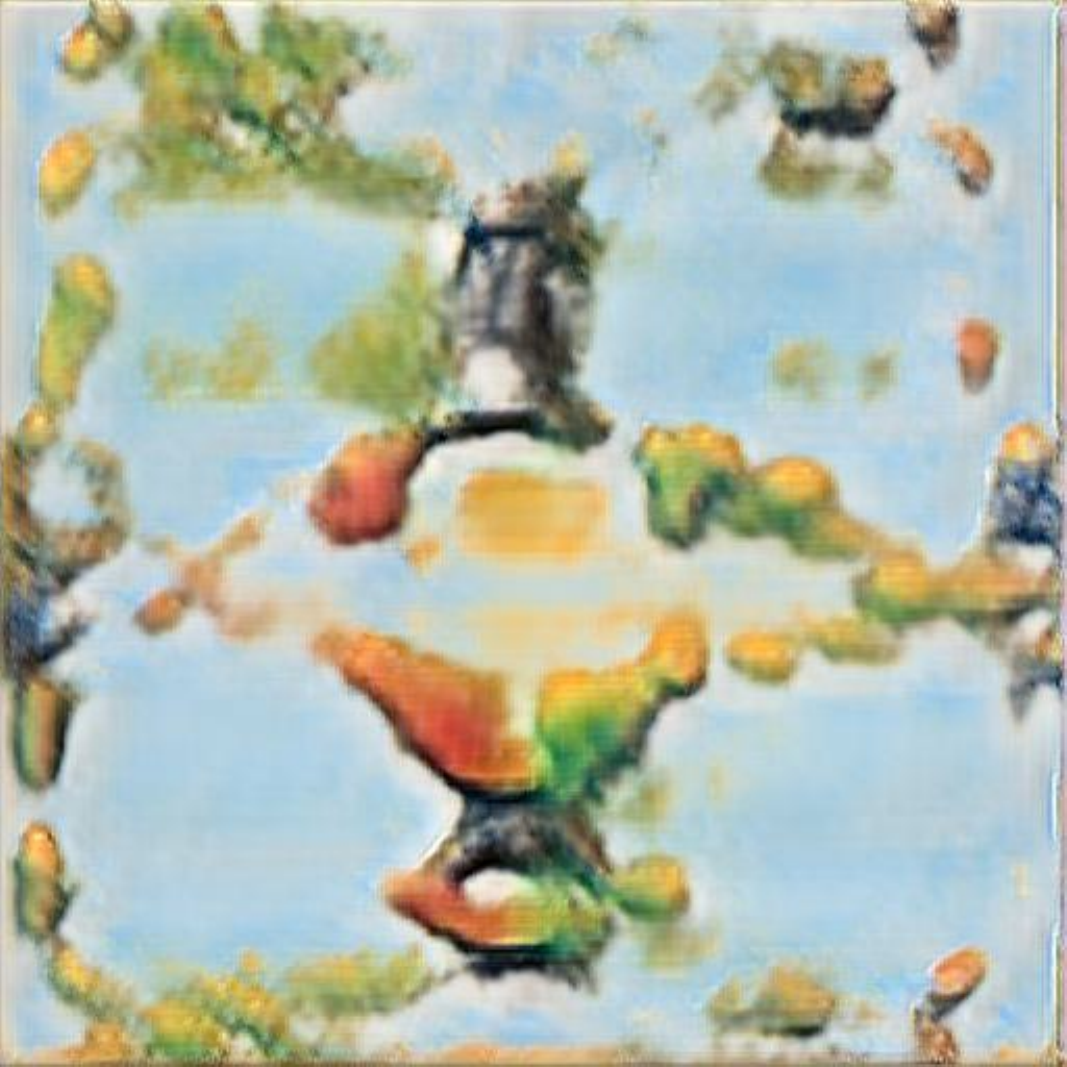}   &
      \includegraphics[width=0.24\linewidth]{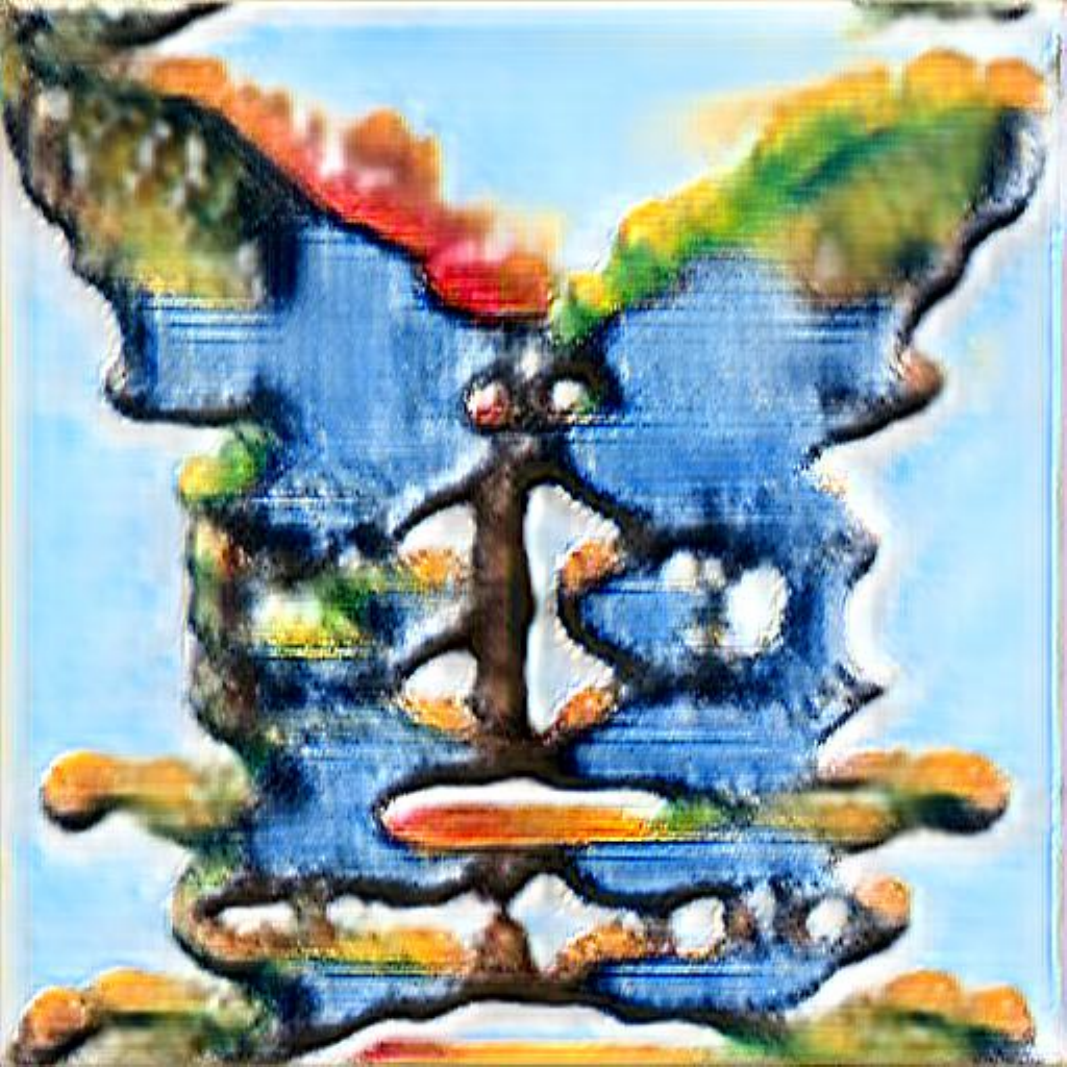}   
    \end{tabular}
\caption{First 16 outputs for the prompt "A colorful tree on a blue sky painted by Cézanne", using the Convolutional + Sorting BM and InfoNCE loss.}
\label{fig:clip_tree}
\end{center}
\end{figure}

\begin{figure}[h!] \
\begin{center}
\setlength{\tabcolsep}{2pt}
    \begin{tabular}{cccc}
      \includegraphics[width=0.24\linewidth]{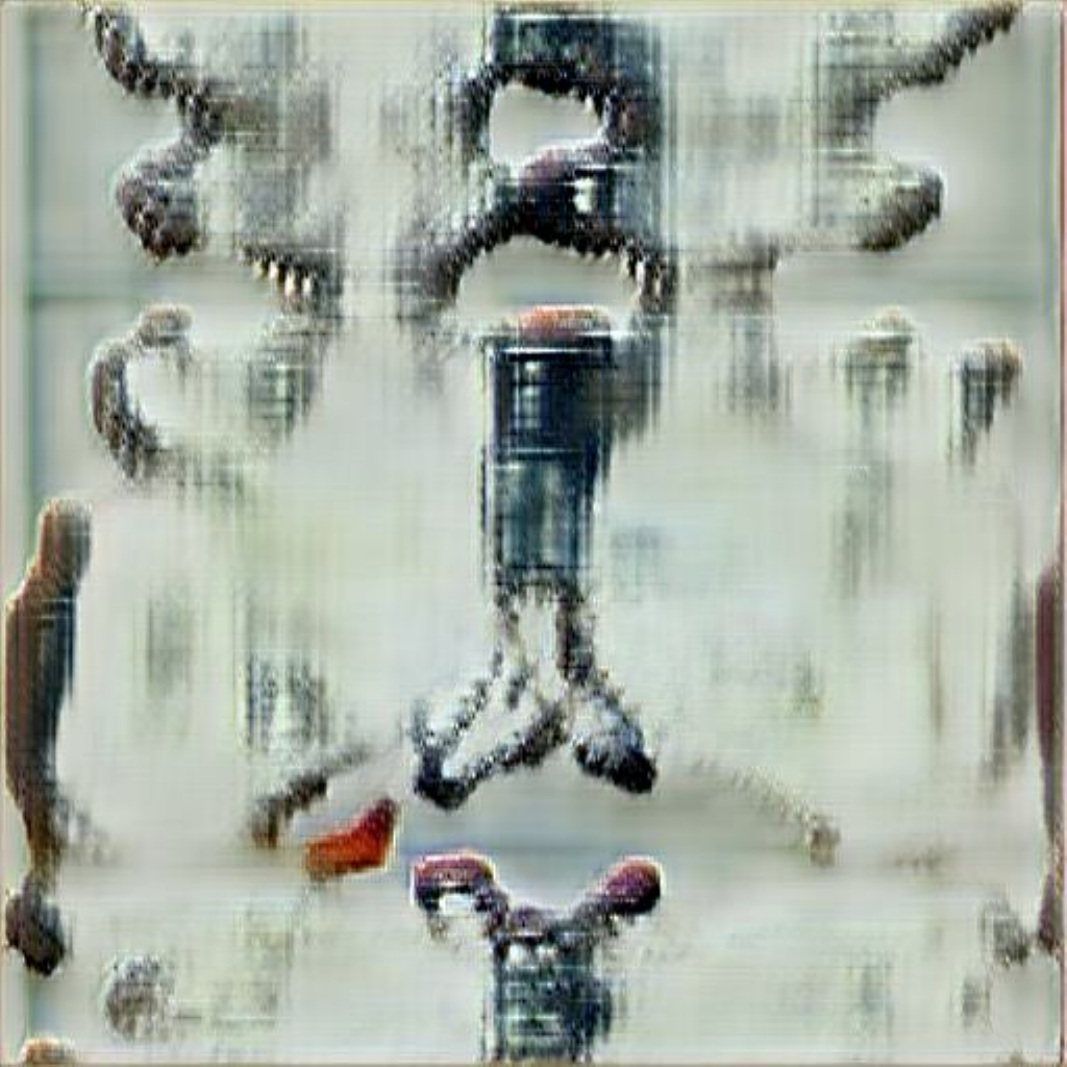}   &
      \includegraphics[width=0.24\linewidth]{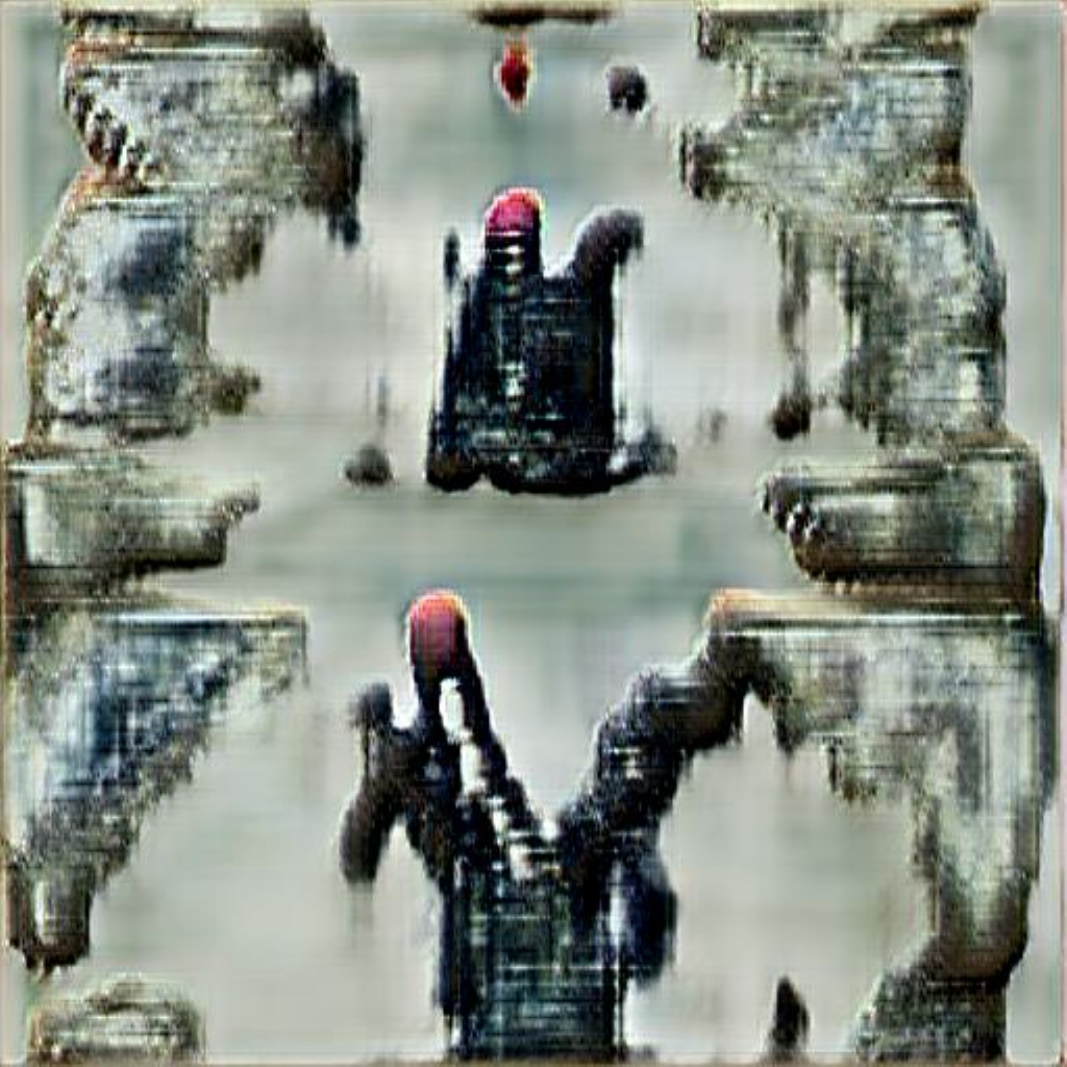}   &
      \includegraphics[width=0.24\linewidth]{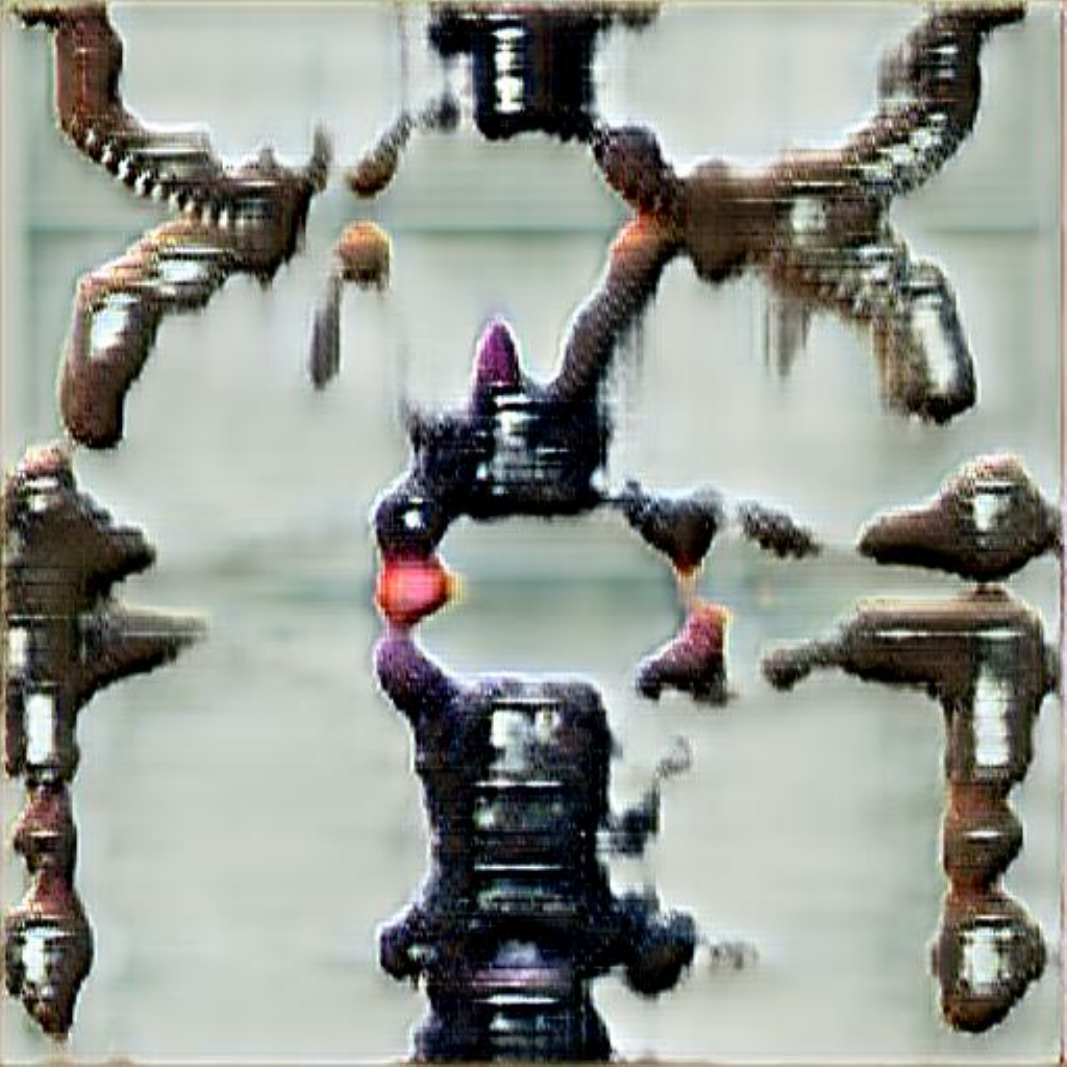}   &
      \includegraphics[width=0.24\linewidth]{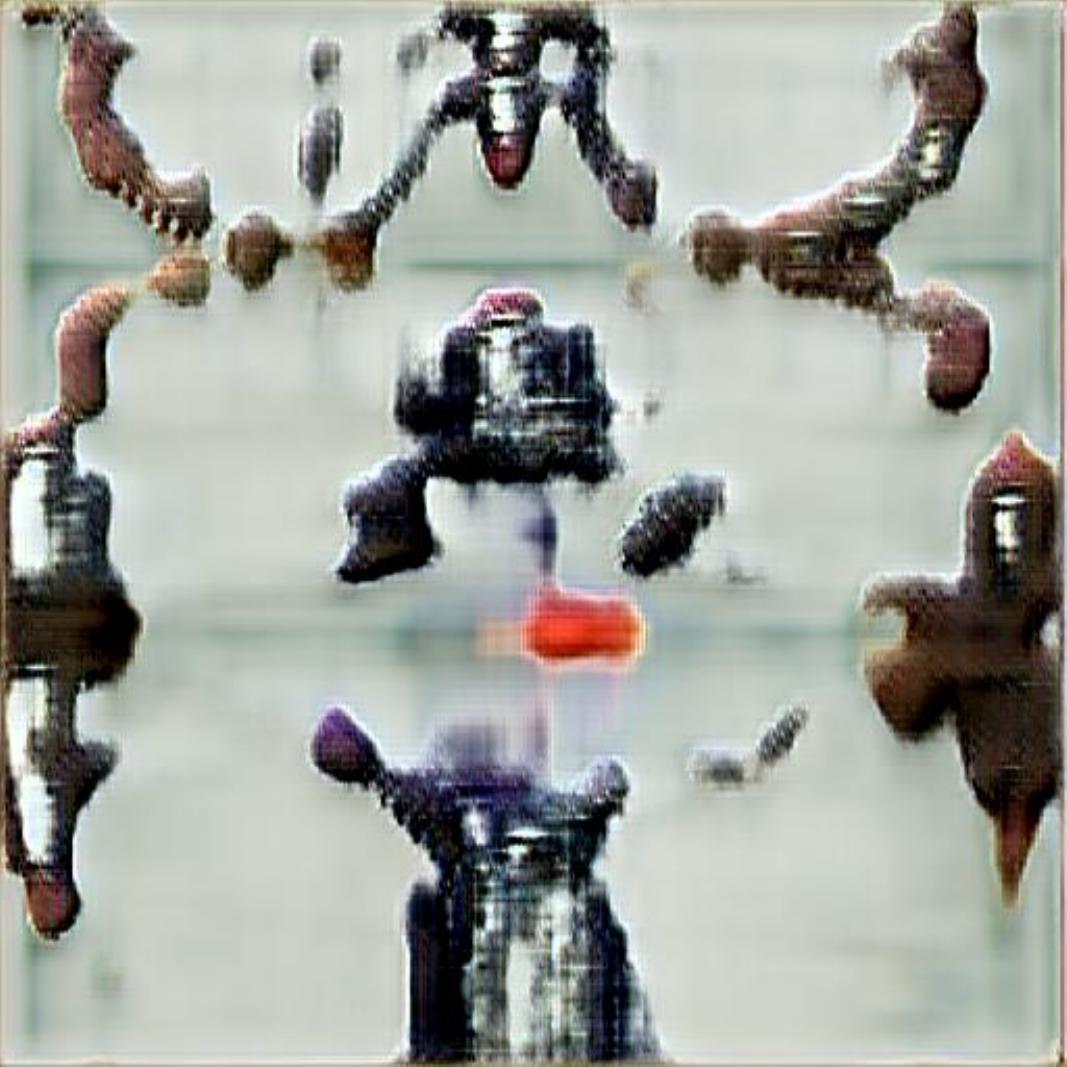}   \\
      \includegraphics[width=0.24\linewidth]{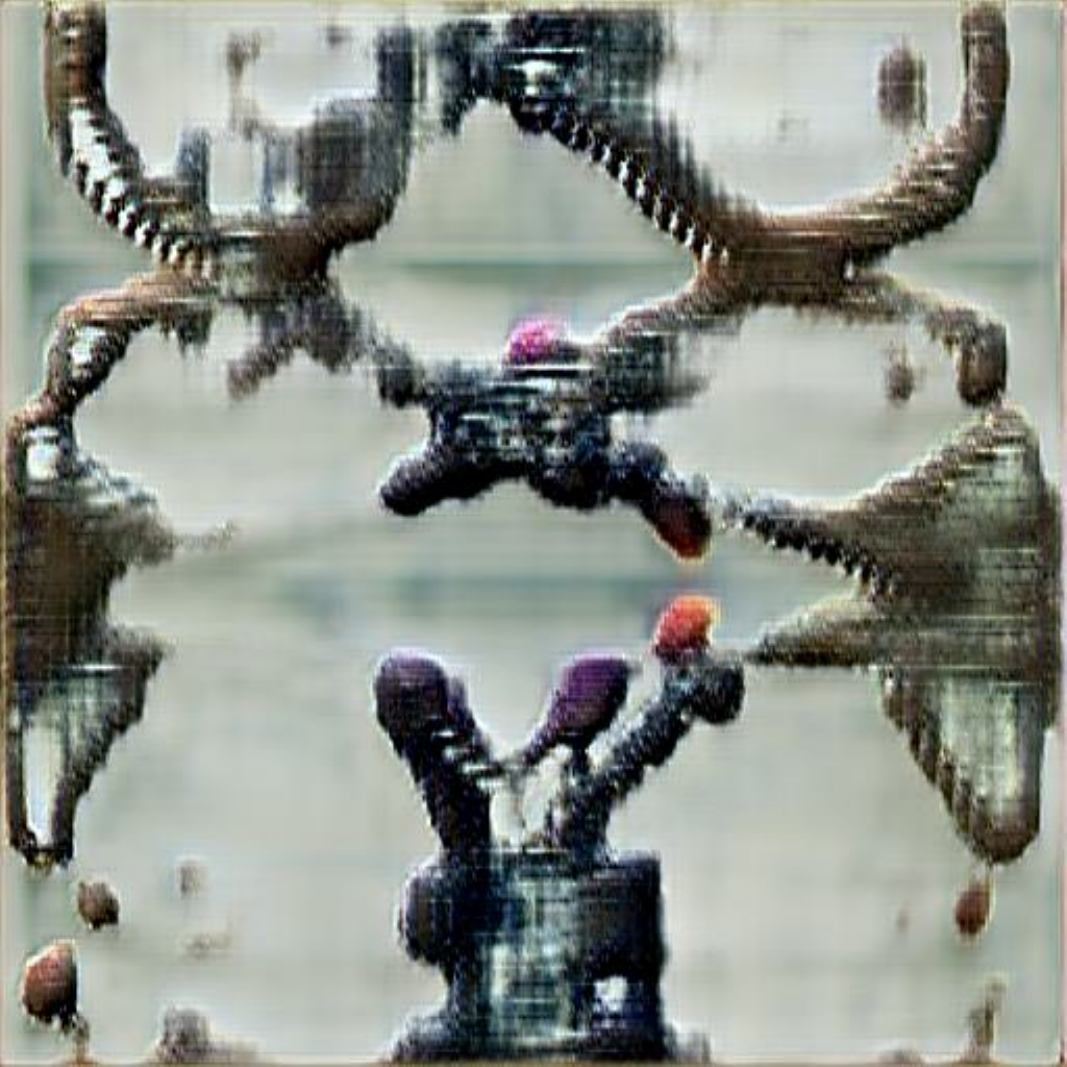}   &
      \includegraphics[width=0.24\linewidth]{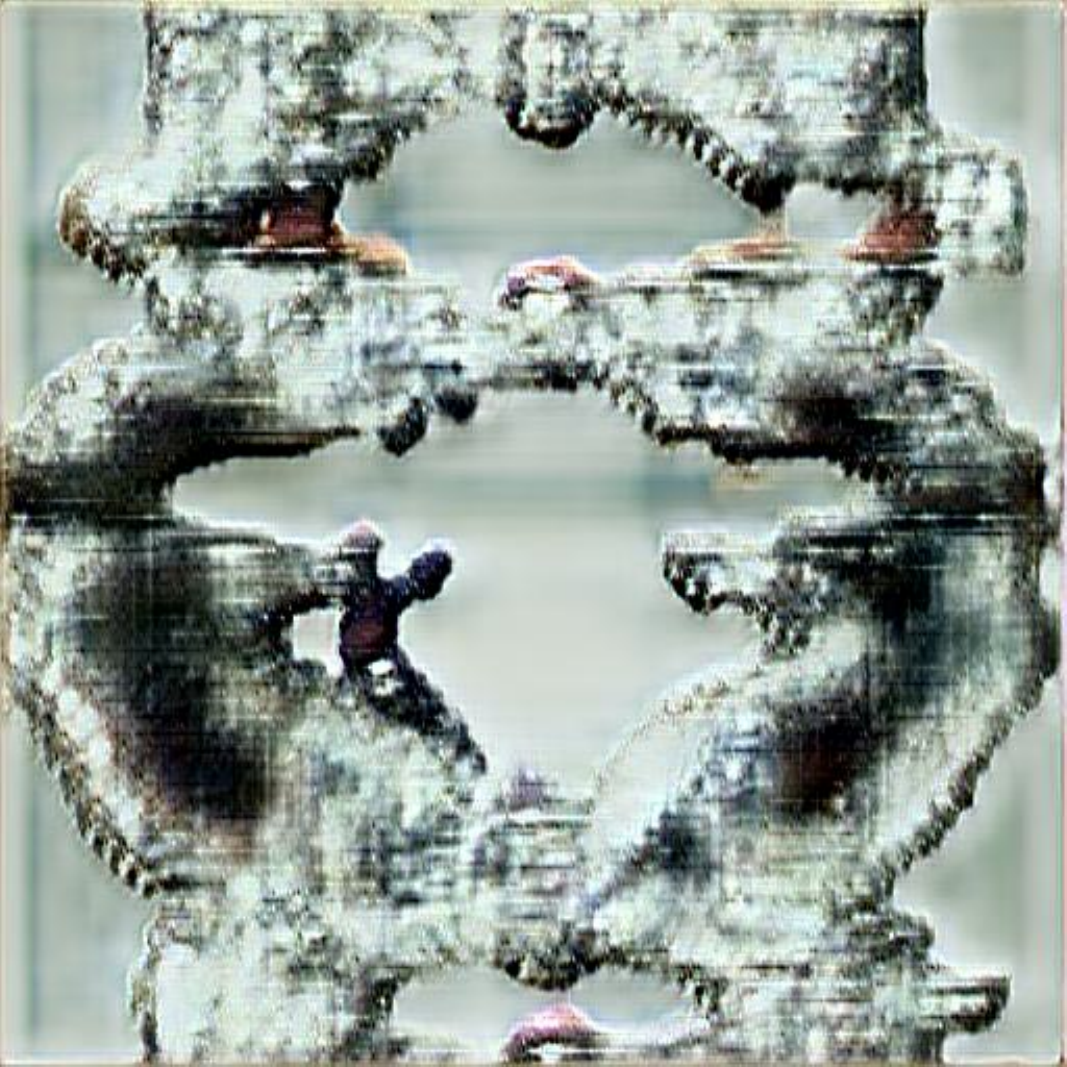}   &
      \includegraphics[width=0.24\linewidth]{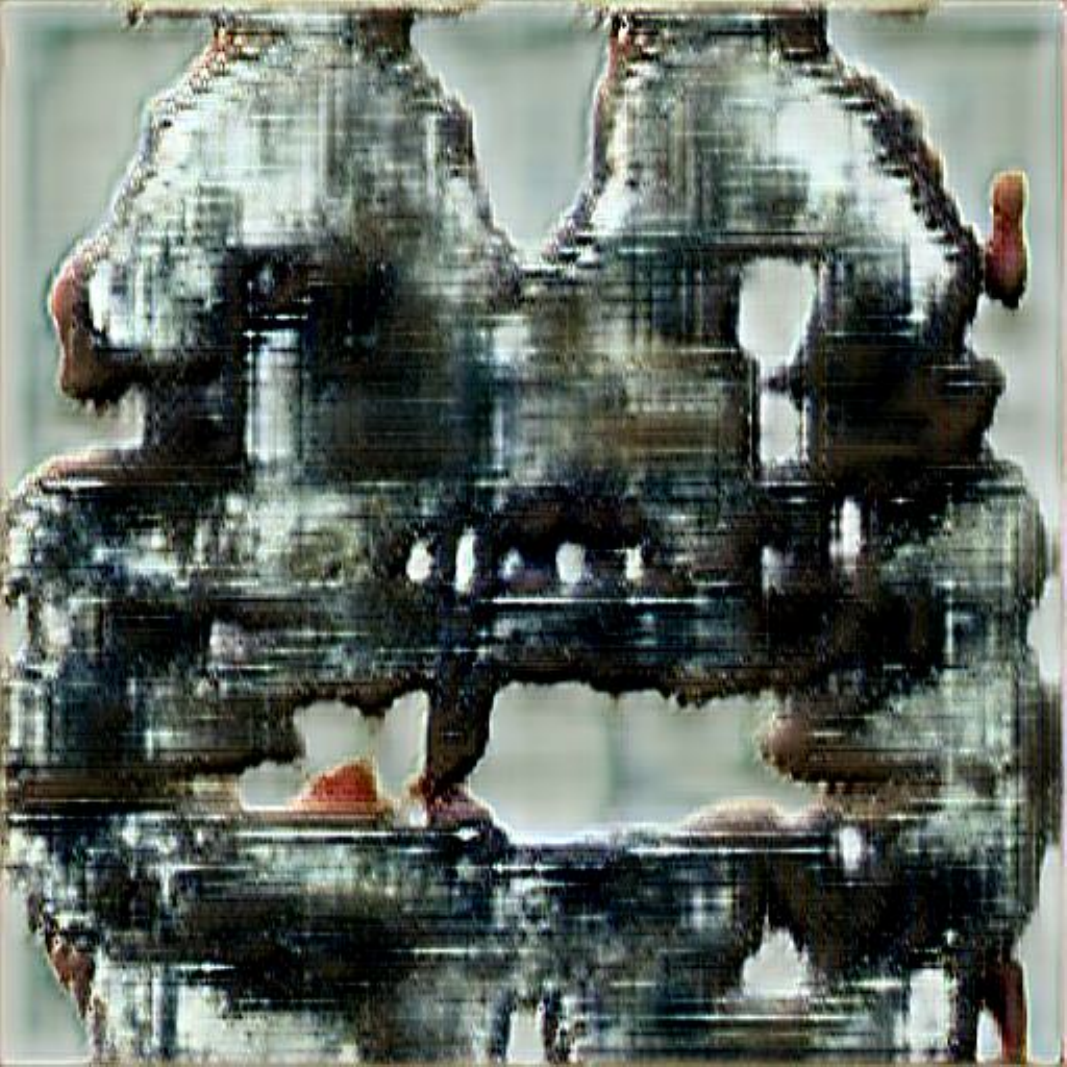}   &
      \includegraphics[width=0.24\linewidth]{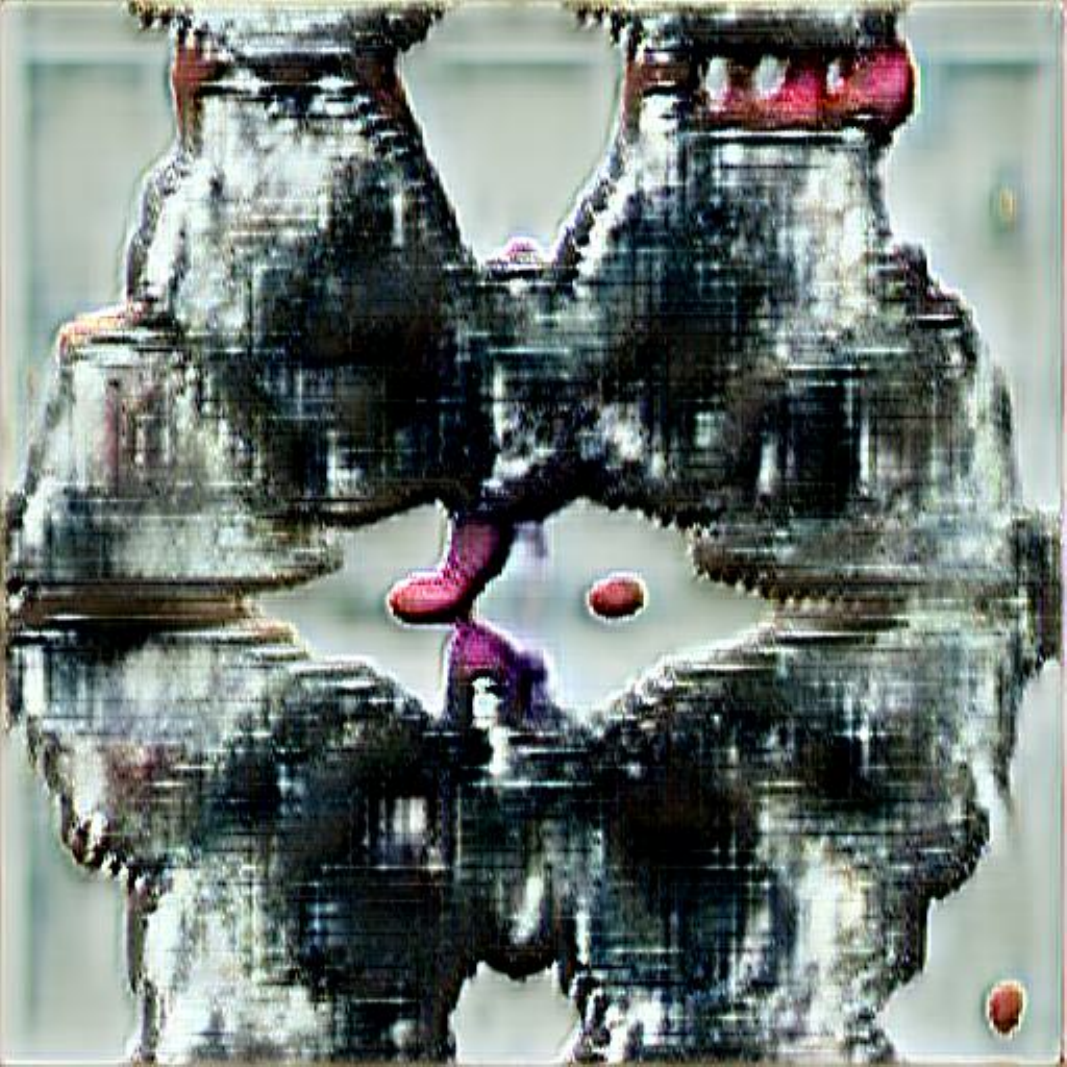}   \\
      \includegraphics[width=0.24\linewidth]{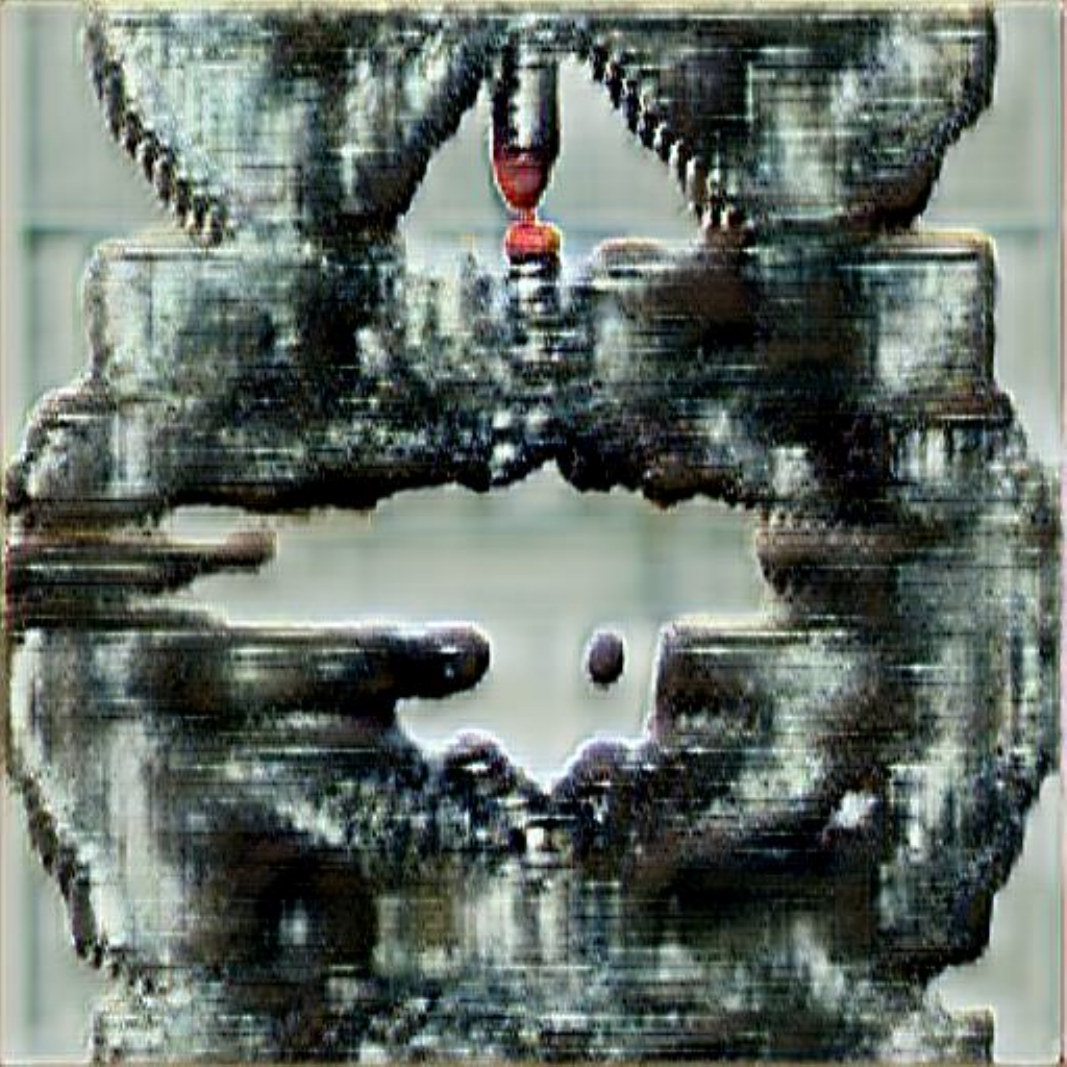}   &
      \includegraphics[width=0.24\linewidth]{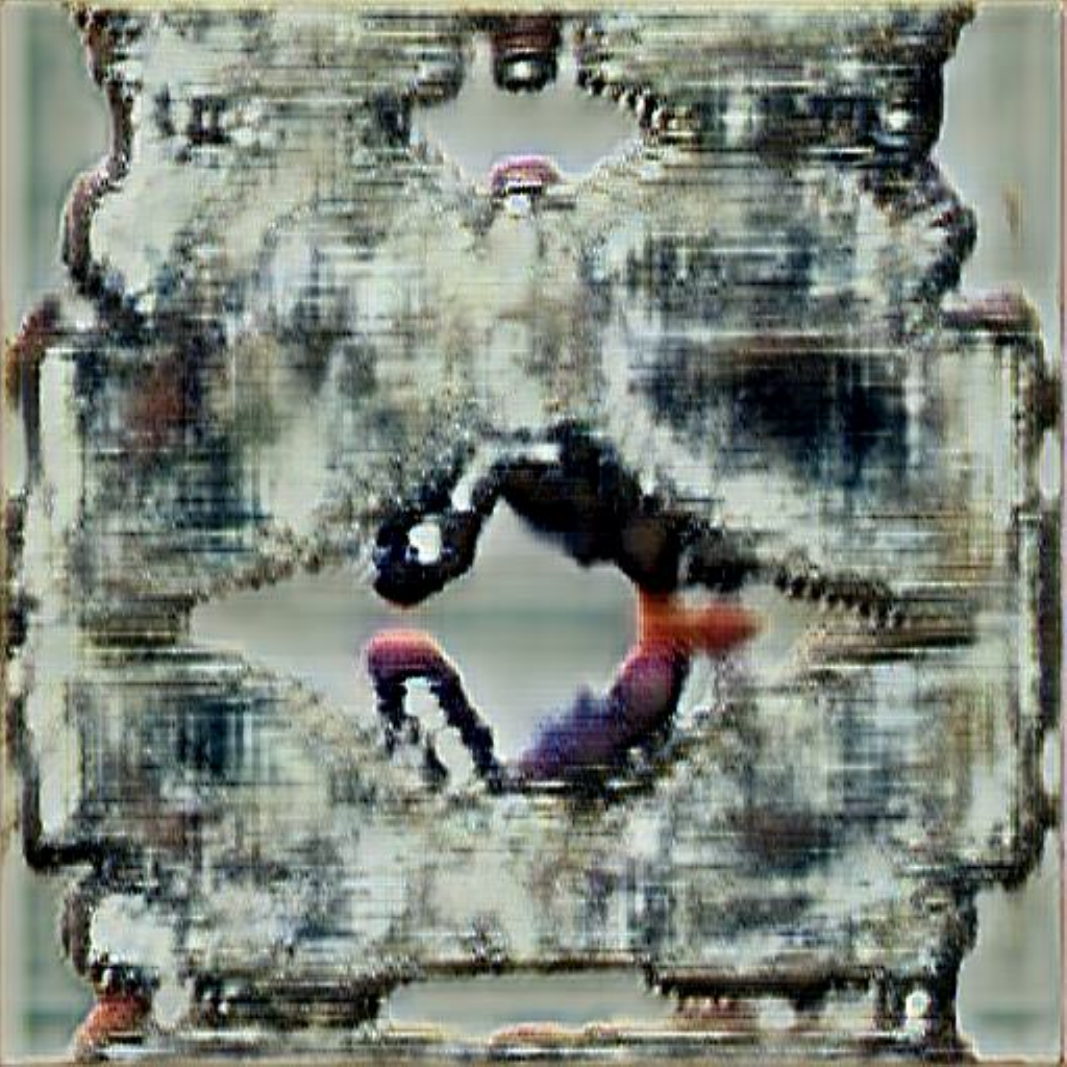}   &
      \includegraphics[width=0.24\linewidth]{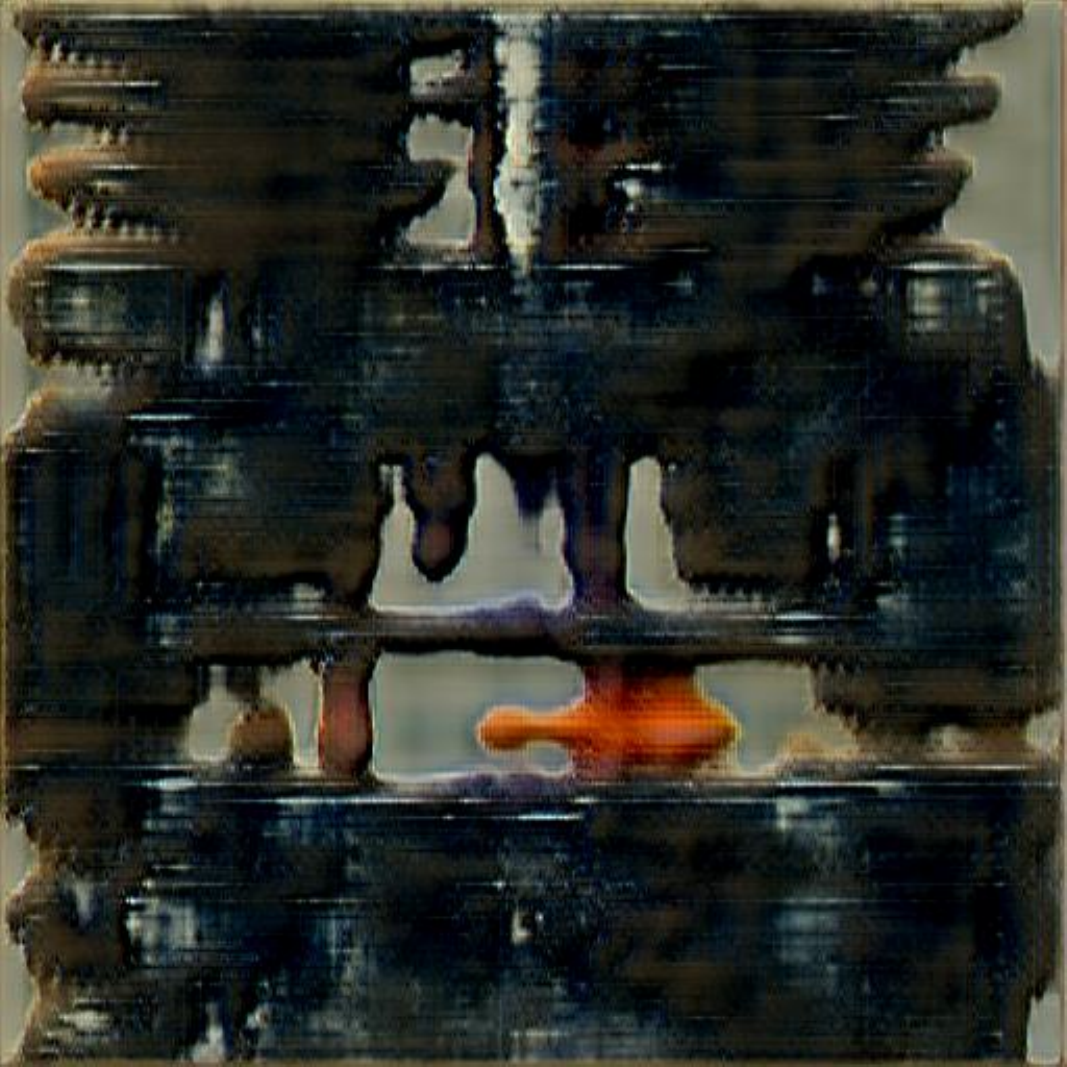}   &
      \includegraphics[width=0.24\linewidth]{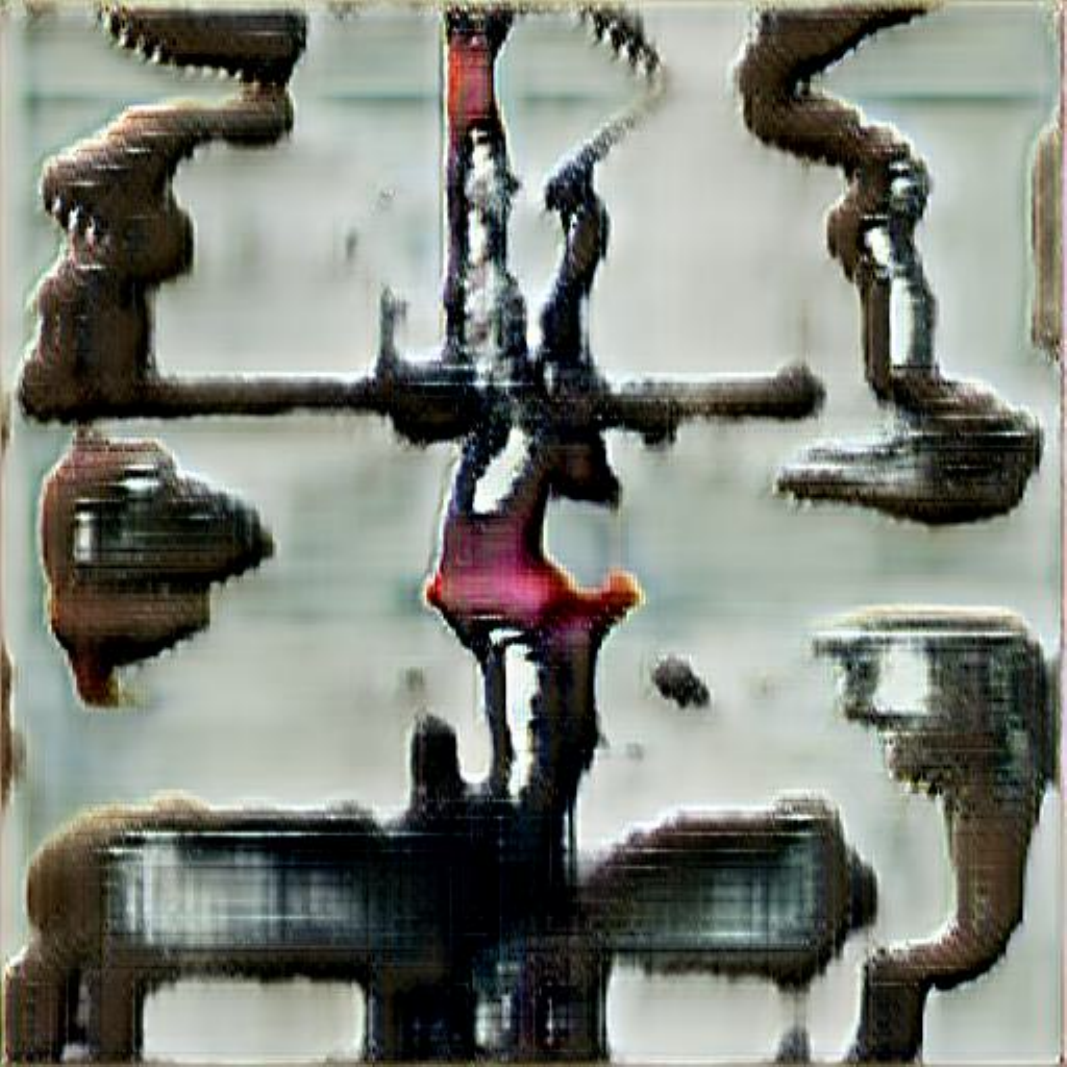}   \\
      \includegraphics[width=0.24\linewidth]{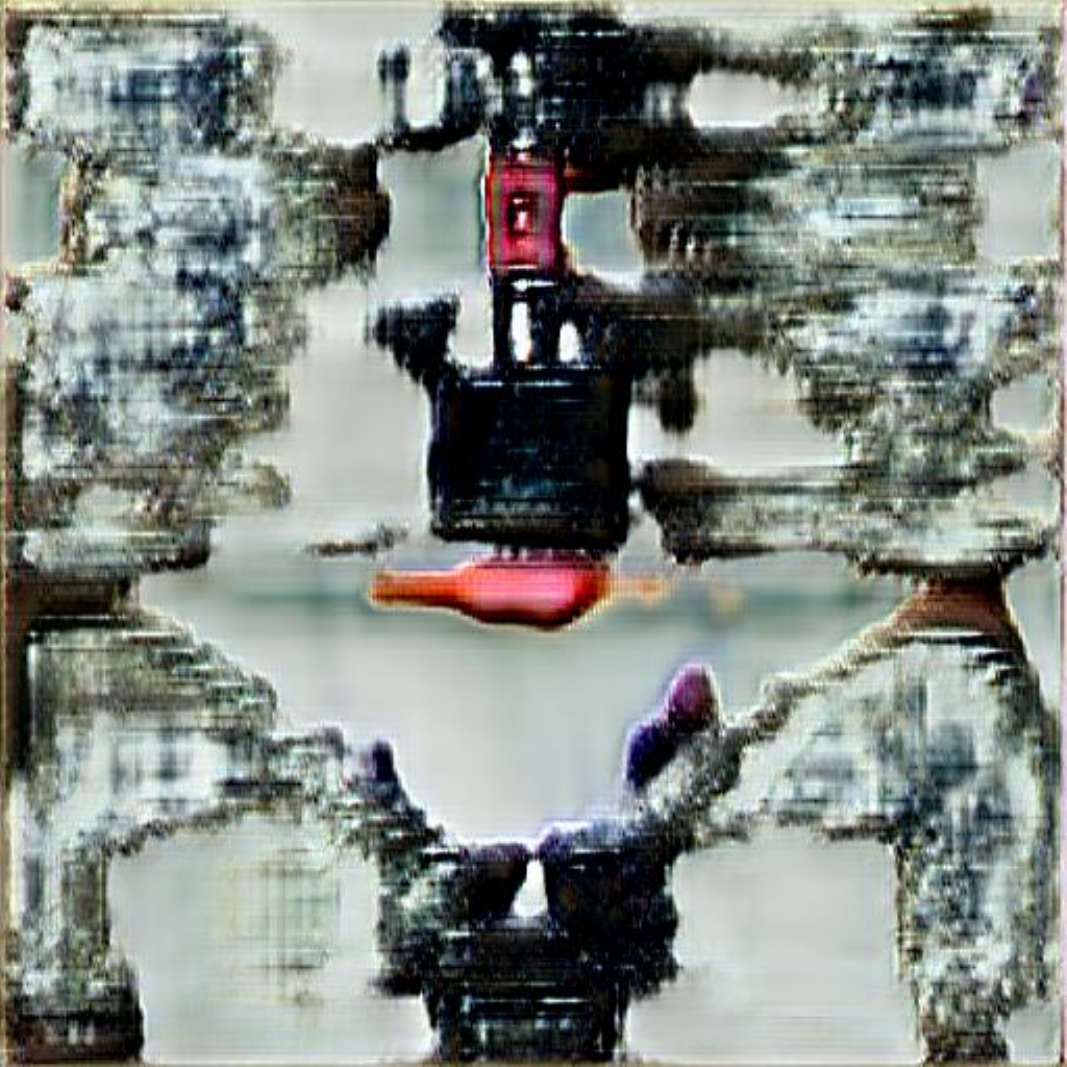}   &
      \includegraphics[width=0.24\linewidth]{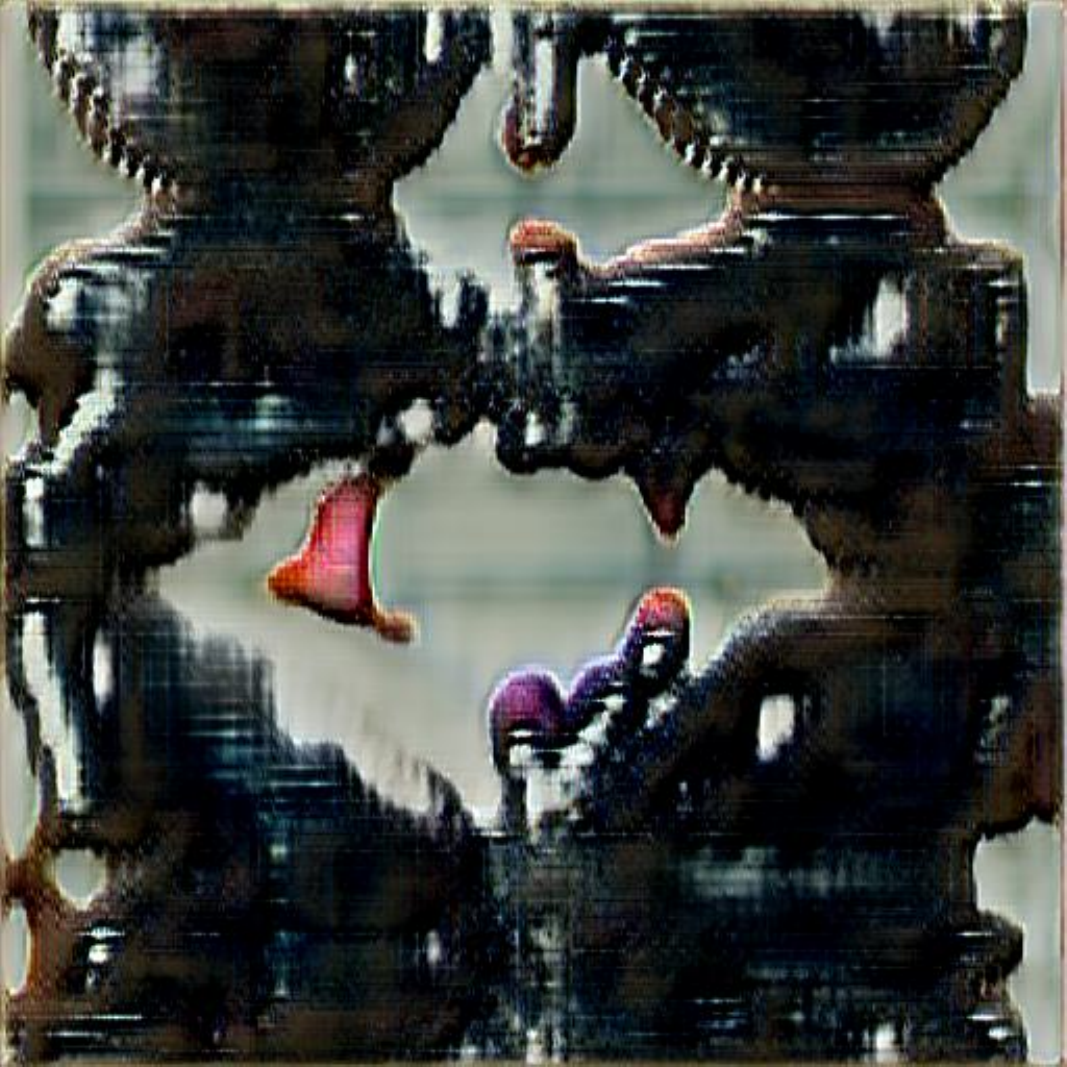}   &
      \includegraphics[width=0.24\linewidth]{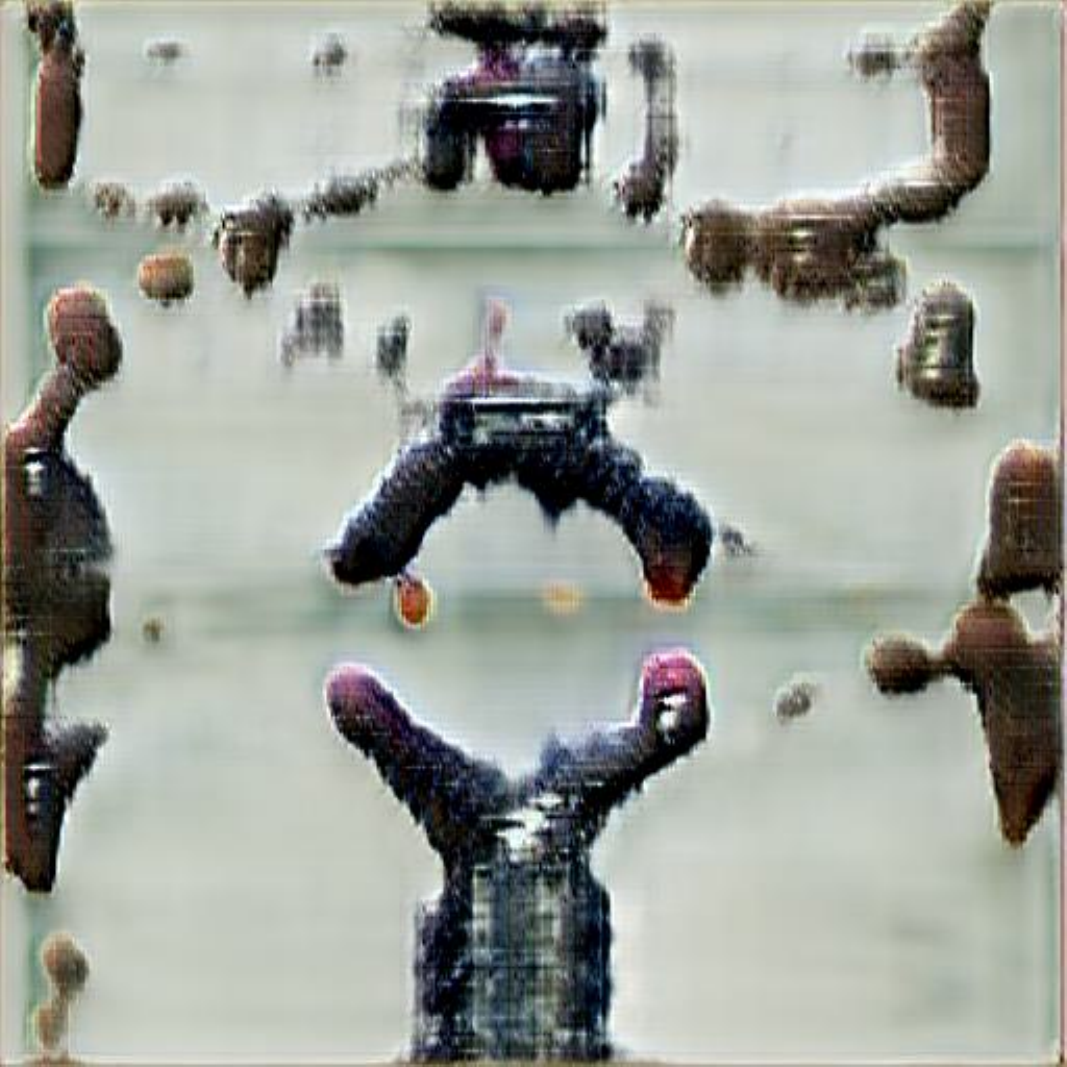}   &
      \includegraphics[width=0.24\linewidth]{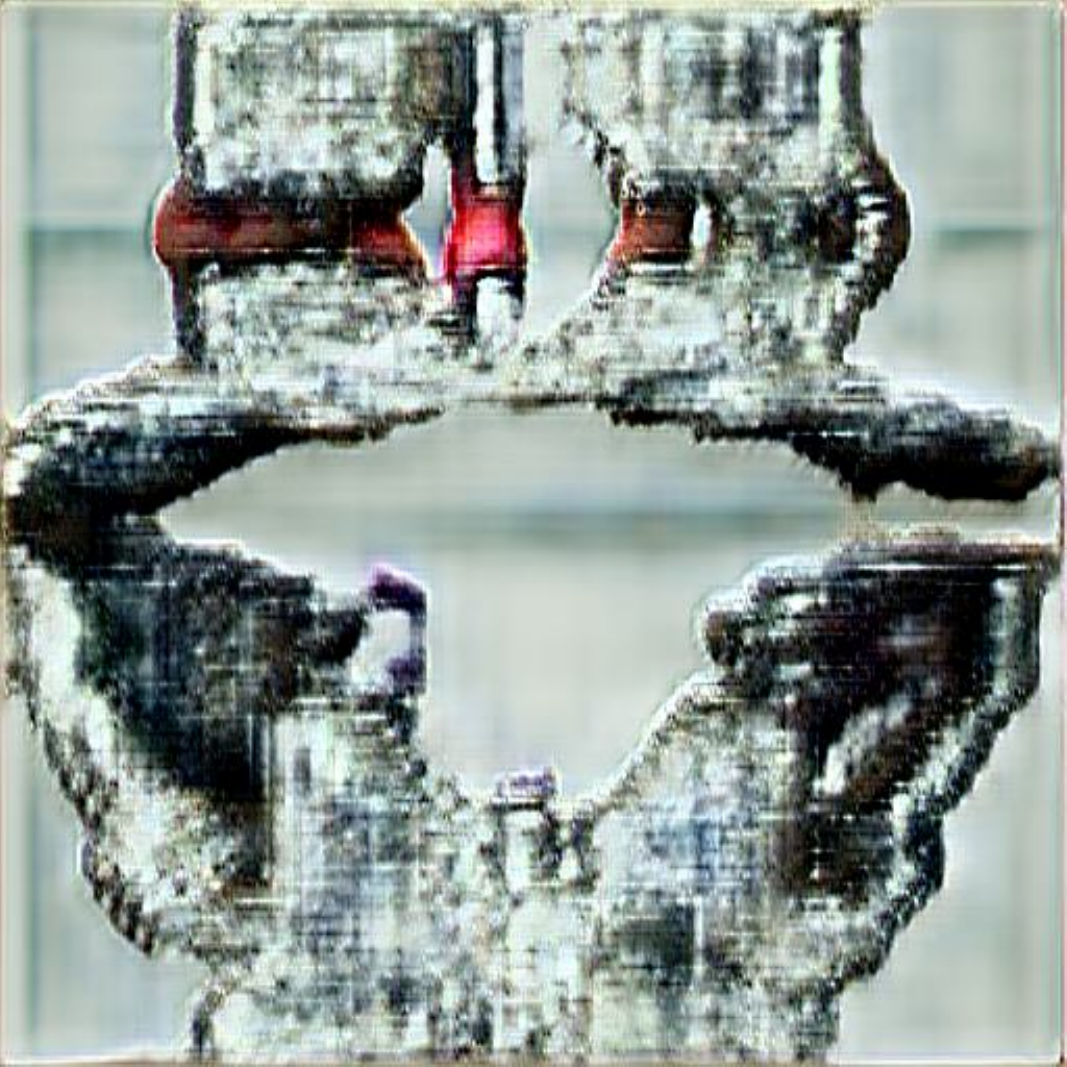}   
    \end{tabular}
\caption{First 16 outputs for the prompt "Brutalist anime robot", using the Convolutional + Sorting BM and InfoNCE loss.}
\label{fig:clip_tree}
\end{center}
\end{figure}

\clearpage
\section{Role of InfoNCE loss}
\label{sec:infonce}

In this section, we compare model outputs with a BM trained with the standard CLIP~\cite{radford2021learning} loss (squared great circle distance between embeddings of the image and caption) and with our novel contrastive loss, based on InfoNCE~\cite{oord2018representation}, which maximizes the similarity (mutual information) between image and caption embeddings while minimizing that among different images in a batch, to encourage output diversity. These outputs were generated using the coordinates-aware BM, and noise vectors generated using the same random seeds were fed to the two models, for a more direct comparison. The temperature $\tau$ for the InfoNCE loss was set at $0.001$.

We can observe that the InfoNCE-generated outputs tend to be more diverse, with several different color combinations and textures, while the standard CLIP outputs tend to be more uniform and to have a smoother quality. These different losses can be used to achieve different artistic effects.

\begin{figure}[h!] \
\begin{center}
\includegraphics[width=1.0\textwidth]{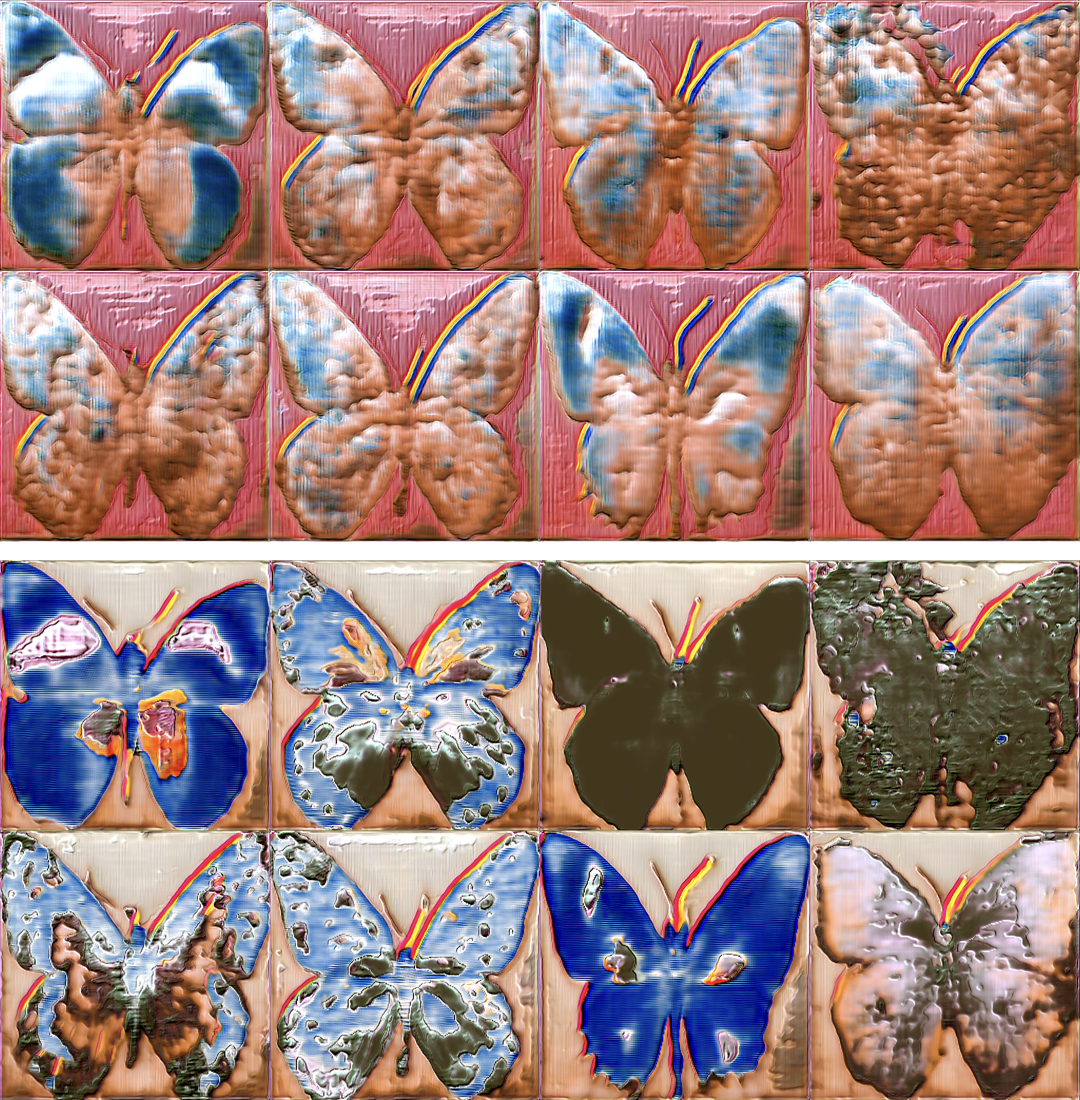}
\caption{Example outputs of a coordinates-aware BM with the standard CLIP loss (top) and InfoNCE loss (bottom). The prompt was "Inflatable plastic bodybuilder in a colorful album cover painted by Magritte".}
\label{fig:results}
\end{center}
\end{figure}

\begin{figure}[h!] \
\begin{center}
\includegraphics[width=1.0\textwidth]{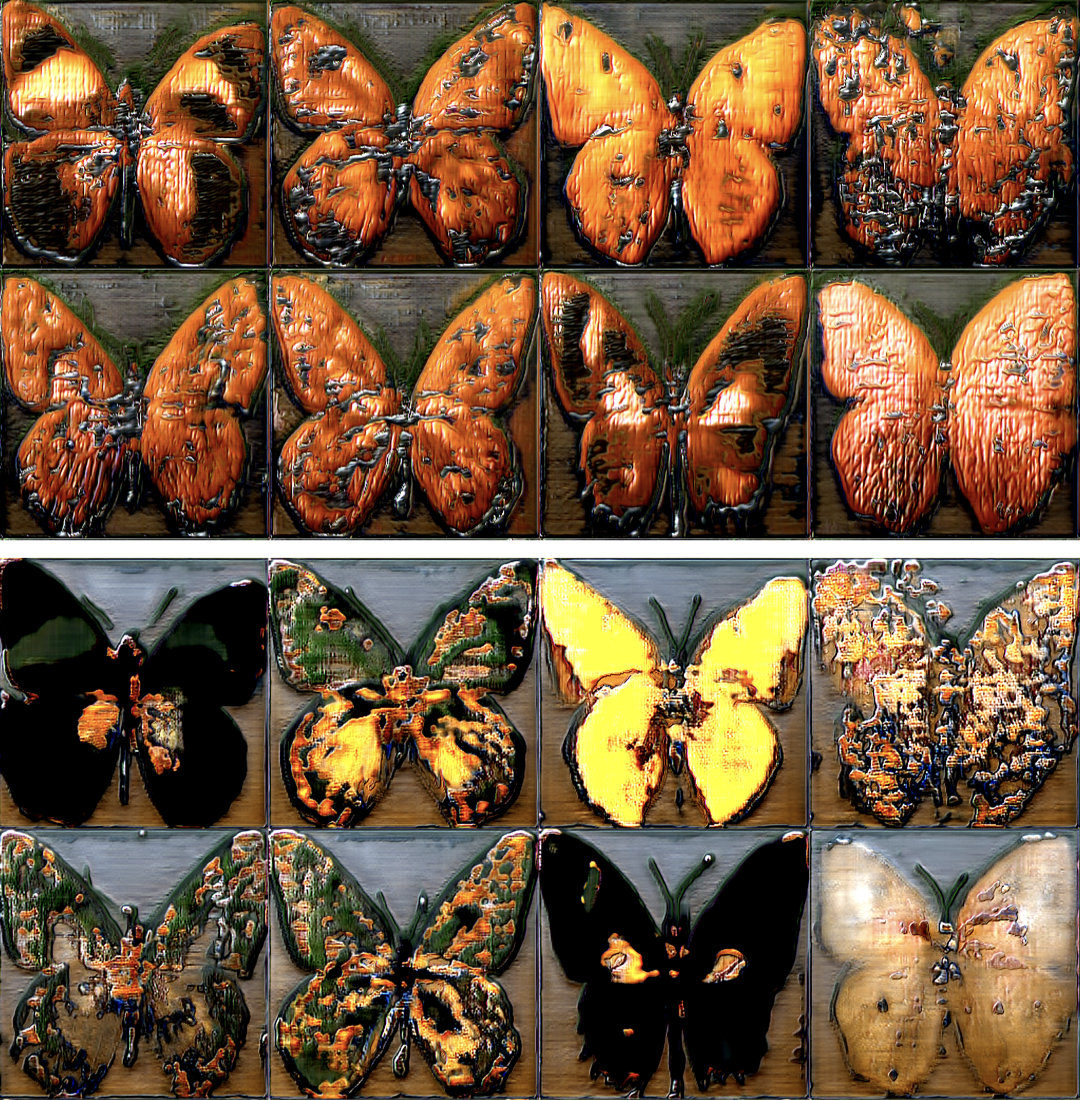}
\caption{Example outputs of a coordinates-aware BM with the standard CLIP loss (top) and InfoNCE loss (bottom). The prompt was "A gang of biker pumpkins painted by Jan van Eyck".}
\label{fig:results}
\end{center}
\end{figure}

\clearpage
\section{Role of spatial coordinates}
\label{sec:spatial_coords}

In this section, we showcase the difference between the coordinates-aware BM, in which $x$ and $y$ spatial coordinates are concatenated to the activation map, and the vanilla convolutional BM. Examples of the two models, for prompts that involve spatial structure, are shown in the figure below. From a qualitative inspection, we can see that the coordinates-aware BM segments different areas of the image more clearly, while the vanilla BM tends to mix different colors together. Somewhat surprisingly, the coordinates-aware BM does not generate the global spatial structure that would be expected given the prompt, such as the three vertical sections of the French flag and the sky being positioned above the landscape.

\begin{figure}[h!] \
\begin{center}
\includegraphics[width=1.0\textwidth]{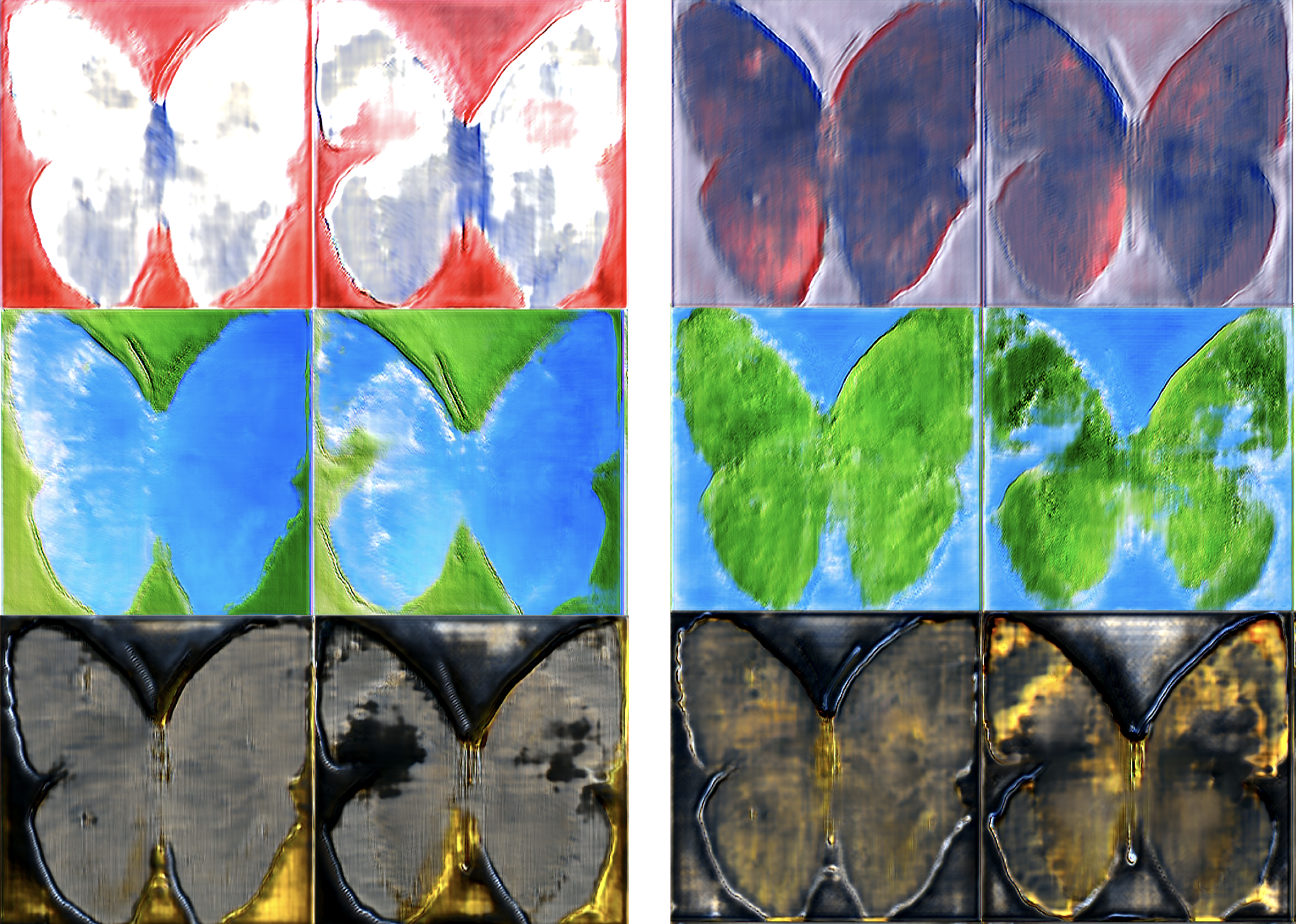}
\caption{Example outputs of a BM with (left) and without (right) spatial coordinates. The prompts were, from top to bottom, "French flag", "Green landscape and blue sky", and "Water and oil".}
\label{fig:results}
\end{center}
\end{figure}
\end{document}